\documentclass[10pt,twocolumn,letterpaper]{article}

\usepackage{iccv}
\usepackage{times}
\usepackage{epsfig}
\usepackage{graphicx}
\usepackage{algorithm} 
\usepackage{algorithmic} 
\usepackage{amsmath}
\usepackage{amssymb}
\usepackage{tabularx}
\usepackage{multirow}
\usepackage{color}
\usepackage{amsmath,amssymb,bm}
\usepackage{epstopdf}

\usepackage[breaklinks=true,bookmarks=false]{hyperref}

\iccvfinalcopy 

\def\iccvPaperID{****} 

\begin{document}

\title{Mode-Seeking on Hypergraphs for Robust Geometric Model Fitting}

\author{Hanzi Wang$^{1,*}$, Guobao Xiao$^{1,*}$, Yan Yan$^{1}$, David Suter$^2$\\
\small{$^1$Fujian Key Laboratory of Sensing and Computing for Smart City, School of Information Science and Engineering, Xiamen University, China}\\
\small{$^2$School of Computer Science, The University of Adelaide, Australia}\\
}

\maketitle

\begin{abstract}
   In this paper, we propose a novel geometric model fitting method, called Mode-Seeking on Hypergraphs (MSH), to deal with multi-structure data even in the presence of severe outliers. The proposed method formulates geometric model fitting as a mode seeking problem on a hypergraph in which vertices represent model hypotheses and hyperedges denote data points. MSH intuitively detects model instances by a simple and effective mode seeking algorithm. In addition to the mode seeking algorithm, MSH includes a similarity measure between vertices on the hypergraph and a ``weight-aware sampling'' technique. The proposed method not only alleviates sensitivity to the data distribution, but also is scalable to large scale problems. Experimental results further demonstrate that the proposed method has significant superiority over the state-of-the-art fitting methods on both synthetic data and real images.
\end{abstract}

\section{Introduction}
Geometric model fitting is a challenging research problem for a variety of applications in computer vision, such as optical flow calculation, motion segmentation and homography/fundamental matrix estimation. Given that data may contain outliers, the task of geometric model fitting is to robustly estimate the number and the parameters of model instances in the data\let\thefootnote\relax\footnotetext{*equal contribution}.
\begin{figure}[ht]
\centering
\begin{minipage}{.23\textwidth}
\centerline{\includegraphics[width=0.90\textwidth]{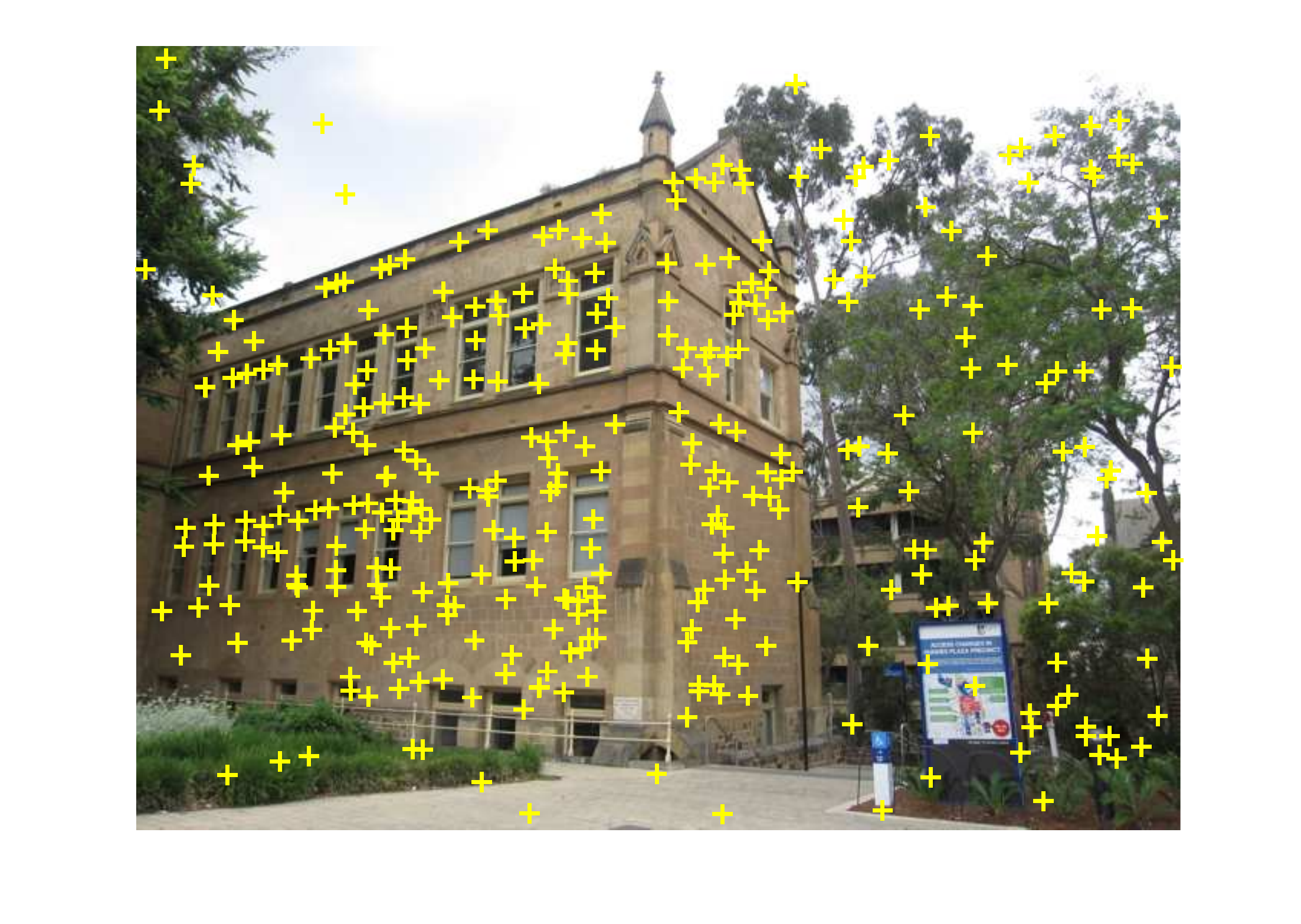}}
  \centerline{(a)}
  \centerline{\includegraphics[width=0.90\textwidth]{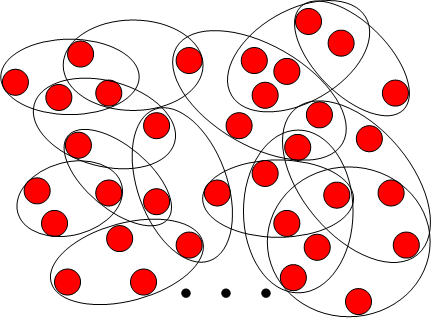}}
  \centerline{(c)}
    \centerline{\includegraphics[width=0.90\textwidth]{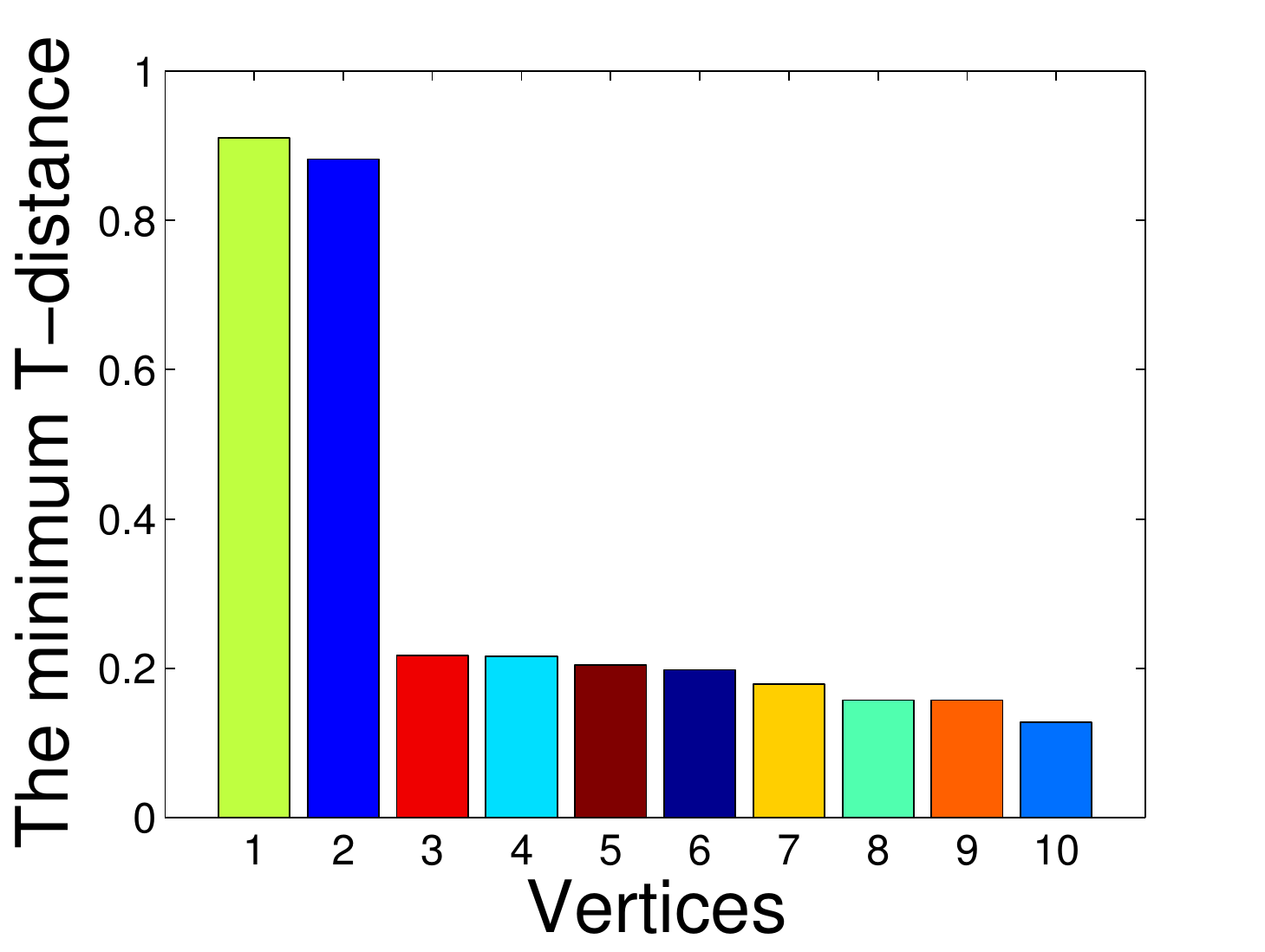}}
  \centerline{(e)}
\end{minipage}
\begin{minipage}{.23\textwidth}
\centerline{\includegraphics[width=0.90\textwidth]{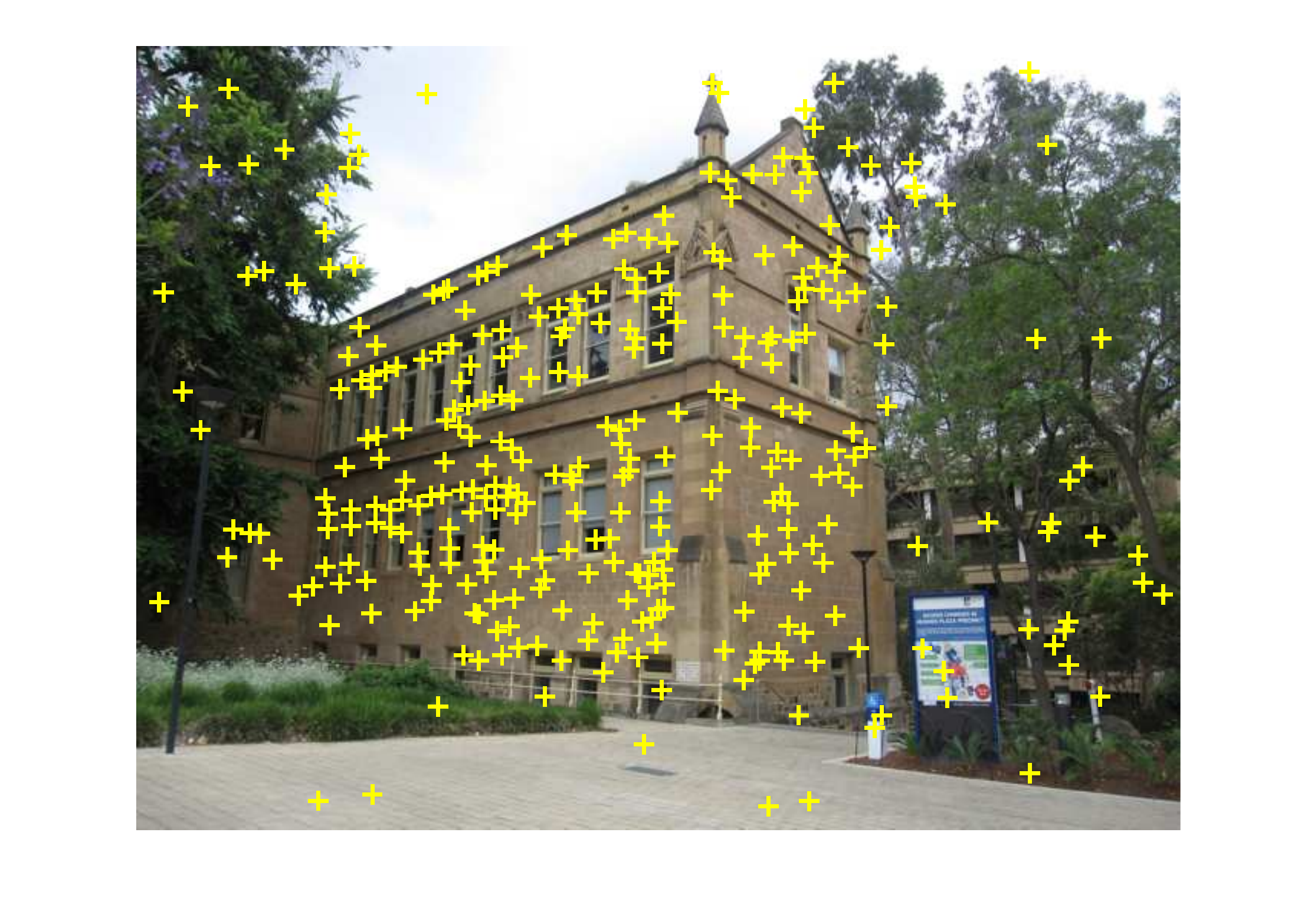}}
  \centerline{(b)}
  \centerline{\includegraphics[width=0.90\textwidth]{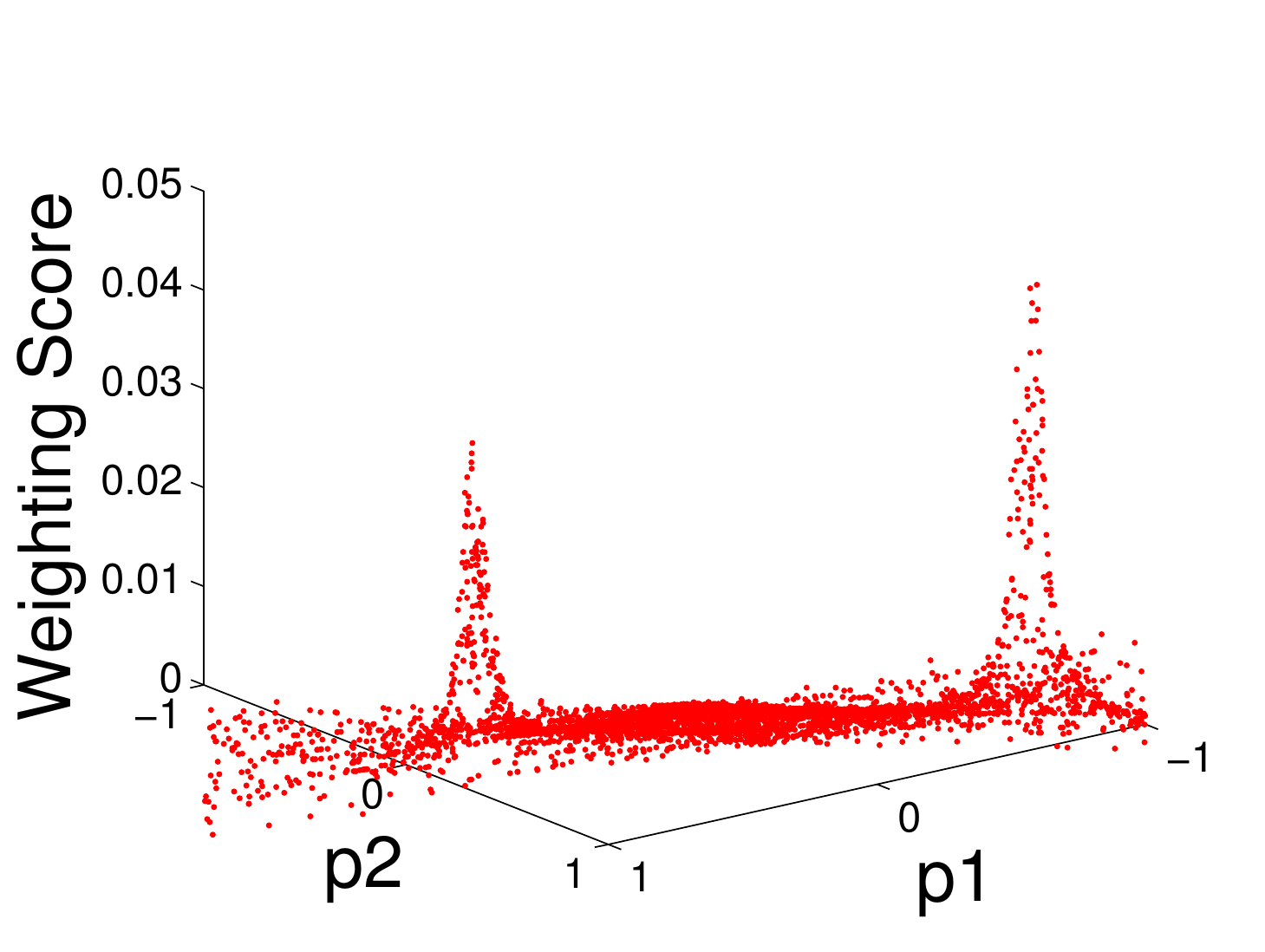}}
  \centerline{(d)}
    \centerline{\includegraphics[width=0.90\textwidth]{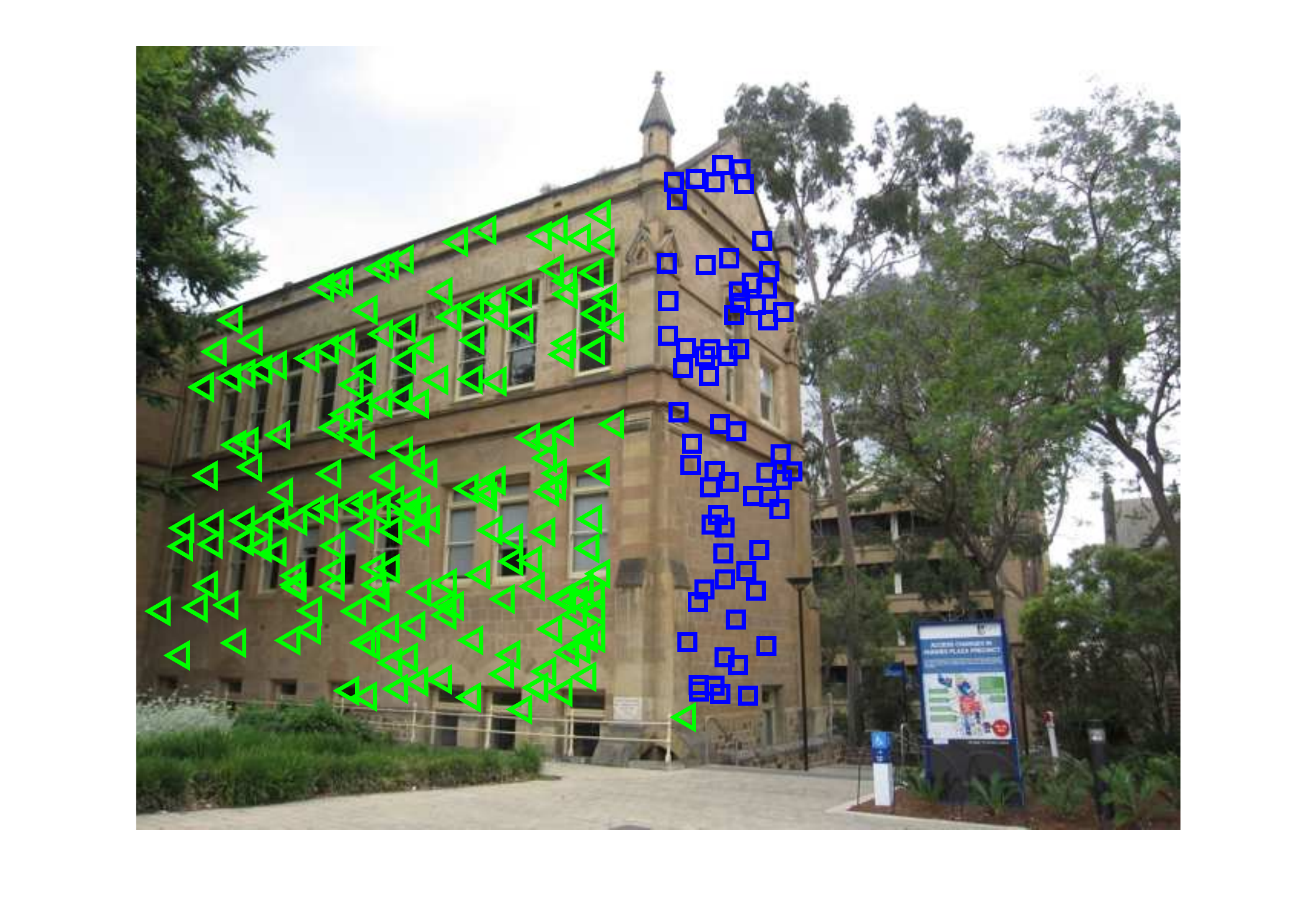}}
  \centerline{(f)}
\end{minipage}
\caption{Overview of the proposed algorithm: (a) and (b) An image pair with SIFT features. (c) Hypergraph modelling in which each vertex represents a model hypothesis and each hyperedge denotes a data point. (d) Weighted vertices (plotted using the first two parameters of the corresponding model hypotheses). (e) Mode seeking by searching for ``authority peaks'' on the hypergraph. (f) Data points segmented according to the detected modes.}
\label{fig:overview}
\end{figure}

A number of robust geometric model fitting methods (e.g., \cite{chin2009robust,fischler1981random,pearl:ijcv12,Magri_2014_CVPR,toldo2008robust,wang2012simultaneously}) have been proposed to work on the task. One of the most popular robust fitting methods is RANSAC~\cite{fischler1981random} due to its efficiency and simplicity. However, RANSAC is sensitive to the inlier scale and is originally designed to fit single-structure data. During the past few decades, many fitting methods have been proposed to deal with multi-structure data, such as KF~\cite{chin2009robust}, PEARL~\cite{pearl:ijcv12}, AKSWH~\cite{wang2012simultaneously}, J-linkage~\cite{toldo2008robust} and T-linkage~\cite{Magri_2014_CVPR}.

Recently, some hypergraph based methods, e.g., \cite{jain2013efficient,liu2012efficient,ochs2012higher,pulak2014clustering,wang2014shifting}, have been proposed for model fitting. Compared with a simple graph, a hypergraph involves high order similarities instead of pairwise similarities used on the graph and it can describe more complex relationships among modes of interest. For example, Liu and Yan \cite{liu2012efficient} proposed to use a random consensus graph (RCG) to fit structures in data. Purkait et al.~\cite{pulak2014clustering} proposed to use large hyperedges for face clustering and motion segmentation.

However, current fitting methods are still far from being practical to deal with real-world problems. Data clustering based fitting methods (e.g., J-linkage and KF), are often sensitive to unbalanced data (i.e., the numbers of inliers belonging to different model instances in data are significantly different), which is quite common in practical applications. In addition, these methods have difficulties in dealing with data points near the intersection of two model instances. Hypergraph based fitting methods (e.g.,~\cite{liu2012efficient,ochs2012higher}) often need to project from a hypergraph to an induced graph, which may cause information-loss and thus impact the accuracy of the methods. Other robust fitting methods (e.g., AKSWH~\cite{wang2012simultaneously}, T-linkage~\cite{Magri_2014_CVPR}, {HS~\cite{wang2014shifting}}, etc.) also have some specific problems, such as: some model hypotheses corresponding to model instances in data may be removed during the selection of significant hypotheses in AKSWH, and the computational cost of T-linkage is typically high due to the agglomerative clustering procedure, and HS also has a complexity problem due to the expansion and dropping strategy.

In this paper, we propose a simple and effective mode-seeking fitting algorithm on hypergraphs to fit and segment multi-structure data in the parameter space. The proposed method (MSH), starts from hypergraph modelling, in which a hypergraph is constructed based on inlier scale estimation for each dataset. Compared with the hypergraph constructed in the previous methods~\cite{jain2013efficient,liu2012efficient,ochs2012higher,pulak2014clustering}, where a hyperedge is constrained to connect with a fixed number of vertices, the hyperedges constructed in this paper can connect with varying number of vertices. We measure the weight of each vertex by using the non-parametric kernel density estimate technique \cite{wand1994kernel}. Based on the hypergraph, a novel mode seeking algorithm is proposed to intuitively detect modes by searching for ``authority peaks'', and we also sample vertices by using a ``weight-aware sampling'' technique to improve the effectiveness of the proposed method. Finally, we estimate the number and the parameters of model instances in data according to the detected modes. The main steps are shown in Fig.~\ref{fig:overview}.

The proposed method (MSH) has three main advantages over previous model fitting methods. First, the constructed hypergraphs can effectively represent the complex relationships among model hypotheses and data points, and it can be directly used for geometric model fitting. Second, MSH deals with geometric model fitting in the parameter space to alleviate sensitivity to the data distribution, even in the presence of seriously unbalanced data. Third, MSH implements mode seeking by analyzing the similarity between vertices on the hypergraphs, which is scalable to large scale problems. We demonstrate that MSH is a highly robust method for geometric model fitting by conducting extensive experimental evaluations and comparisons in Sec.~\ref{sec:experiments}.

\section{Hypergraphs and Weighting Score}
\label{sec:HypergraphModelling}
In this study, the geometric model fitting problem is formulated as a mode-seeking problem on a hypergraph. In Sec.~\ref{sec:problemsetting}, we express the relationships among model hypotheses and data points with the hypergraph, in which a vertex represents a model hypothesis and a hyperedge denotes a data point. We also assign each vertex a weighting score based on the non-parametric kernel density estimate technique \cite{wand1994kernel} in Sec.~\ref{sec:authority}.
\subsection{Hypergraphs}
\label{sec:problemsetting}
A hypergraph $G=(\mathcal{V},\mathcal{E}, \mathcal{W})$ consists of vertices $\mathcal{V}$, hyperedges  $\mathcal{E}$, and weights $\mathcal{W}$. Each vertex $v$ is weighted by a weighting score $w(v)$. When $v\in e$, a hyperedge $e$ is incident with a vertex $v$. Then an incident matrix $\mathbf{H}$, satisfying $h(v,e)=1$ if $v\in e$ and 0 otherwise, is used to represent the relationships between vertices and hyperedges in the hypergraph $G$. For a vertex $v\in \mathcal{V}$, its degree is defined by $\delta(v)=\sum_{e\in \mathcal{E}} h(v,e)$.

Now we describe the detailed procedure of hypergraph construction as follows: Given a set of data points $\mathbf{X}=\{\mathbf{x}_i\}_{i=1}^n$, we first sample a set of minimal subsets from $\mathbf{X}$. A minimal subset contains the minimum number of data points which is necessary to estimate a model hypothesis (e.g., $2$ for line fitting and $4$ for homography fitting). Then we generate a set of model hypotheses using the minimal subsets and estimate their inlier scales. In this paper, we use IKOSE~\cite{wang2012simultaneously} as the inlier scale estimator due to its efficiency. After that, we connect each vertex (i.e., a model hypothesis) to the corresponding hyperedges (i.e., the inliers of the model hypothesis).
Therefore, the complex relationships among model hypotheses and data points can be effectively characterized on by the hypergraph. In this manner, we can directly perform mode-seeking on the hypergraph for model fitting.
\subsection{Weighting Score}
\label{sec:authority}
We weight a model hypothesis (i.e., a vertex $v$) and assign a weighting score for the model hypothesis using the density estimate technique through the following equation which is similar to~\cite{wang2012simultaneously}
\begin{align}
\label{equ:score1}
\pi(v)=\frac{1}{n}\sum_{e\in\mathcal{E}} \frac{\Psi(r_e(v)/b(v))}{\hat{s}(v)b(v)},
\end{align}
where $\Psi(\cdot)$ is a kernel function (such as the Epanechnikov kernel); $r_e(v)$ is a residual measured with the Sampon Distance~\cite{torr1997development} from the model hypothesis ($v$) to a data point (i.e., a hyperedge $e$); $n$ and $\hat{s}(v)$ are the number of data points and the inlier scale of the model hypothesis, respectively; $b(v)$ is a bandwidth.

Since the ``good" model hypotheses corresponding to the model instances in data have significantly more data points with small residuals than the other ``bad" model hypotheses, the weighting scores of the vertices corresponding to the ``good" model hypotheses should be higher than those of the other vertices~\cite{wang2012simultaneously}. However, weighting a vertex based on residuals may be not robust to outliers, especially for extreme outliers. To weaken the impacts of outliers, we only consider the residuals of the corresponding inlier data points belonging to the model hypotheses. Thus, based on a hypergraph $G$, Eq.~(\ref{equ:score1}) can be rewritten as
\begin{align}
\label{equ:score}
w(v)=\frac{1}{\delta(v)}\sum_{e\in\mathcal{E}} \frac{h(v,e)\Psi(r_e(v)/b(v))}{\hat{s}(v)b(v)},
\end{align}
where $\delta(v)$ is the degree of vertex $v$ and $h(v,e)$ is an entry of the incident matrix $\mathbf{H}$ belonging to the hypergraph $G$.

Based on the weighting score, authority peaks on a hypergraph can be defined as follows:

\textbf{Definition 1}~~\emph{Authority peaks are the vertices that have the local maximum values of weighting scores on the hypergraph.}

The vertices that have the local maximum values of weighting scores correspond to the modes on a hypergraph, i.e., the model instances in data. This definition is consistent with the conventional concept of modes, which are defined as the significant peaks of the density distribution in the parameter space~\cite{cho2012mode,comaniciu2002mean,xu1990new}.

\section{Mode-Seeking on Hypergraphs}
\label{sec:proposedalgorithm}
In this section, we perform mode seeking by analyzing the similarity between vertices on a hypergraph. We develop an effective similarity measure between vertices in Sec.~\ref{sec:similaritymeasure} and propose a mode seeking algorithm in Sec.~\ref{sec:algorihmanalysis}. In addition, we further propose a weight-aware sampling (WAS) technique in Sec.~\ref{sec:authorityaware} to improve the effectiveness of the proposed algorithm.
\subsection{Similarity Measure}
\label{sec:similaritymeasure}
An effective similarity measure is proposed to describe the relationships between any two vertices in a hypergraph based on the Tanimoto distance \cite{tanimoto1957internal} (referred to as T-distance), which measures the degree of overlap between two hyperedge sets connected by two vertices.

Similar to \cite{Magri_2014_CVPR}, we first define the preference function of a vertex $v_p$ as
\begin{align}
\label{equ:preferencefunction1}
  \mathcal{C}_{v_p} &=\left\{ \begin{array}    {r@{\quad \quad} l}
\exp\{-\frac{r_e(v_p)}{\hat{s}(v_p)}\}, & if~r_e(v_p)\leq E\hat{s}(v_p),\\
0,&otherwise,
\end{array}\right.
\end{align}
where $E$ is a threshold ($E$ is usually set to 2.5 to include $98\%$ inliers of a Gaussian distribution). Note that the preference function of each vertex can be effectively expressed by Eq.~(\ref{equ:preferencefunction1}), which takes advantages of the information of residuals of data points.

Considering a hypergraph, we can rewrite Eq.~(\ref{equ:preferencefunction1}) as
\begin{align}
\label{equ:preferencefunction}
\mathcal{C}_{v_p}=h(v_p,e)\exp\{-\frac{r_e(v_p)}{\hat{s}(v_p)}\}, {\forall e\in\mathcal{E}}.
\end{align}

Then the T-distance between two vertices $v_p$ and $v_q$ based on the corresponding preference functions is given by \cite{tanimoto1957internal}
\begin{equation}
\label{equ:tdistance}
\begin{split}
\mathcal{T}(\mathcal{C}_{v_p},\mathcal{C}_{v_q})
=1-\frac{\langle\mathcal{C}_{v_p},\mathcal{C}_{v_q}\rangle}{\|\mathcal{C}_{v_p}\|^2+\|\mathcal{C}_{v_q}\|^2-\langle\mathcal{C}_{v_p},\mathcal{C}_{v_q}\rangle},
\end{split}
\end{equation}
where $\langle\cdot,\cdot\rangle$ and $\|\cdot\|$ indicate the standard inner product and the corresponding induced norm, respectively.

Although \cite{Magri_2014_CVPR} also employs the T-distance as a similarity measure, our use of T-distance has significant differences: 1) We define the preference function of a hyperedge set (i.e., the inlier data points) with respect to a vertex (i.e., a model hypothesis), while the authors in~\cite{Magri_2014_CVPR} define the preference function of model hypotheses with respect to a data point. We analyze the preference of a model hypothesis instead of a data point to alleviates sensitivity to the data distribution. 2) The T-distance in the proposed method is calculated without using iterative processes. In contrast, the T-distance in \cite{Magri_2014_CVPR} is iteratively calculated until an agglomerative clustering algorithm segments all data points. Therefore, the T-distance is used much more efficiently in this study than that in \cite{Magri_2014_CVPR}.
\subsection{The Mode Seeking Algorithm}
\label{sec:algorihmanalysis}
Given the vertices of a hypergraph $G$, we aim to seek modes by searching for authority peaks which correspond to model instances in data. Inspired by \cite{rodriguez2014clustering}, where each cluster center is characterized by two attributes (i.e., a higher local density than their neighbors and a relatively large distance from any point that has higher densities to itself), we search for authority peaks, which are the vertices that are not only surrounded by their neighbors with lower local weighting scores, but also significantly dissimilar to any other vertices that have higher local weighting scores.

More specifically, based on the similarity measure and weighting scores, we compute the Minimum T-Distance (MTD) $\eta^v_{min}$ of a vertex $v$ in $G$ as follows:
\begin{align}
\label{equ:minimumtdistance}
\eta^v_{min}=\min_{v_i\in \Omega(v)}\{\mathcal{T}(\mathcal{C}_{v},\mathcal{C}_{v_i})\},
\end{align}
where $\Omega(v)=\{v_i|v_i\in\mathcal{V},w(v_i)>w(v)\}$. That is, $\Omega(v)$ contains all vertices with higher weighing scores than $w(v)$ in $G$. For the vertex $v_max$ with the highest weighting score, we set $\eta^{v_max}_{min}=\max\{\mathcal{T}(\mathcal{C}_{v_max},\mathcal{C}_{v_i})\}_{v_i\in\mathcal{V}}$.

Note that a vertex with the local maximum value of weighting score, has a larger MTD value than the other vertices in $G$. Therefore, we propose to seek modes by searching for the authority peaks, i.e., the vertices with significantly large MTD values.

\begin{figure}[ht]
\centering
\begin{minipage}{.20\textwidth}
\centerline{\includegraphics[width=1.02\textwidth]{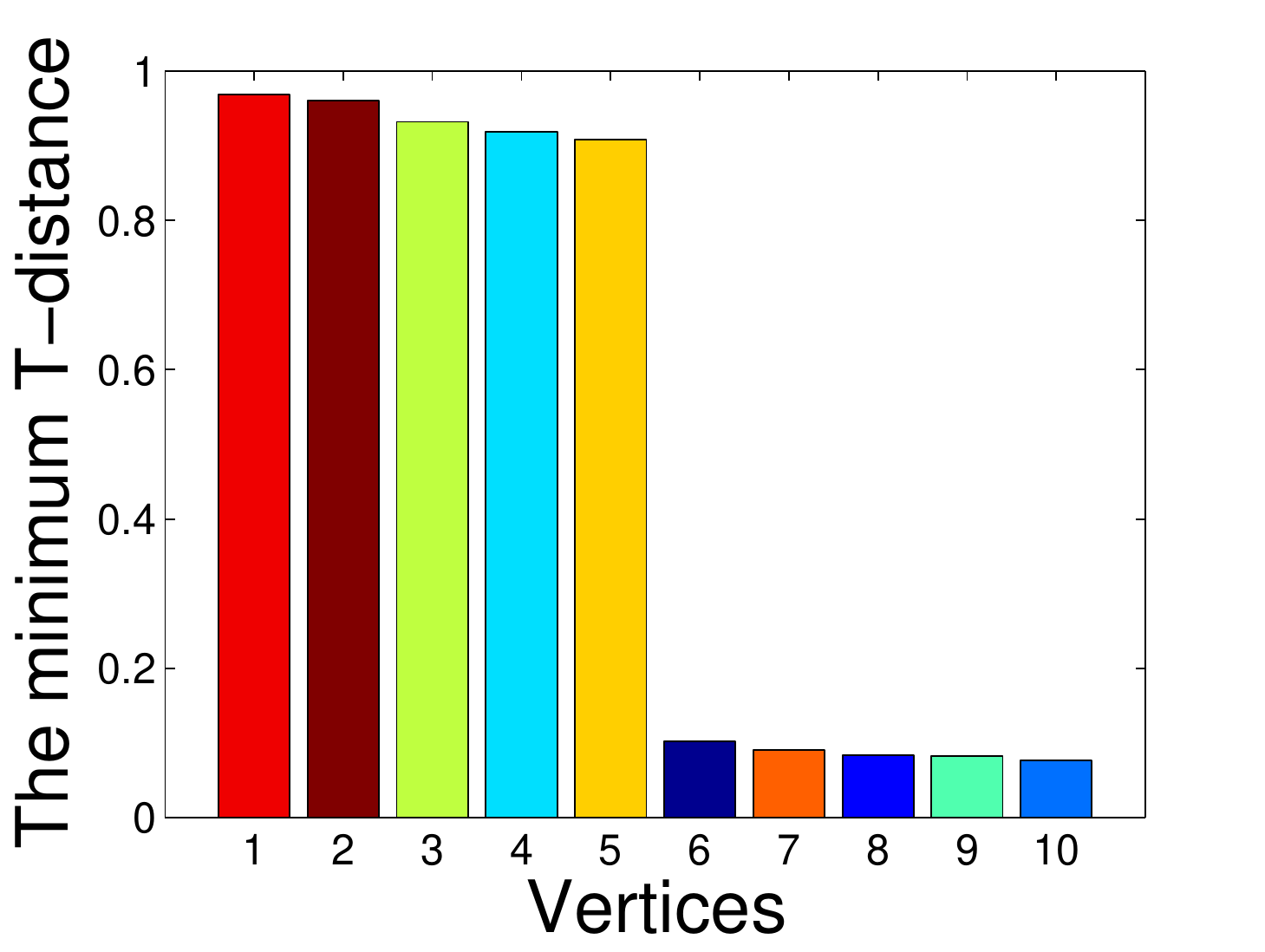}}
  \begin{center} (a) \end{center}
\end{minipage}
\begin{minipage}{.20\textwidth}
\centerline{\includegraphics[width=1.02\textwidth]{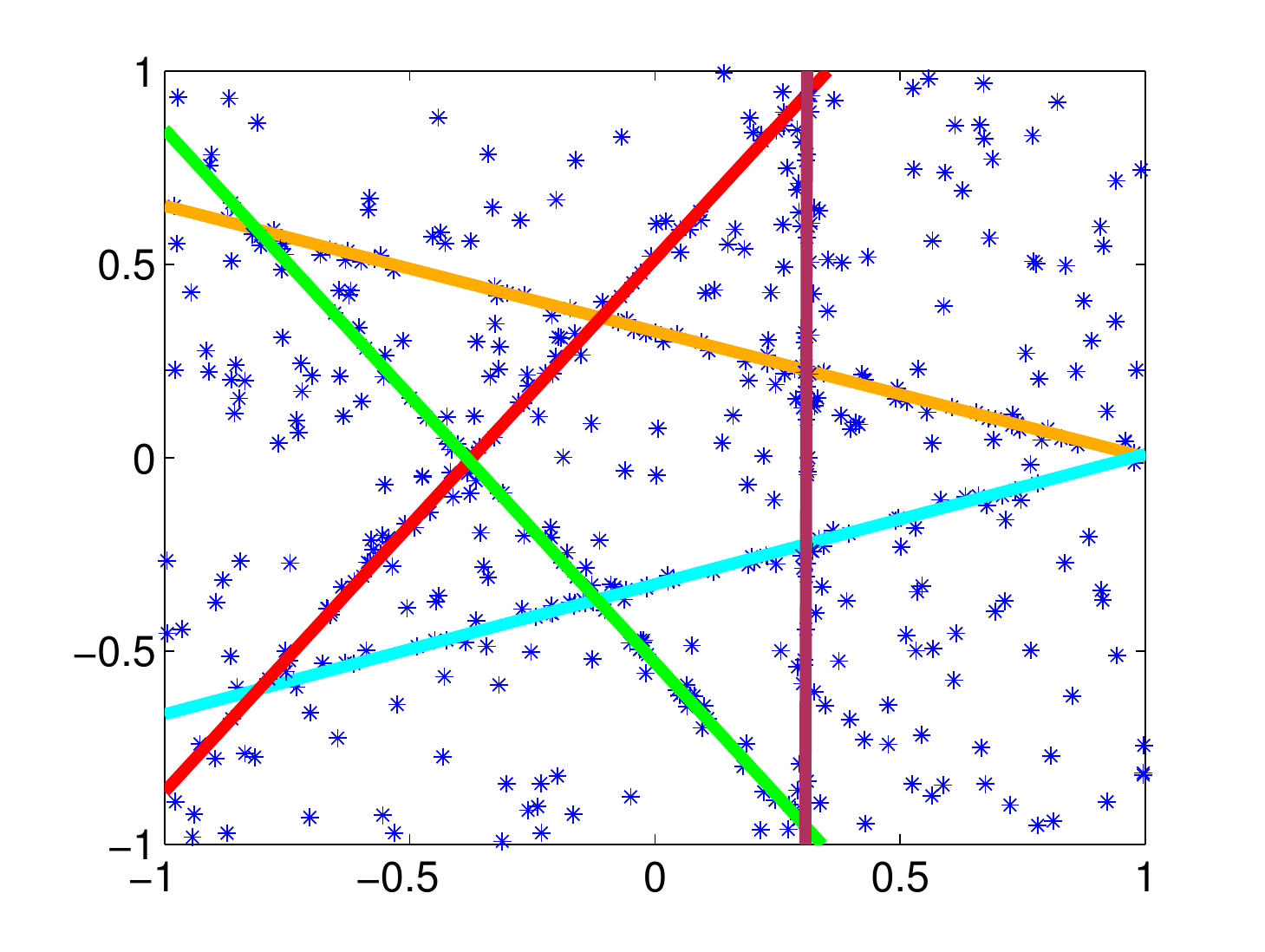}}
  \begin{center} (b) \end{center}
\end{minipage}
\caption{Line fitting on the ``Star5'' dataset. (a) The top $10$ largest MTD values of the corresponding vertices. (b) The five lines corresponding to the vertices with the top $5$ largest MTD values.}
\label{fig:twolinedistribution}
\end{figure}
We further illustrate the proposed mode seeking algorithm by using a simple example on the ``Star5'' dataset. Fig.~\ref{fig:twolinedistribution}(a) shows the top $10$ largest MTD values belonging to the corresponding vertices (sorted in the descending order). We can see that the top $5$ largest MTD values are significantly larger than those of the other vertices, and the lines corresponding to the vertices with the top $5$ largest MTD values are shown in Fig.~\ref{fig:twolinedistribution}(b).

The proposed mode seeking algorithm works well for line fitting. This is because the distribution of model hypotheses generated for line fitting is dense in the parameter space. However, the distribution of model hypotheses generated for higher order model fitting applications, such as homography based segmentation or two-view based motion segmentation, is often sparse, in which a few bad model hypotheses (with low weighting score values) may show anomalously large MTD values as good model hypotheses (with high weighting score values). This problem will cause the proposed algorithm to seek modes ineffectively.

\subsection{The Weight-Aware Sampling Technique}
\label{sec:authorityaware}
To solve the above problem, we further propose a simple technique called the weight-aware sampling (WAS) technique, which samples vertices according to the weighting scores on a hypergraph $G$. In WAS, the probability of sampling a vertex $v$ is computed as $w(v)/\sum_{v\in\mathcal{V}}w(v)$. As mentioned before, vertices corresponding to good model hypotheses often have significantly higher weighting score values than the other vertices. Thus WAS tends to sample good model hypotheses while rejecting bad model hypotheses. Therefore, for a few bad model hypotheses that may also show anomalously large MTD values, the probability of the vertices corresponding to these bad model hypotheses are sampled is quite low due to their low weighting score values.

To improve the effectiveness of the proposed mode seeking algorithm (as analyzed above), we use WAS to sample vertices of $G$ to approximate $G$, obtaining a new hypergraph $G^*$. Then we directly perform mode seeking by searching for authority peaks on $G^*$ instead of $G$. In this manner, we can find that a vertex, which is regarded as an authority peak, not only has a high weighting score but also has a large MTD value.
%
\begin{figure}[ht]
\centering
\begin{minipage}{.20\textwidth}
\centerline{\includegraphics[width=1.02\textwidth]{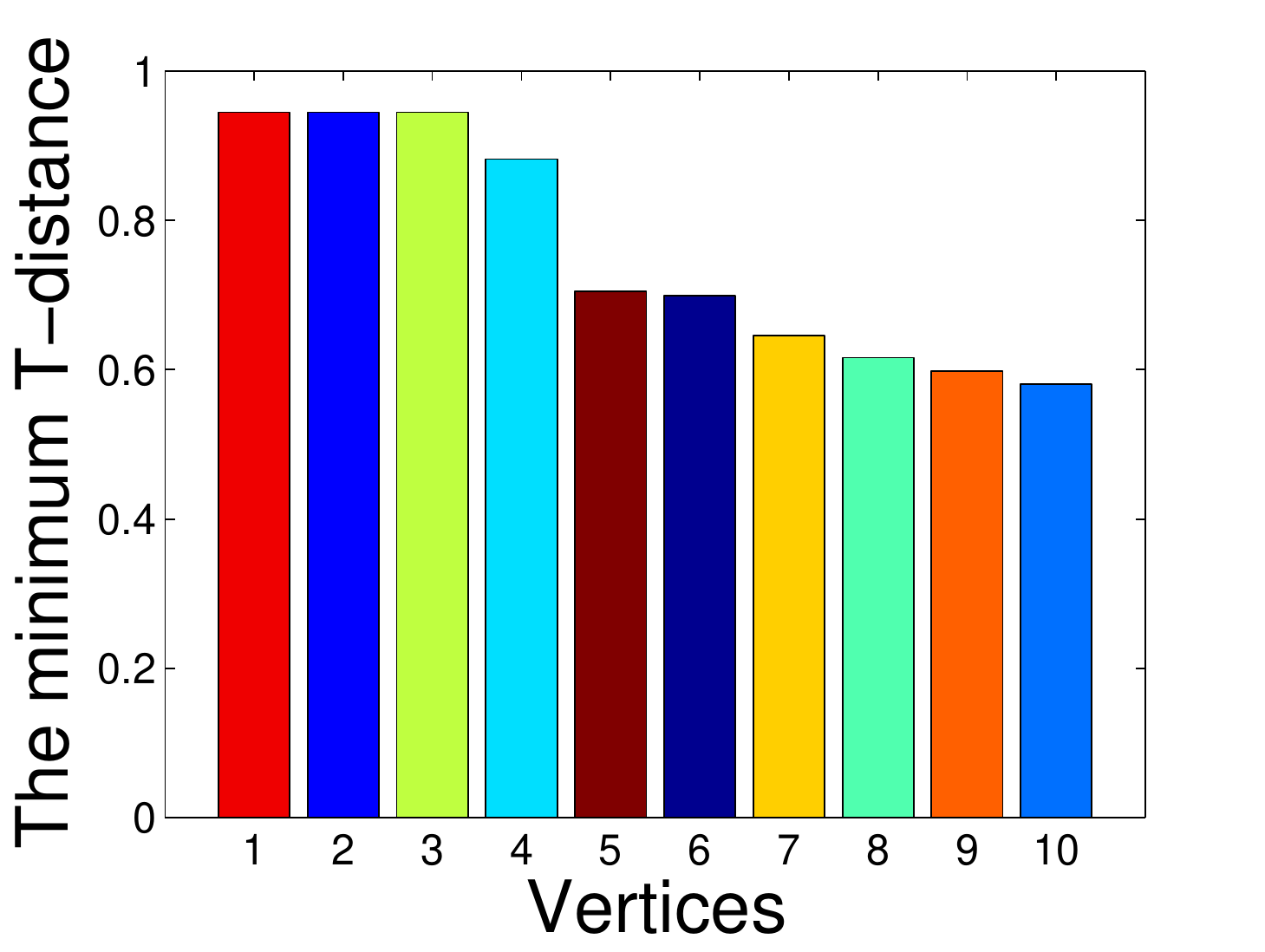}}
  \centerline{(a)}
  \centerline{\includegraphics[width=1.02\textwidth]{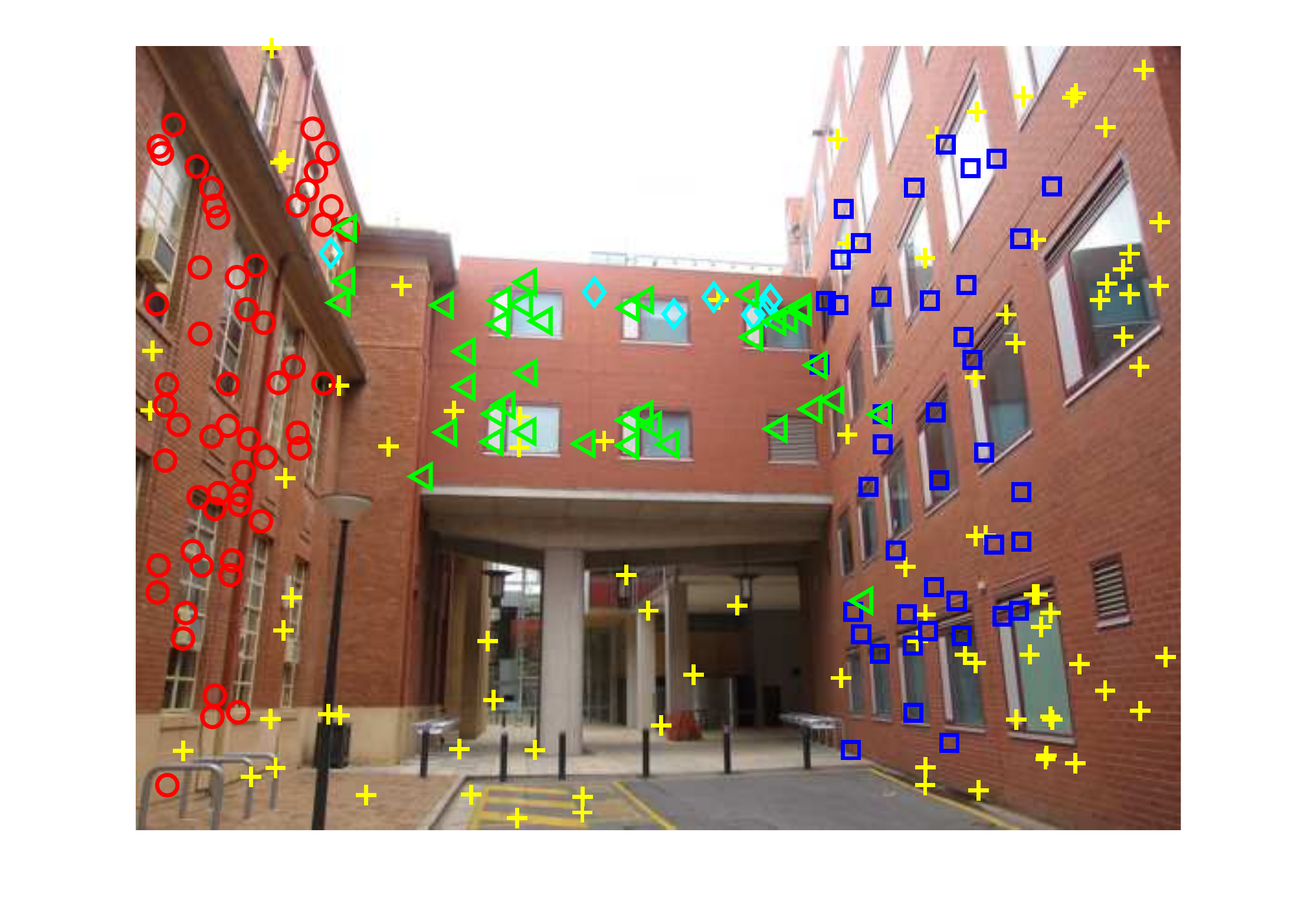}}
  \centerline{(c)}
\end{minipage}
\begin{minipage}{.20\textwidth}
\centerline{\includegraphics[width=1.02\textwidth]{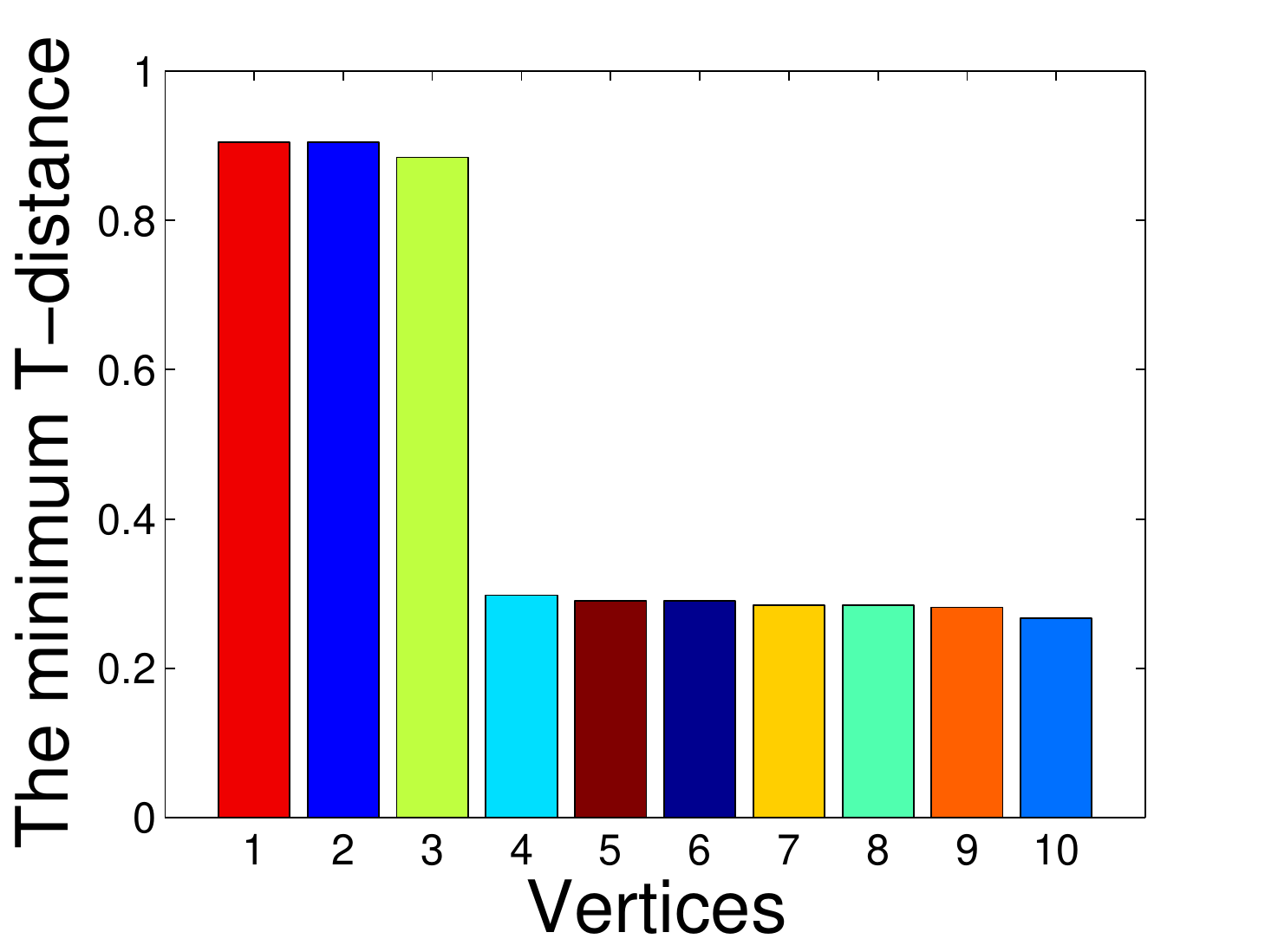}}
  \centerline{(b)}
  \centerline{\includegraphics[width=1.02\textwidth]{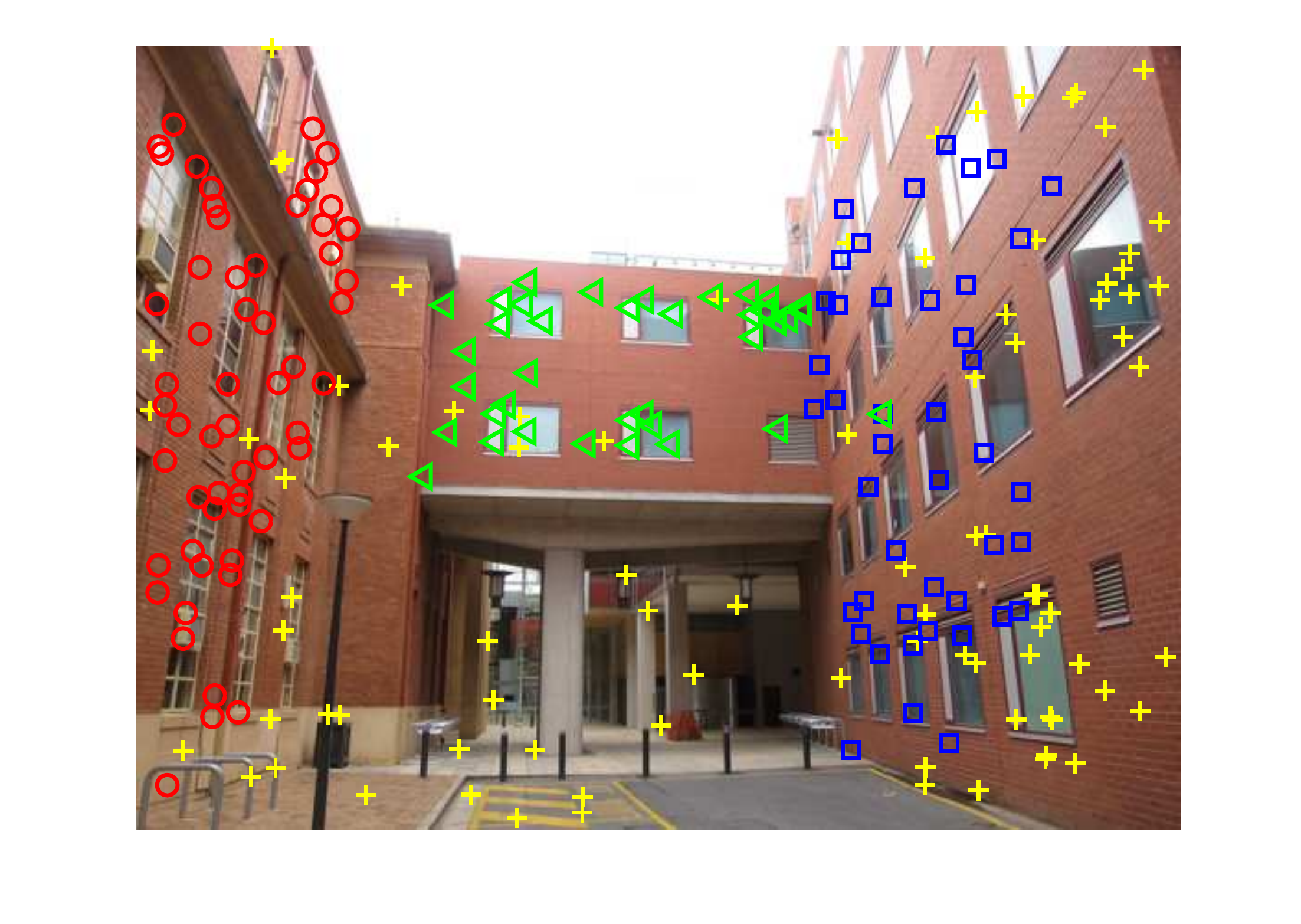}}
  \centerline{(d)}
\end{minipage}
\caption{Homography based segmentation on the ``Neem''~\cite{wong2011dynamic}. (a) and (b) The top $10$ largest MTD values of the corresponding vertices obtained by the proposed mode seeking algorithm based on $G$ and $G^*$, respectively. (c) and (d) The segmentation results obtained by the proposed MSH method based on $G$ and $G^*$, respectively.}
\label{fig:homotop10}
\end{figure}

To show the influence of WAS on the performance of the proposed mode seeking algorithm, we evaluate the algorithm for fitting multiple homographies based on the two hypergraphs, i.e., $G$ and $G^*$, as shown in Fig.~\ref{fig:homotop10}. We show the top $10$ largest MTD values (sorted in descending order) in Fig.~\ref{fig:homotop10}(a) and Fig.~\ref{fig:homotop10}(b) which correspond to $G$ and $G^*$, respectively. We can see that the proposed mode seeking algorithm based on $G$ has difficulty to distinguish the three significant model hypotheses from the MTD values. In contrast, the proposed mode seeking algorithm based on $G^*$ can effectively find the three significant model hypotheses by seeking the largest drop in the MTD values. As shown in Fig.~\ref{fig:homotop10}(c) and \ref{fig:homotop10}(d), the segmentation results further show the influence of WAS on the proposed MHS method--leading to more accurate results.
\begin{figure*}[ht]
\centering
\begin{minipage}[t]{.15\textwidth}
  \centerline{\includegraphics[width=1.00\textwidth]{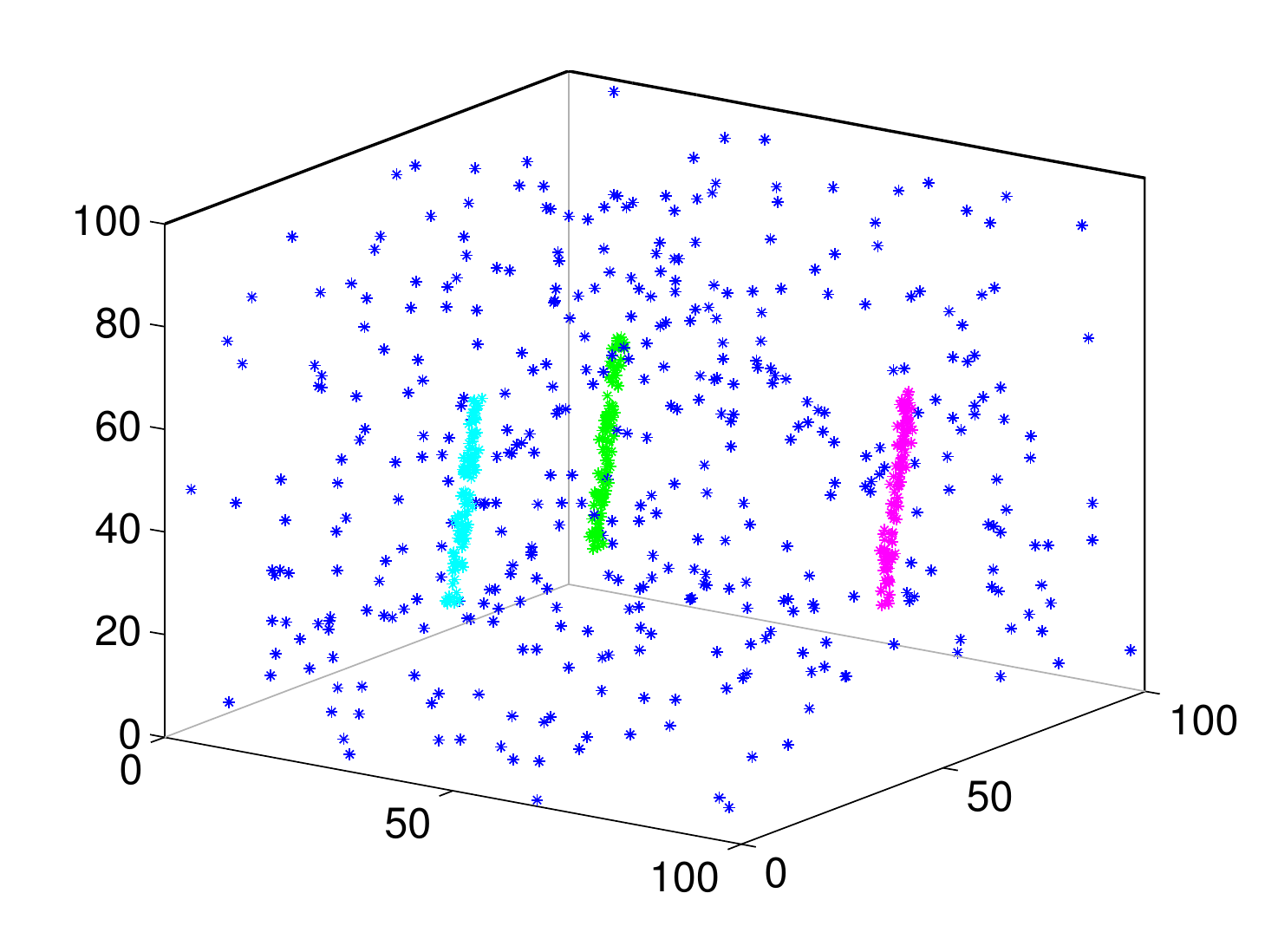}}
  \centerline{\includegraphics[width=1.00\textwidth]{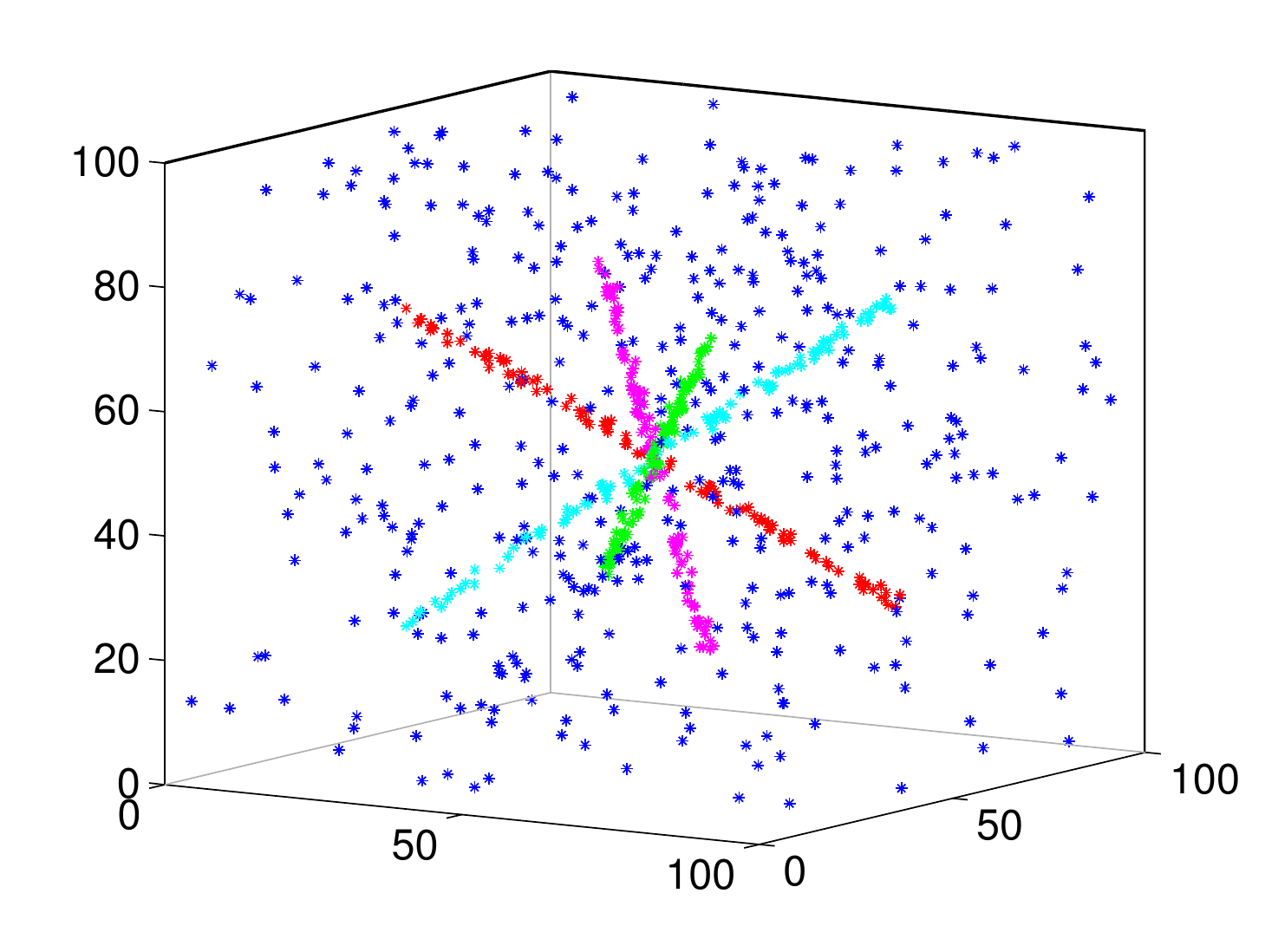}}
  \centerline{\includegraphics[width=1.00\textwidth]{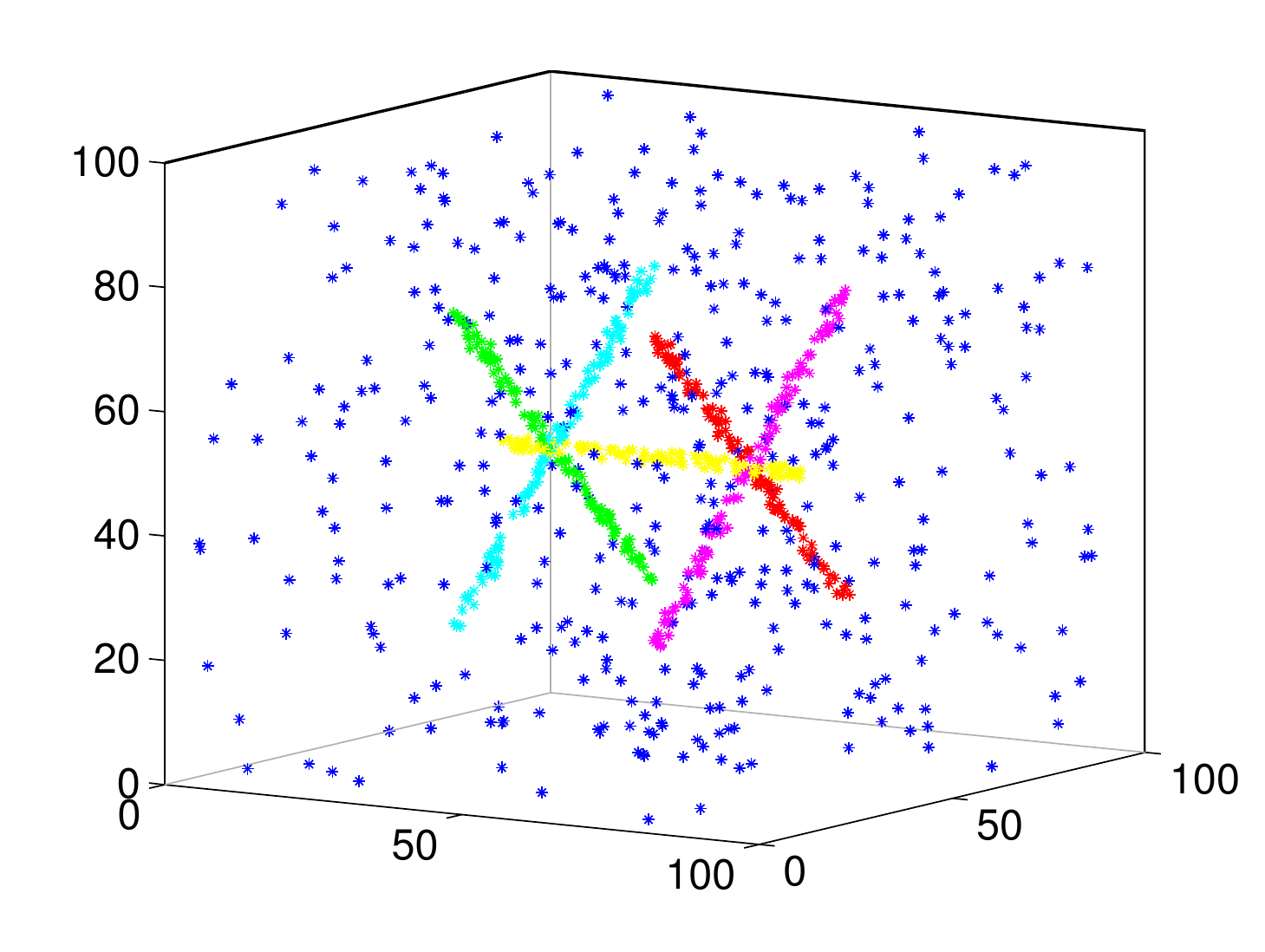}}
  \centerline{\includegraphics[width=1.00\textwidth]{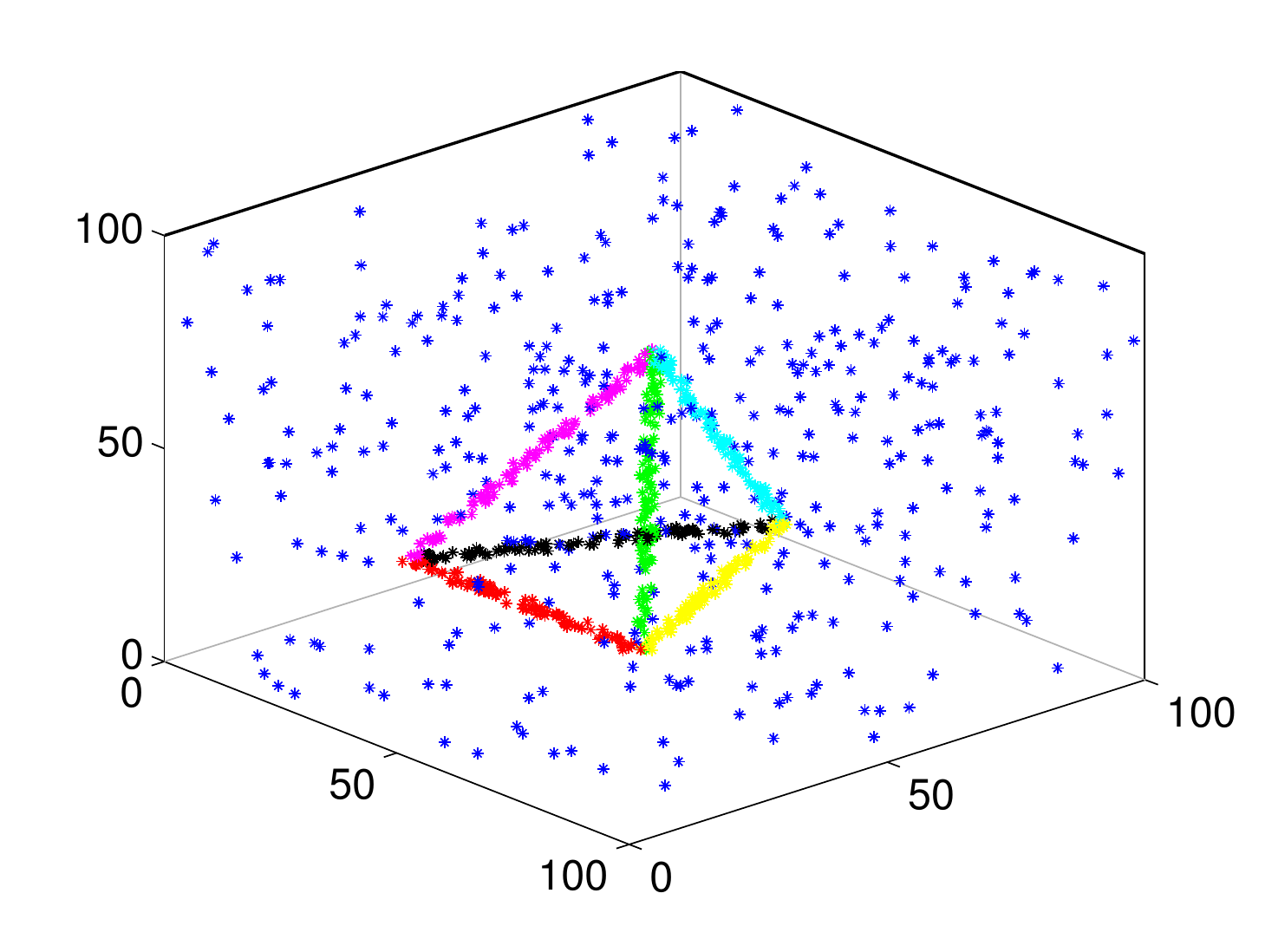}}
  \begin{center} (a) Datasets  \end{center}
\end{minipage}
\begin{minipage}[t]{.15\textwidth}

 \centerline{\includegraphics[width=1.00\textwidth]{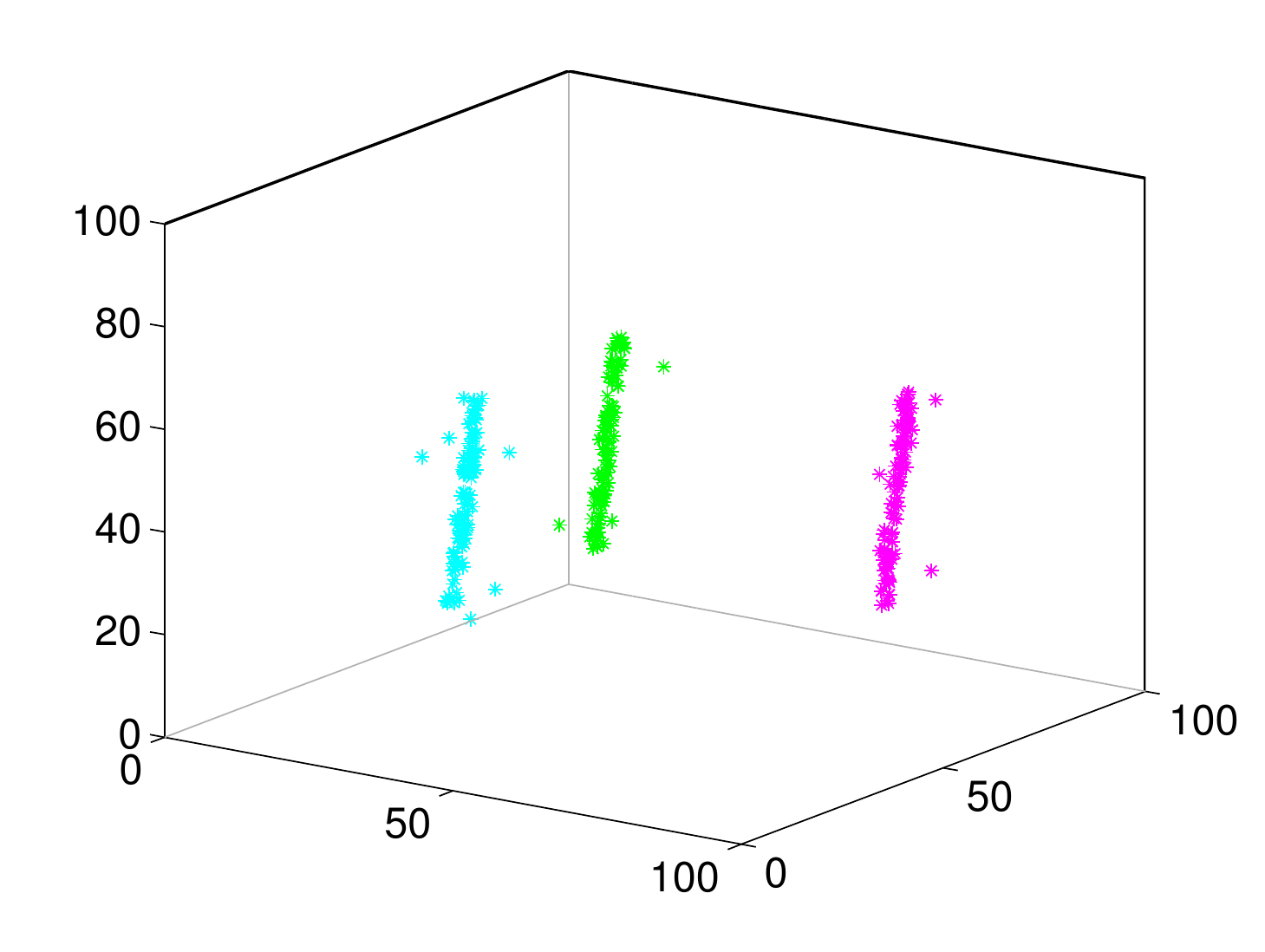}}
  \centerline{\includegraphics[width=1.00\textwidth]{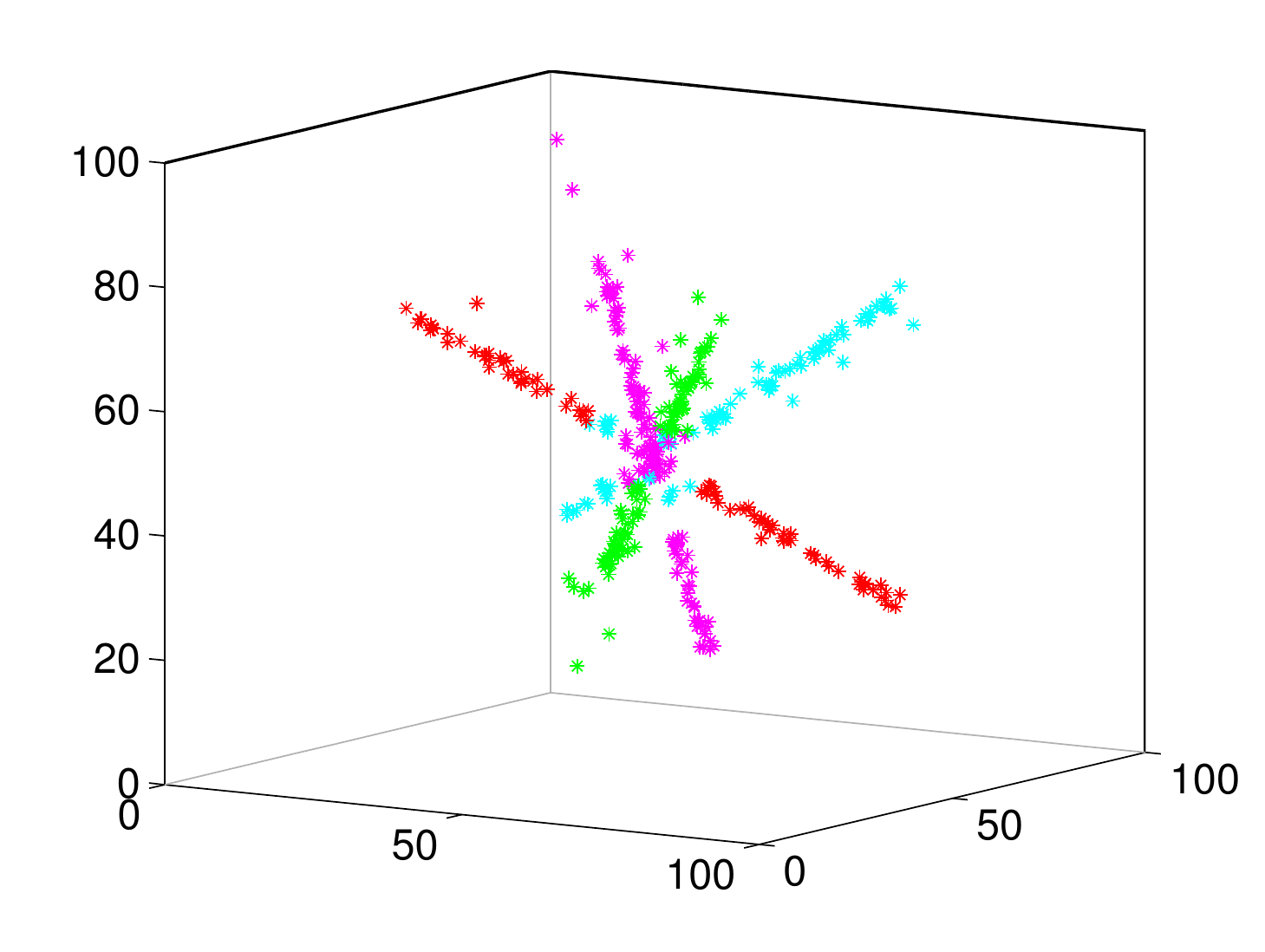}}
  \centerline{\includegraphics[width=1.0\textwidth]{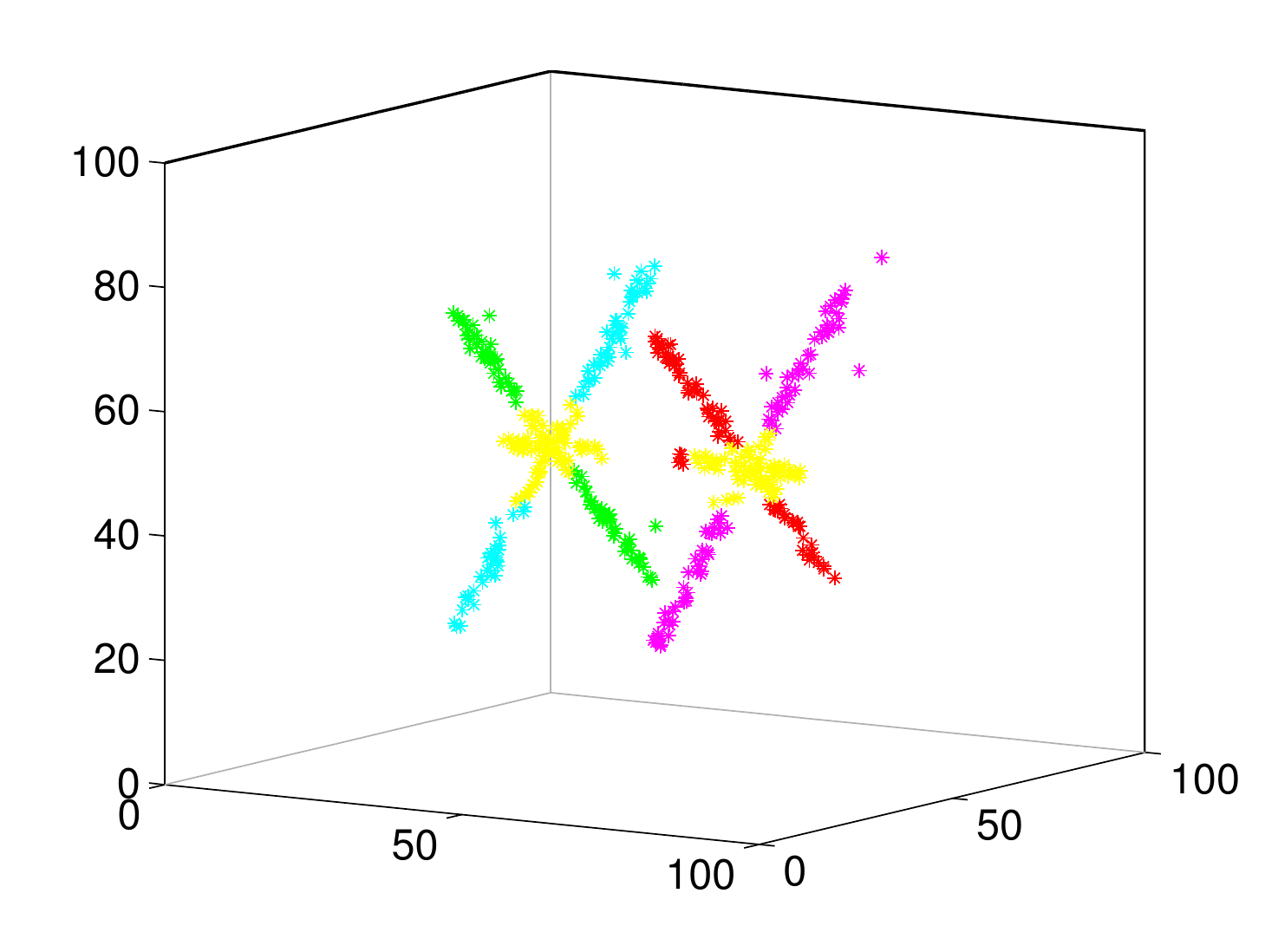}}
  \centerline{\includegraphics[width=1.0\textwidth]{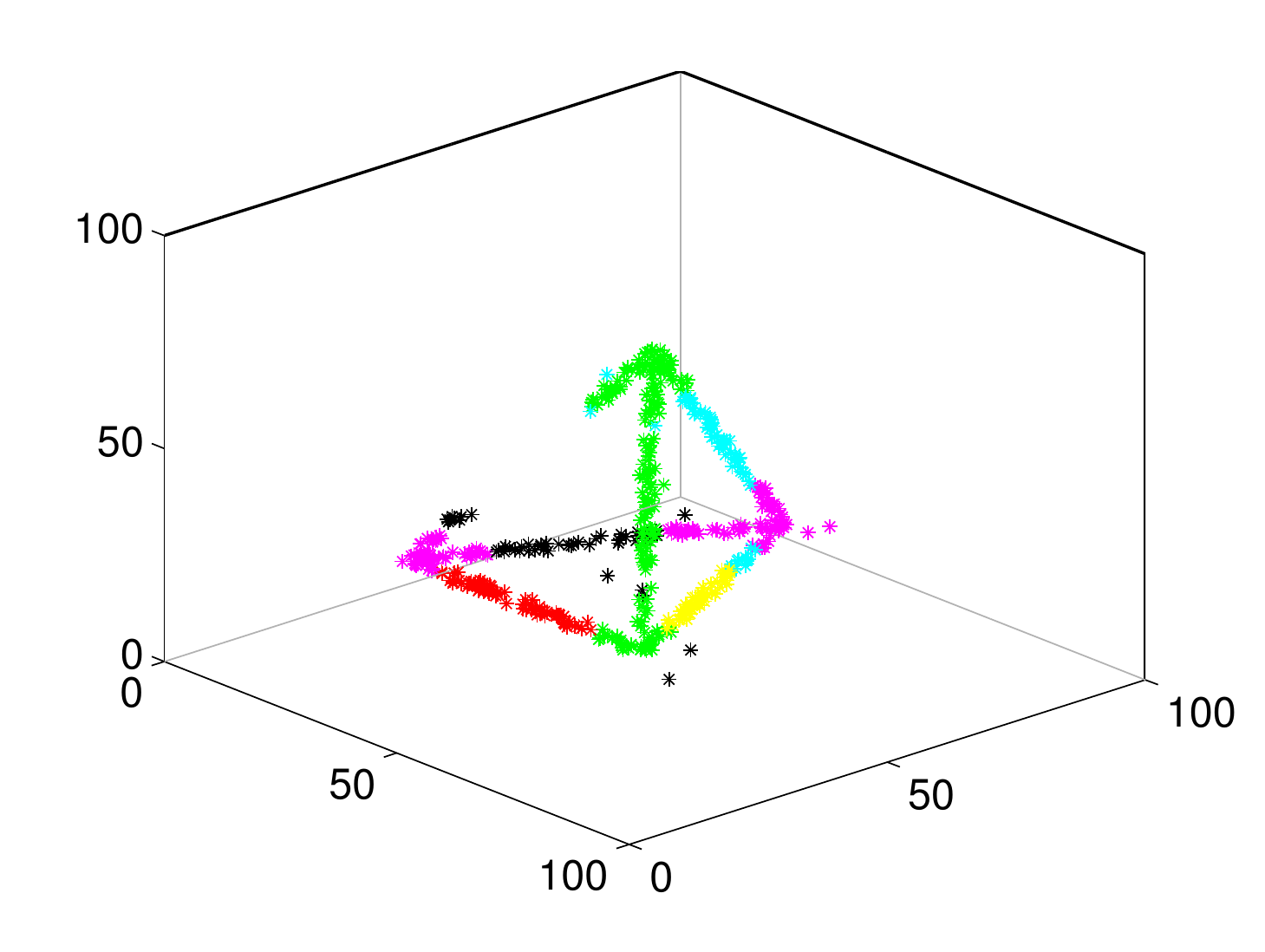}}
  \centerline{(b) KF }\medskip
\end{minipage}
\begin{minipage}[t]{.15\textwidth}
\centerline{\includegraphics[width=1.0\textwidth]{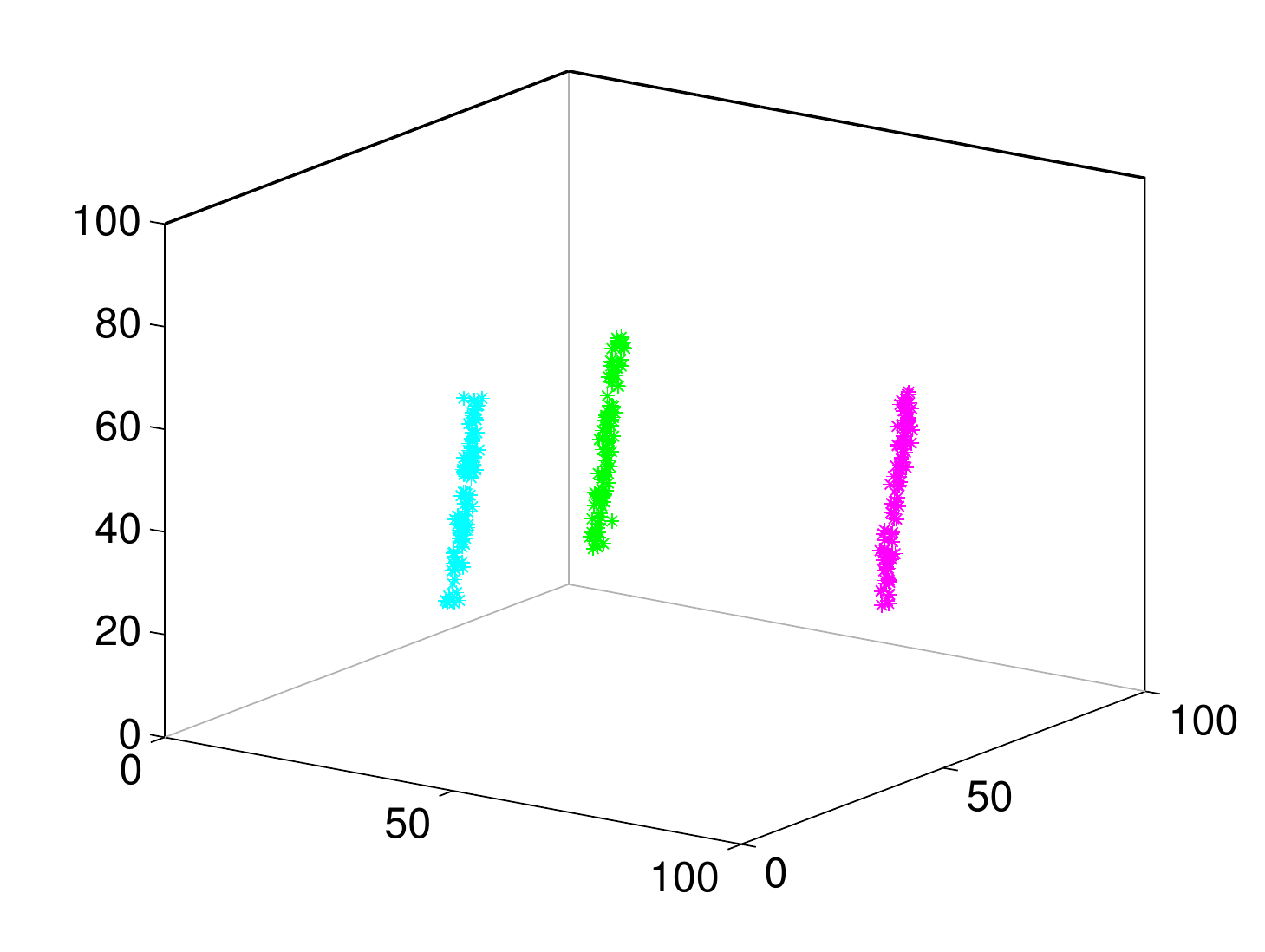}}
  \centerline{\includegraphics[width=1.0\textwidth]{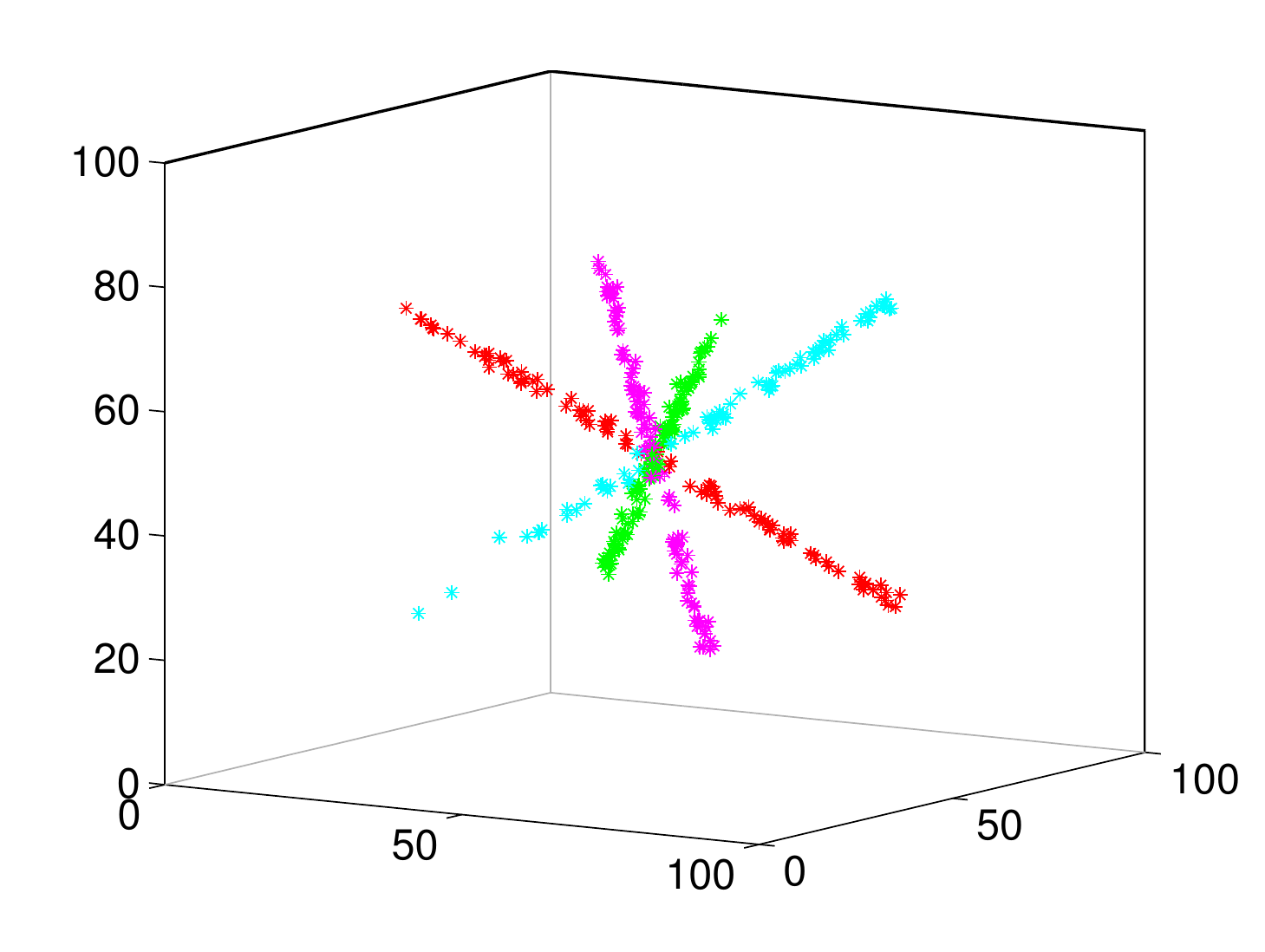}}
  \centerline{\includegraphics[width=1.0\textwidth]{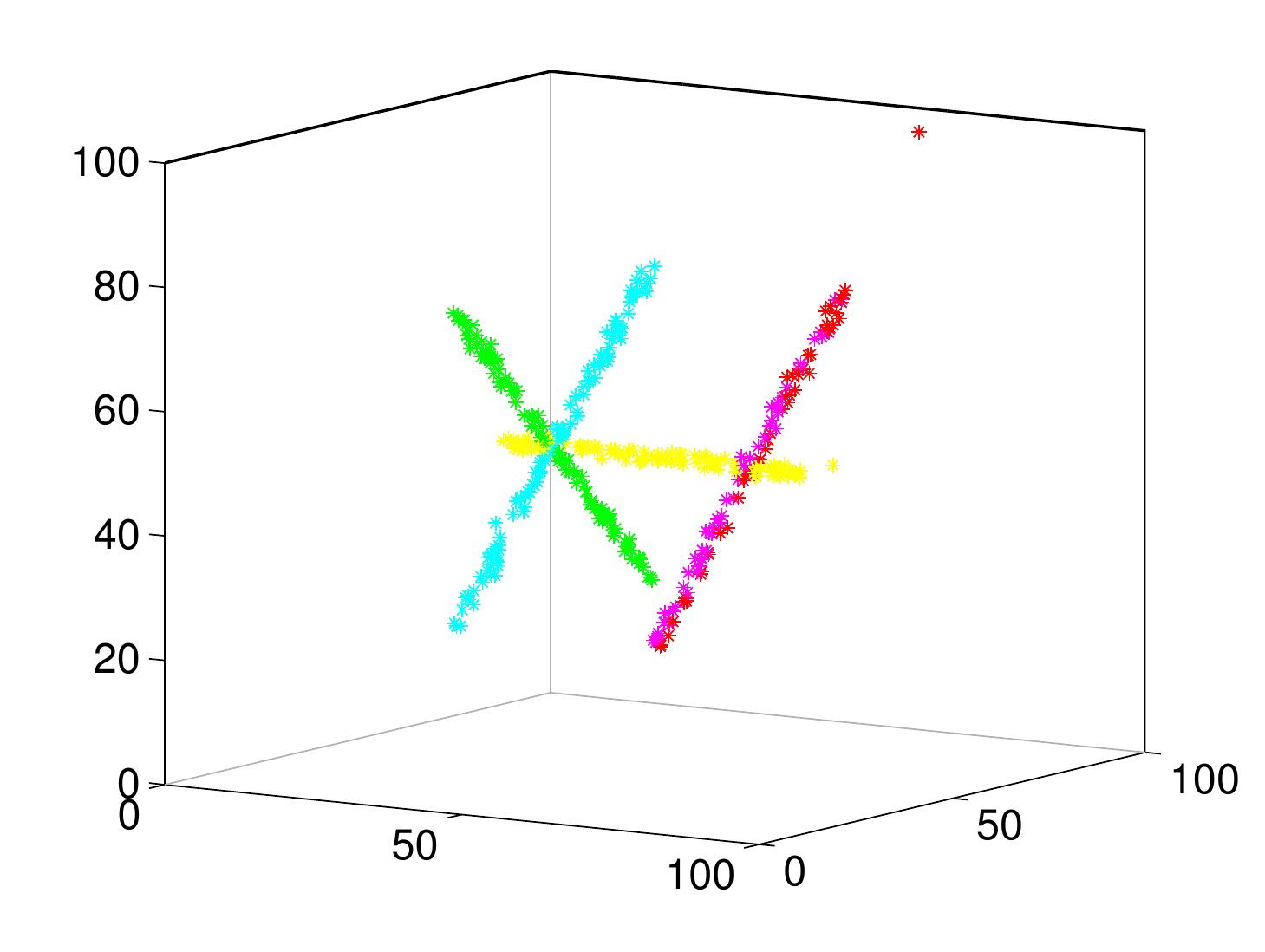}}
  \centerline{\includegraphics[width=1.0\textwidth]{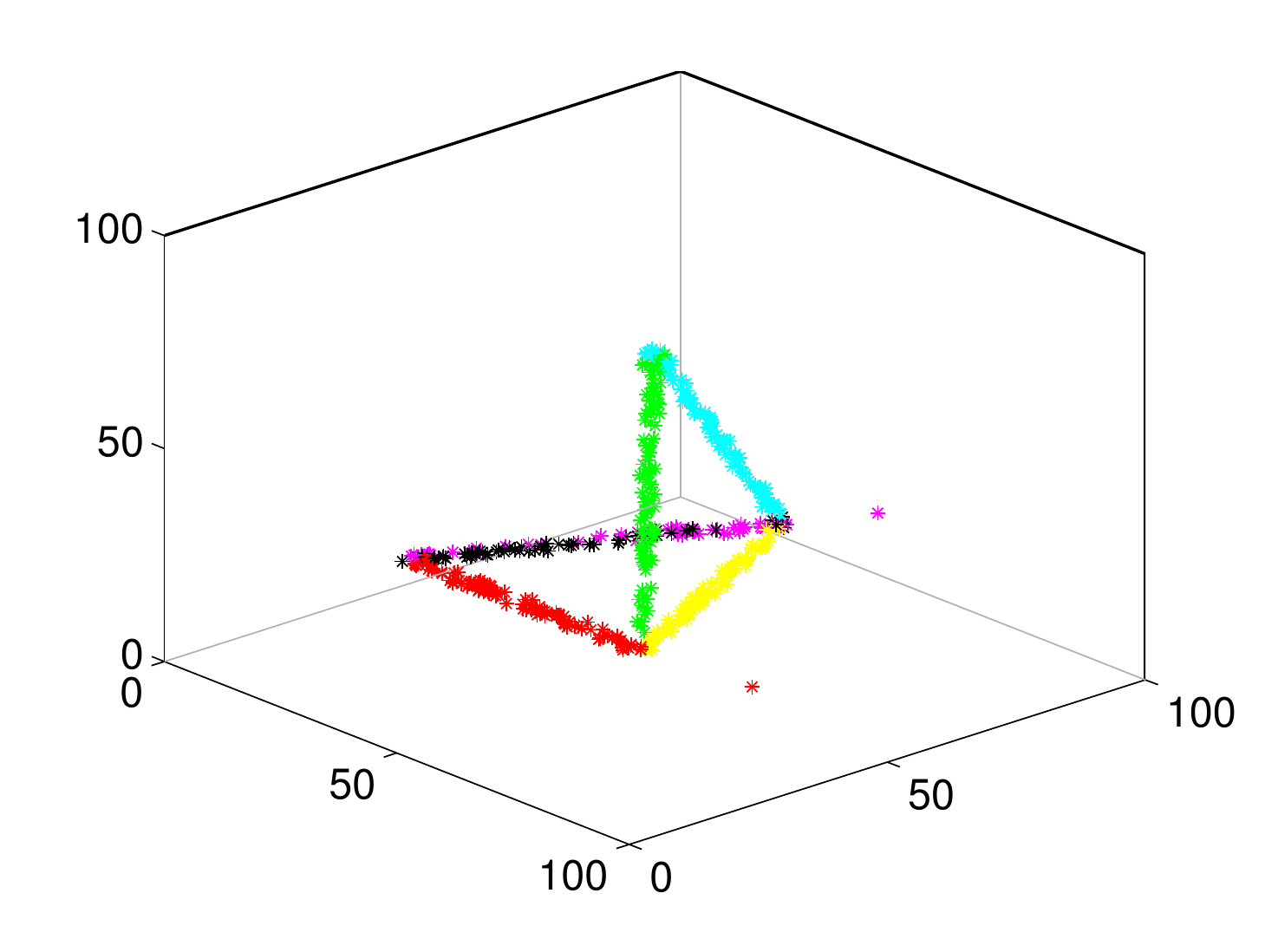}}
  \begin{center} (c) RCG  \end{center}
\end{minipage}
\begin{minipage}[t]{.15\textwidth}
\centerline{\includegraphics[width=1.0\textwidth]{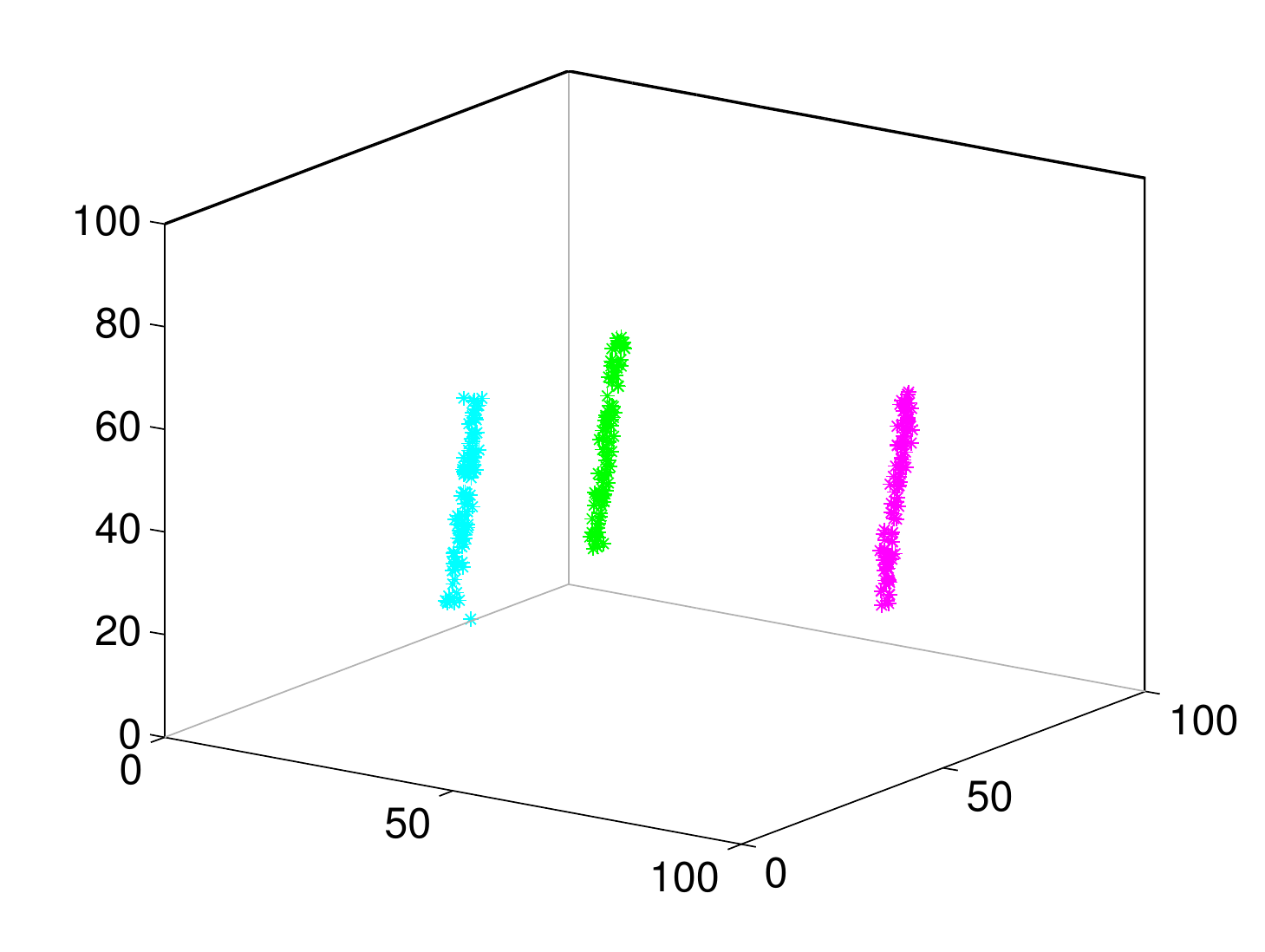}}
  \centerline{\includegraphics[width=1.0\textwidth]{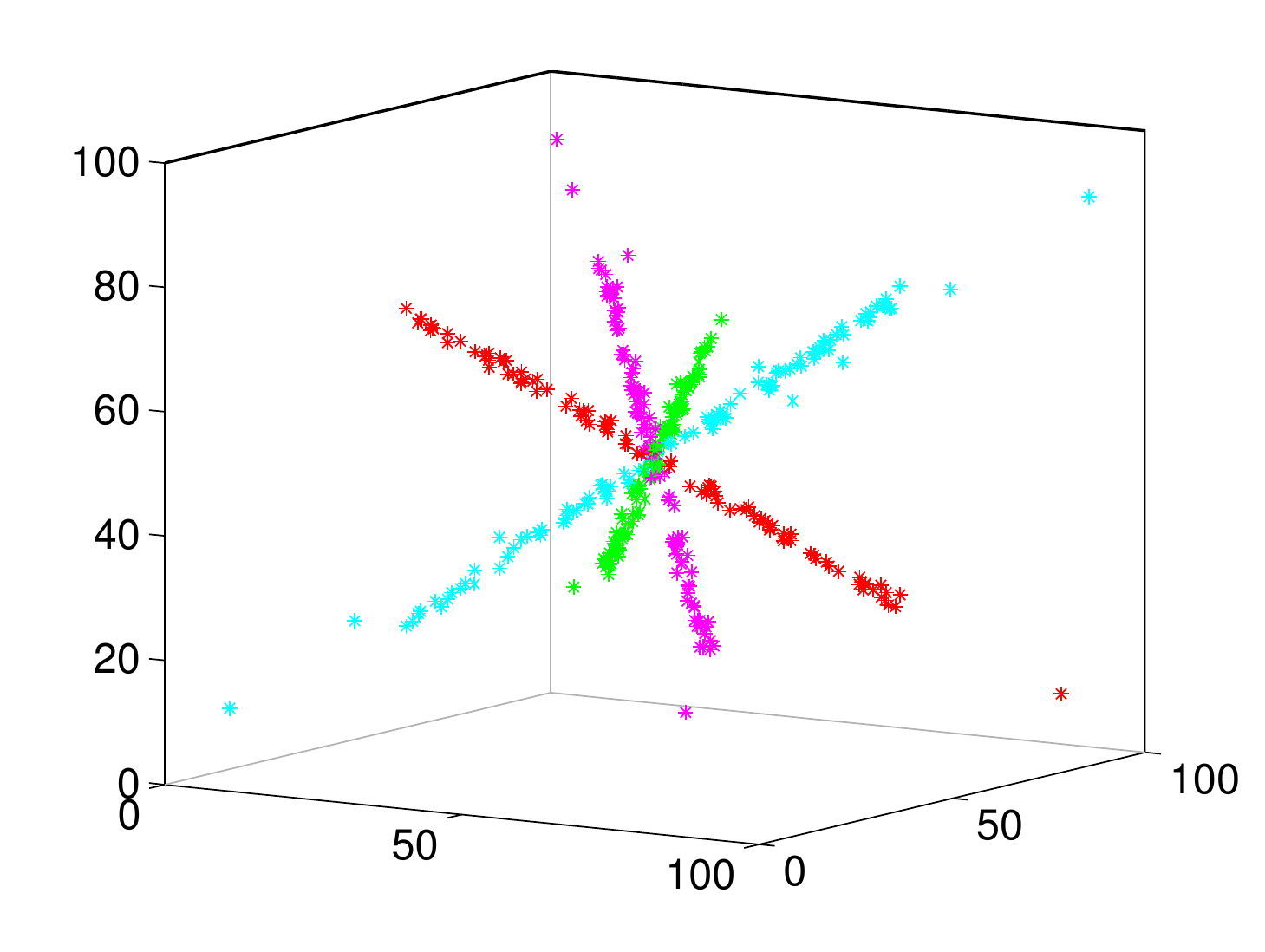}}
  \centerline{\includegraphics[width=1.0\textwidth]{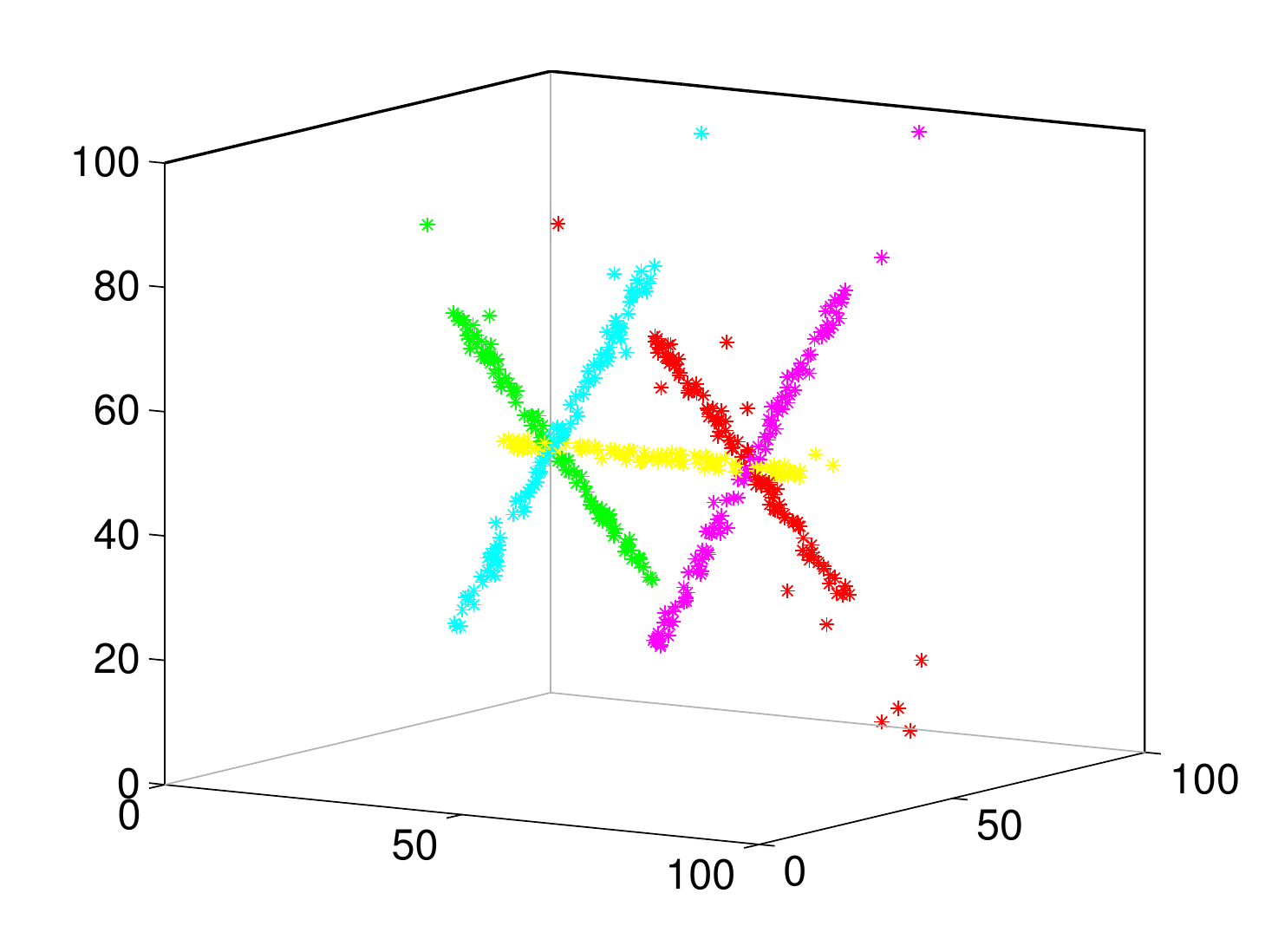}}
  \centerline{\includegraphics[width=1.0\textwidth]{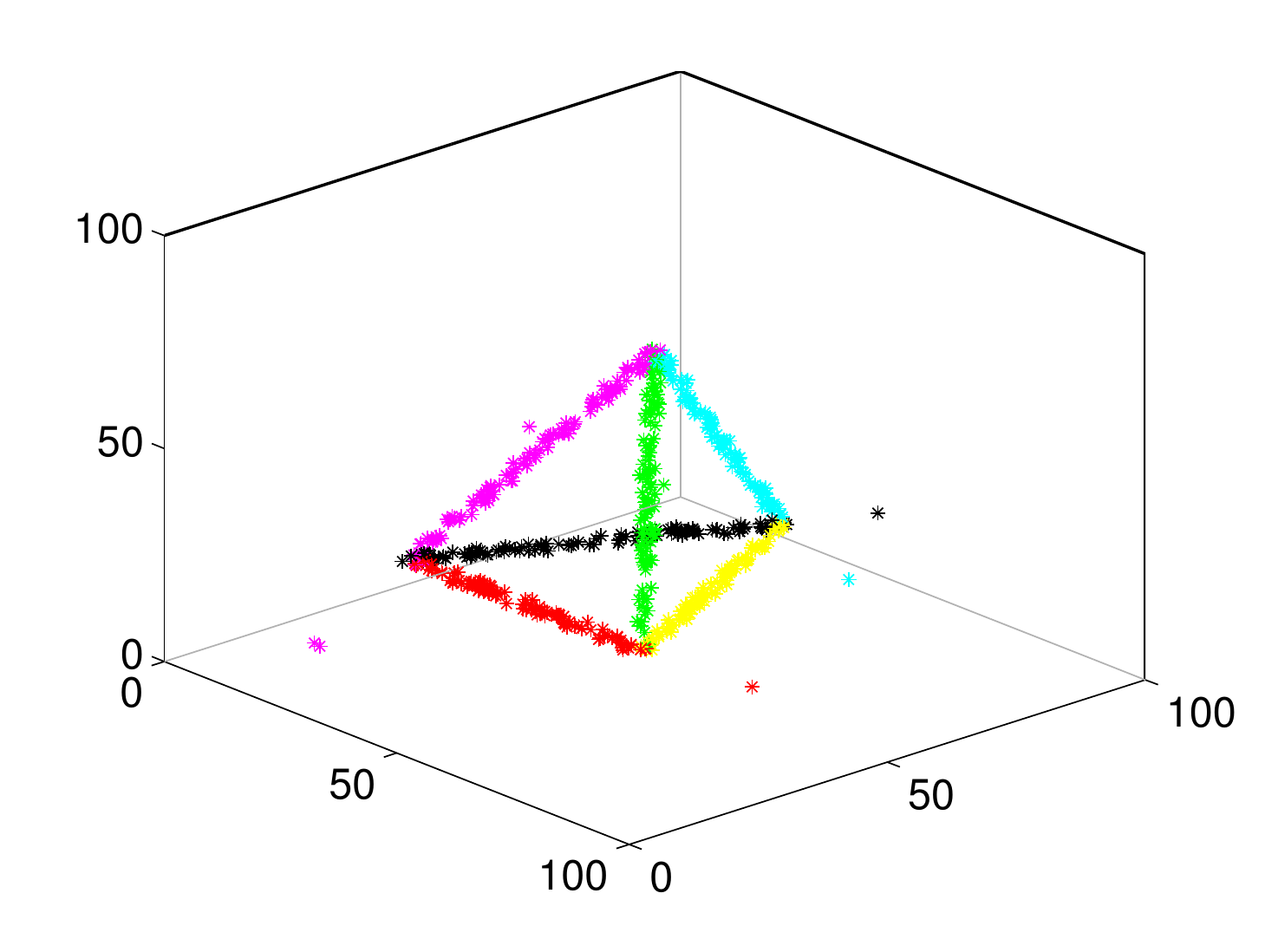}}
  \begin{center} (d) AKSWH  \end{center}
\end{minipage}
\begin{minipage}[t]{.15\textwidth}
\centerline{\includegraphics[width=1.0\textwidth]{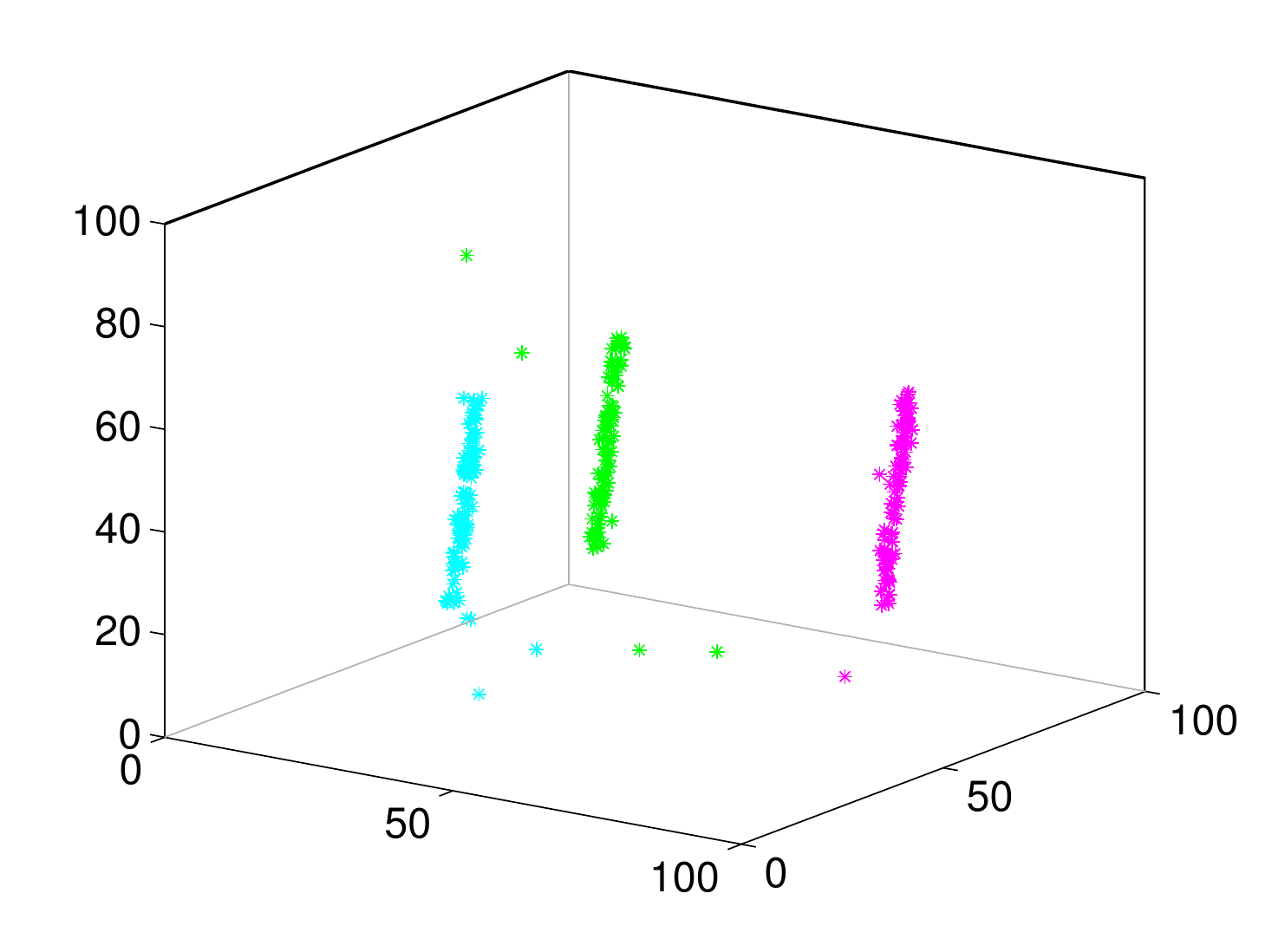}}
  \centerline{\includegraphics[width=1.0\textwidth]{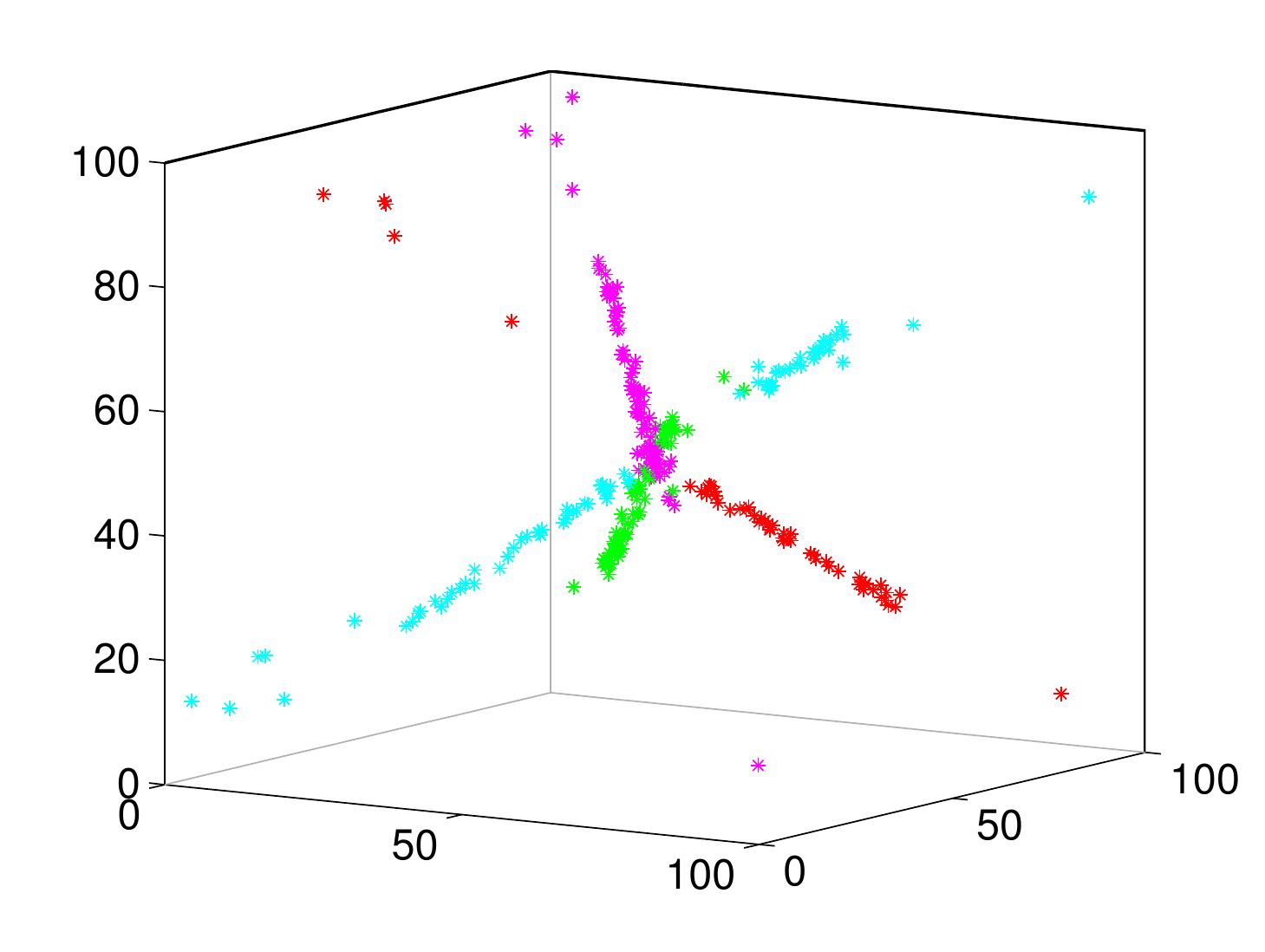}}
  \centerline{\includegraphics[width=1.0\textwidth]{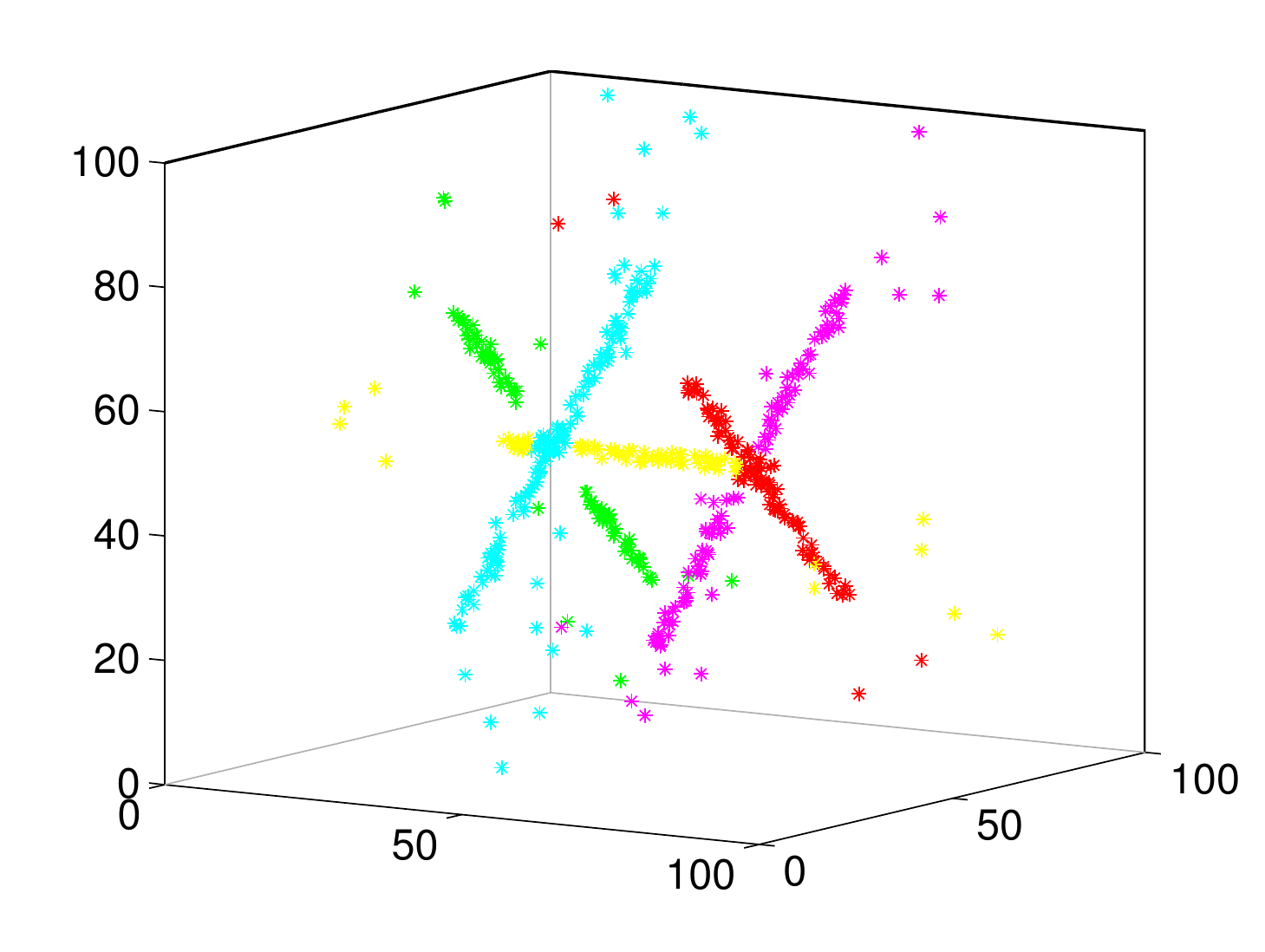}}
  \centerline{\includegraphics[width=1.0\textwidth]{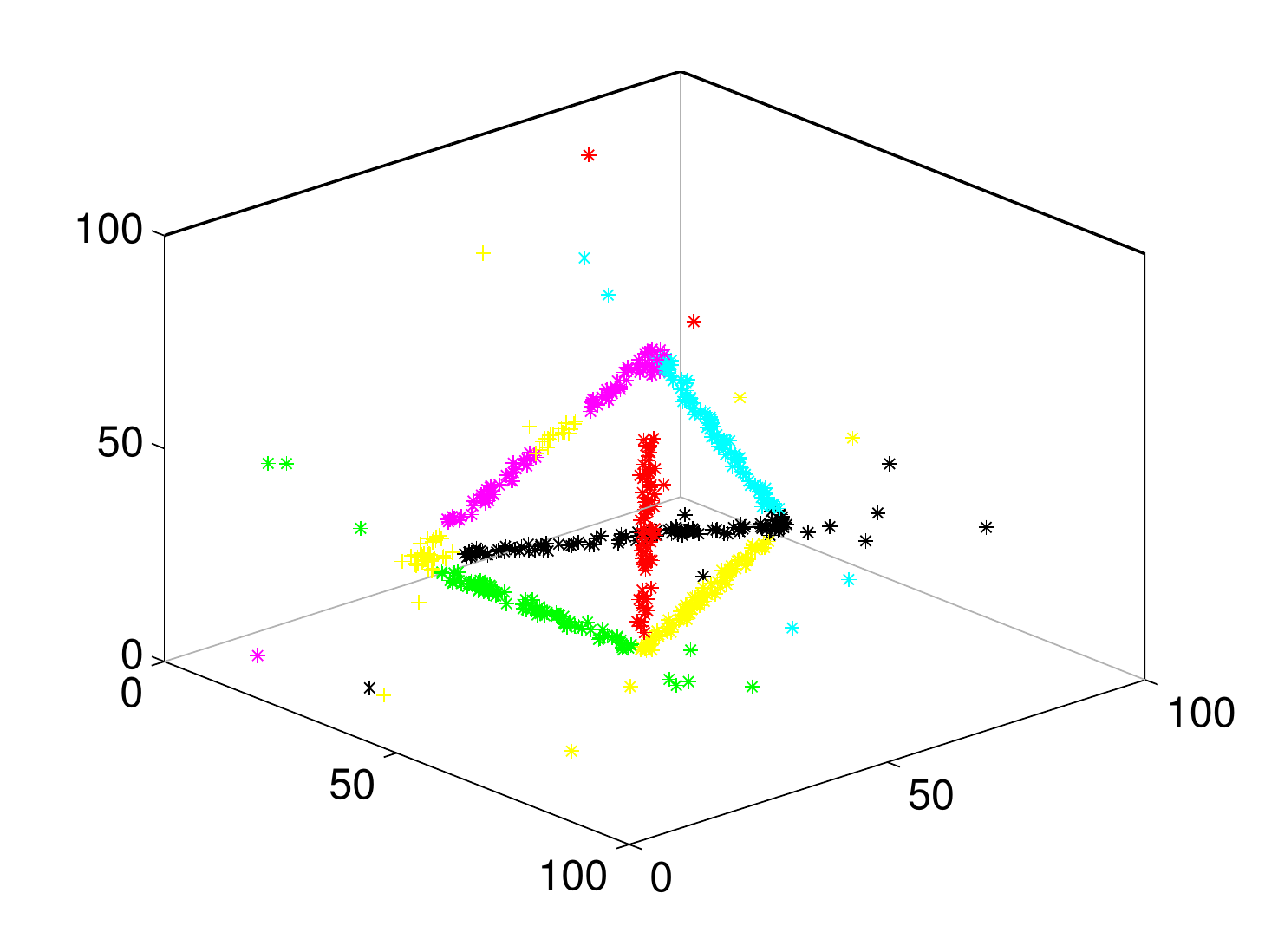}}
  \centerline{(e) T-linkage }\medskip
\end{minipage}
\begin{minipage}[t]{.15\textwidth}
  \centerline{\includegraphics[width=1.0\textwidth]{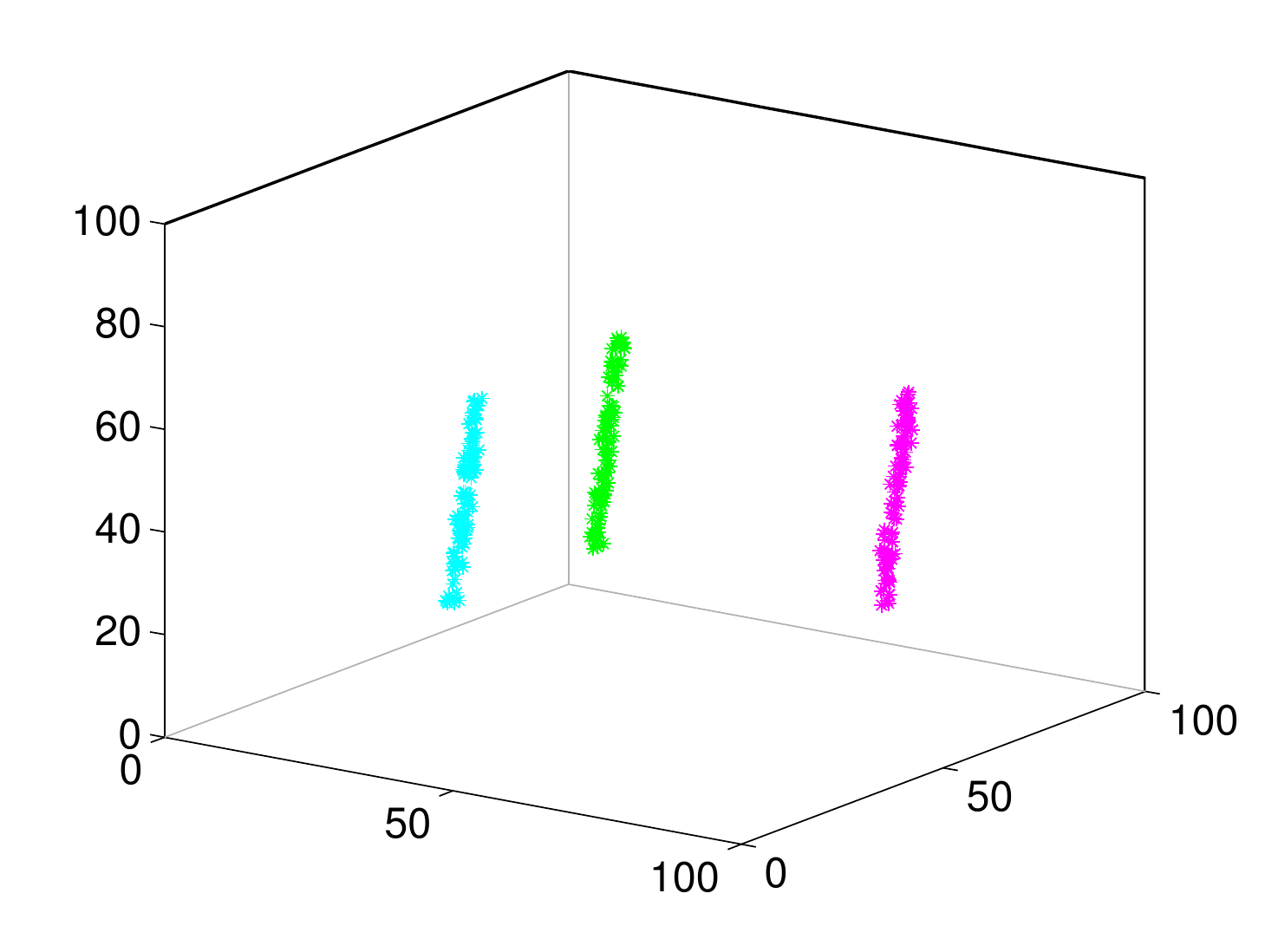}}
  \centerline{\includegraphics[width=1.0\textwidth]{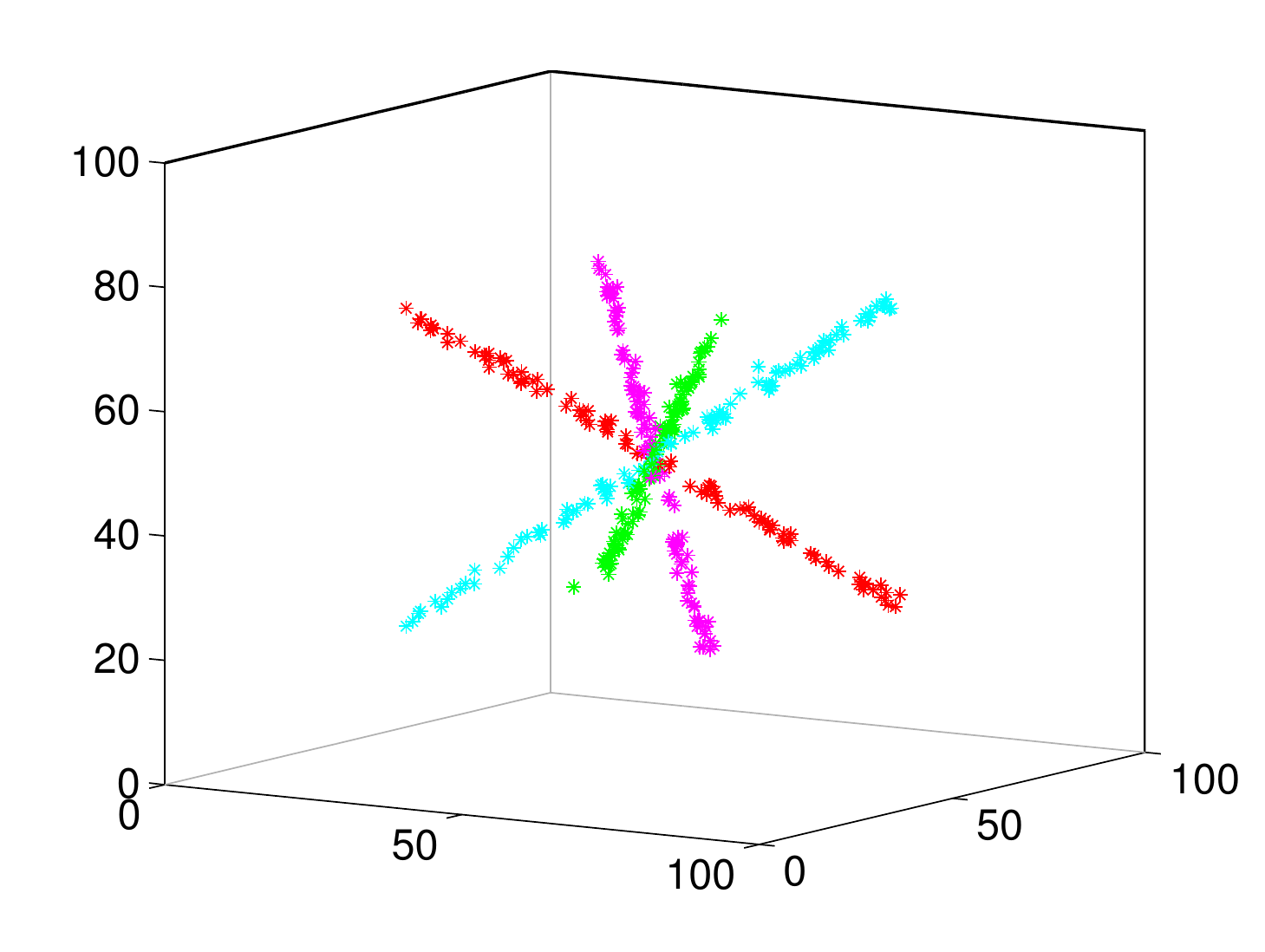}}
  \centerline{\includegraphics[width=1.0\textwidth]{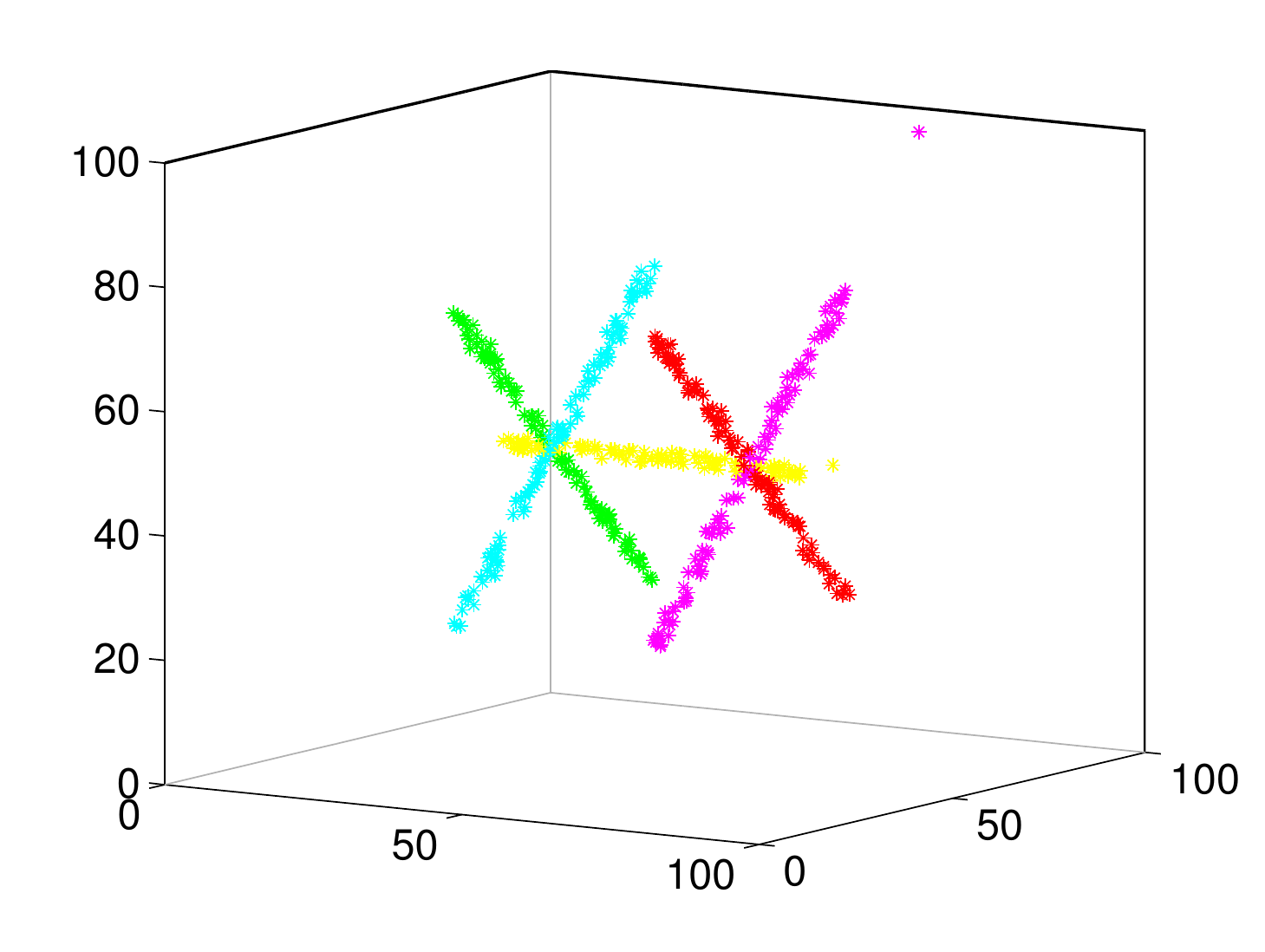}}
  \centerline{\includegraphics[width=1.0\textwidth]{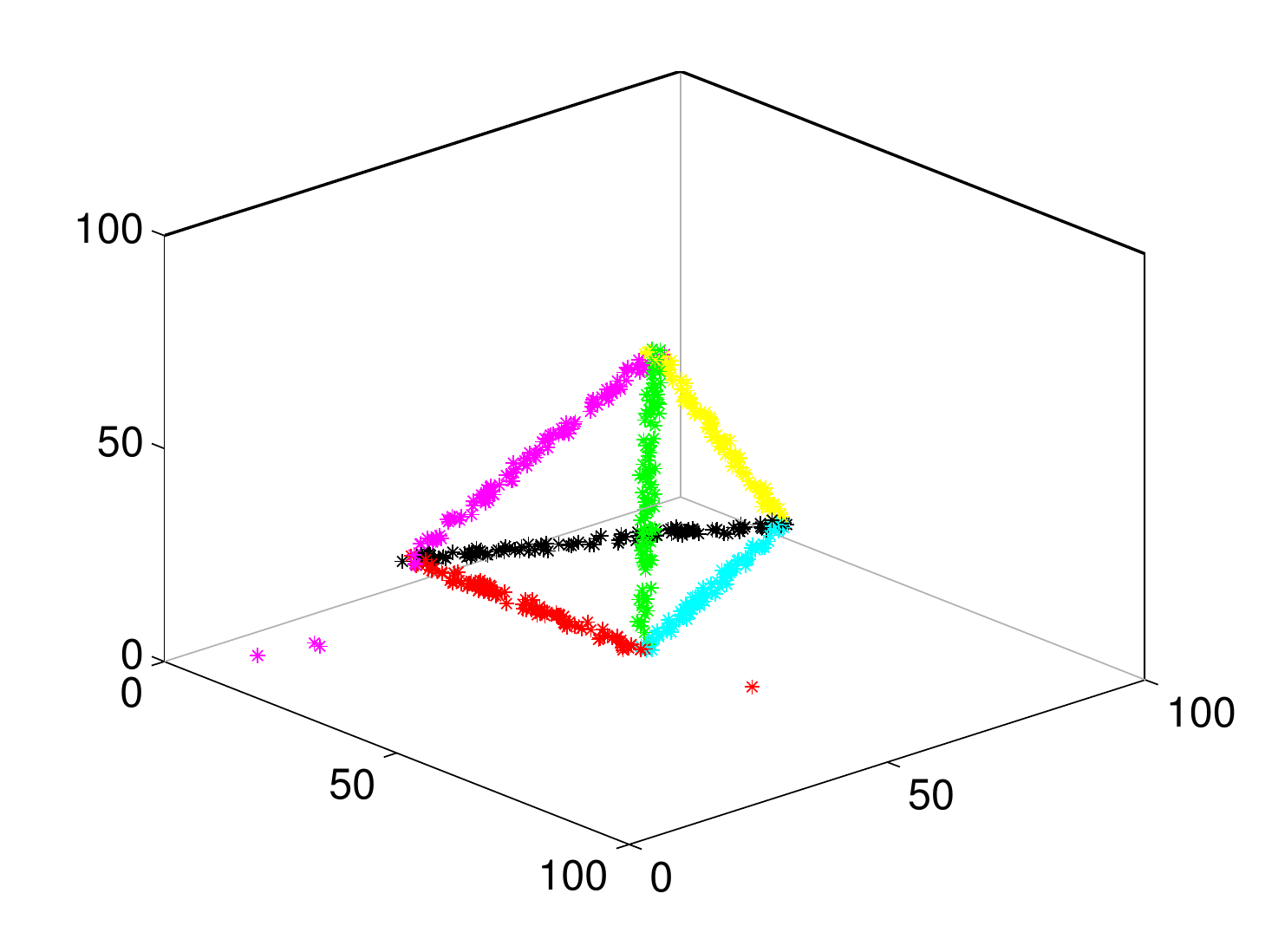}}
  \centerline{(f) MSH }\medskip
\end{minipage}
\hfill
\caption{Examples for line fitting in the $3$D space. $1^{st}$ to $4^{th}$ rows respectively fit three, four, five and six lines. The corresponding outlier percentages are respectively $86\%$, $88\%$, $89\%$ and $90\%$. The inlier scale is set to 1.0. (a) The original data with $400$ outliers. Each line includes 100 inliers. (b) to (f) The results obtained by KF, RCG, AKSWH, T-linkage and MSH, respectively.}
\label{fig:fivelines}
\end{figure*}

\begin{table*}
\small\
  \caption{The fitting errors (in percentage) for line fitting on four datasets (the best results are boldfaced)}
\medskip
\centering
\begin{tabular}{|c|c|c|c|c|c|c|c|c|c|c|c|c|}
\hline
\multirow{2}{*}{} & \multicolumn{3}{c|}{3 lines} & \multicolumn{3}{c|}{4 lines}& \multicolumn{3}{c|}{5 lines} & \multicolumn{3}{c|}{6 lines}\\
\cline{2-13}
& Std. & Avg. &  Min. & Std. & Avg. & Min. &Std. & Avg. &Min.&Std.& Avg. & Min. \\
\hline
 KF       & {0.00} & {1.76} & {1.71} & {0.03}  & {18.25} & {13.25} & {0.03} & {15.27} & {11.42} & {0.03} & {33.71} & {27.10} \\
 \hline
 RCG      & {0.00} & {0.33} & {0.29} & {0.02} & {4.13} & {1.63}  & {0.07} & {18.00} & {2.44} &{0.05} & {15.69}& {5.00} \\
 \hline
AKSWH     & {0.00} & {0.34} & {0.29} & {0.02}  & {3.00}  & {2.88}  & {0.05} & {3.78} & {2.67} & {0.03} & {4.57}& {2.70} \\
\hline
T-linkage & {0.00} & {1.87} & {1.71} & {0.05}  & {31.40} & {23.75} & {0.05} & {17.29} & {11.89} & {0.03} & {16.26}& {11.70} \\
\hline
MSH      & {0.00} & {\bf0.16} & {\bf0.14} & {\bf0.01}  & {\bf1.29}  & {\bf0.88}  & {\bf0.00} & {\bf1.76} & {\bf1.44} & {\bf0.01} & {\bf3.34} & {\bf2.30}  \\
\hline
\end{tabular}
 \label{table:3Dlinetable}
\end{table*}
\section{The Complete Method}
\label{sec:completealgorithm}
Based on the ingredients described in the previous sections, we present the complete fitting method in this section. We summarize the proposed Mode Seeking on Hypergraphs (MSH) method for geometric model fitting in Algorithm \ref{alg:MSH}.

The proposed MSH seeks modes by directly searching hypergraphs for authority peaks in the parameter space without requiring iterative processes. The computational complexity of MSH is mainly governed by Step $3$ for computing the T-distance between pairs of vertices. Therefore, the total complexity approximately amounts to $O(M^2)$, where $M$ is the number of sampling vertices in $G^*$ and $M$ is empirically about $10\%$ $\sim$ $20\%$ of vertices in $G$.
\begin{algorithm}[htb] 
\renewcommand{\algorithmicrequire}{\textbf{Input:}}
\renewcommand\algorithmicensure {\textbf{Output:} }
\caption{The mode seeking on hypergraphs method for geometric model fitting} 
\label{alg:MSH} 
\begin{algorithmic}[1] 
\REQUIRE 
Data points $X$, the $K$ value for IKOSE
\STATE Construct a hypergraph $G$ and compute the weighting score for each vertex (described in Sec.~\ref{sec:HypergraphModelling}).
\STATE Sample the vertices in $G$ by WAS to generate a new hypergraph $G^*$ (described in Sec.~\ref{sec:authorityaware}).
\STATE Compute the minimum T-distance $\eta^v_{min}$ for each sampled vertex ${v}$ by Eq.~(\ref{equ:minimumtdistance}).
\STATE Sort the vertices in $G^*$ according to their MTD values satisfying $\eta^{v_1}_{min}\geq \eta^{v_2}_{min} \geq \cdots$.
\STATE Find the vertex $v_i$ whose MTD value ($\eta^{v_i}_{min}$) has the largest drop from $\eta^{v_i}_{min}$ to $\eta^{v_i+1}_{min}$. Then reject the vertices whose values of $\eta^v_{min}$ are smaller than $\eta^{v_i}_{min}$.
\STATE Derive the inliers/outliers dichotomy from the hypergraph $G^*$ and the remaining vertices (modes).
\ENSURE The modes (model instances) and the hyperedges (inliers) connected by the modes.
\end{algorithmic}
\end{algorithm}
\section{Experiments}
\label{sec:experiments}
In this section, we compare the proposed MSH with several state-of-the-art model fitting methods, including KF~\cite{chin2009robust}, RCG~\cite{liu2012efficient}, AKSWH~\cite{wang2012simultaneously}, and T-linkage~\cite{Magri_2014_CVPR}, on both synthetic data and real images. We choose these representative methods because KF is a data clustering based method, RCG is a hypergraph based method, and AKSWH is a parameter space based method. These fitting methods are related to the proposed method (recall that MSH seeks modes on hypergraphs and fits multi-structure data in the parameter space). In addition, we also choose T-linkage due to its good performance.

To be fair, we first generate a set of model hypotheses by using the proximity sampling~\cite{kanazawa2004detection,toldo2008robust} for all the competing algorithms in each experiment.  Then the competing methods perform model fitting based on the same set of model hypotheses. We generate a number of model hypotheses as~\cite{wang2012simultaneously}, i.e., there are $5,000$ model hypotheses generated for line fitting (Sec.~\ref{sec:SyntheticData} and Sec.~\ref{sec:linefitting}) and circle fitting (Sec.~\ref{sec:circlefitting}), $10,000$ model hypotheses generated for homography based segmentation (Sec.~\ref{sec:homographbasedsegmentation}), and $20,000$ model hypotheses generated for two-view based motion segmentation (Sec.~\ref{sec:motionsegmentation}). We have optimized the parameters of all the competing fitting methods on each dataset for the best performance. The fitting error is computed as~\cite{Magri_2014_CVPR,mittal2012generalized}.

\begin{figure*}
\centering
\begin{minipage}[t]{.16\textwidth}
  \centering
  \centerline{\includegraphics[width=0.90\textwidth]{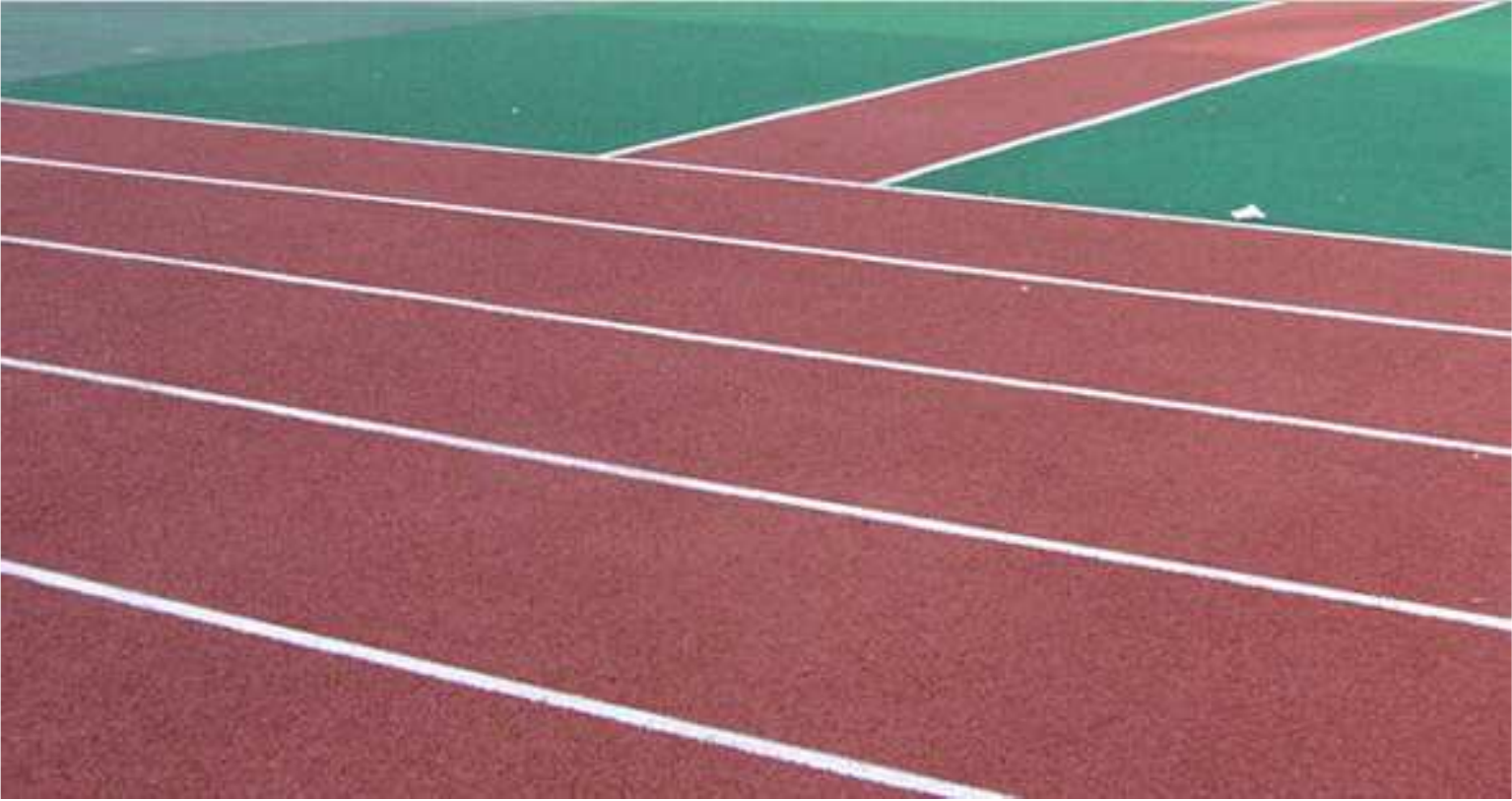}}
  \centerline{\includegraphics[width=0.90\textwidth]{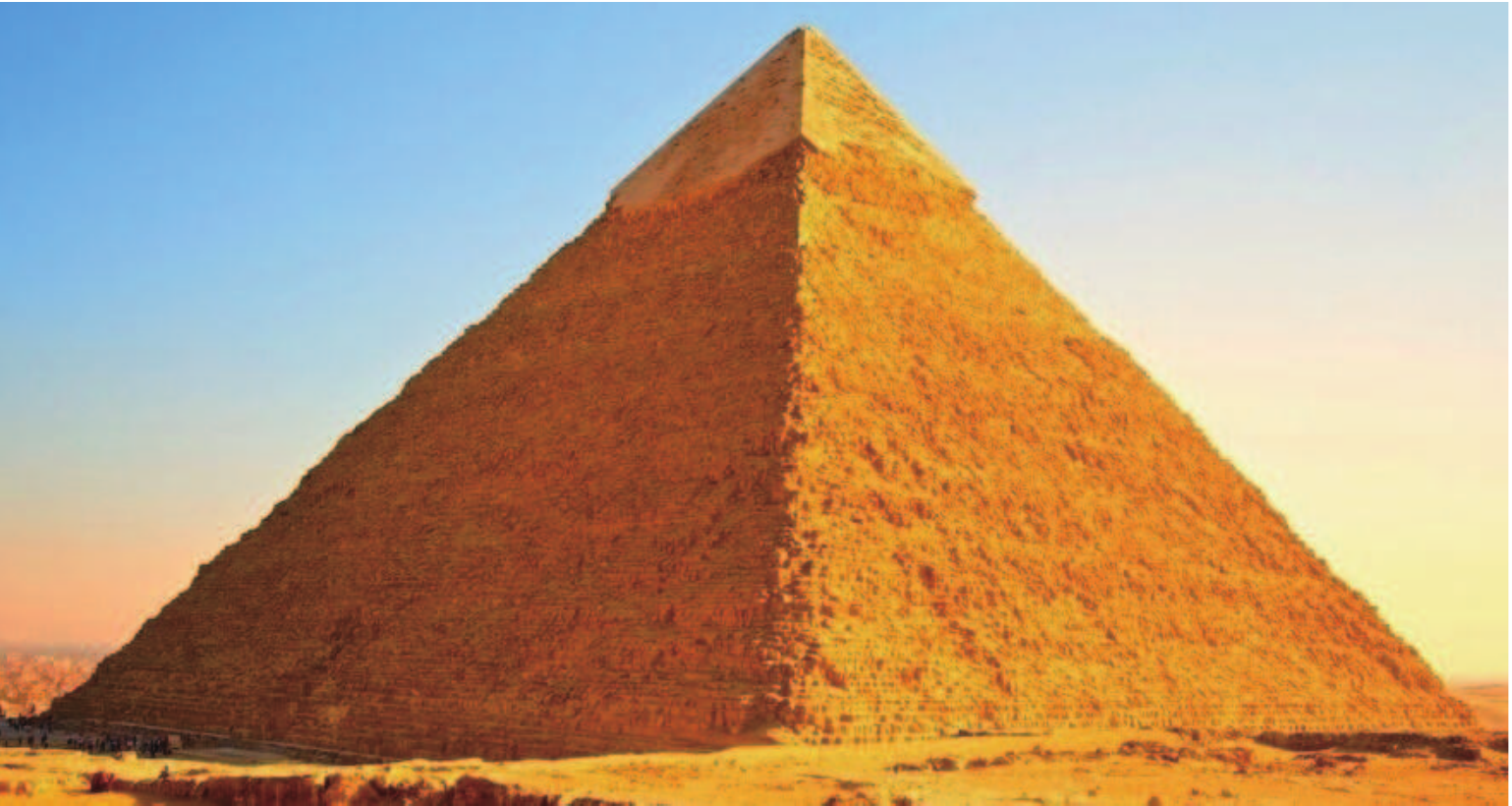}}
  \begin{center} (a) Datasets  \end{center}
\end{minipage}
\begin{minipage}[t]{.16\textwidth}
  \centering
  \centerline{\includegraphics[width=0.90\textwidth]{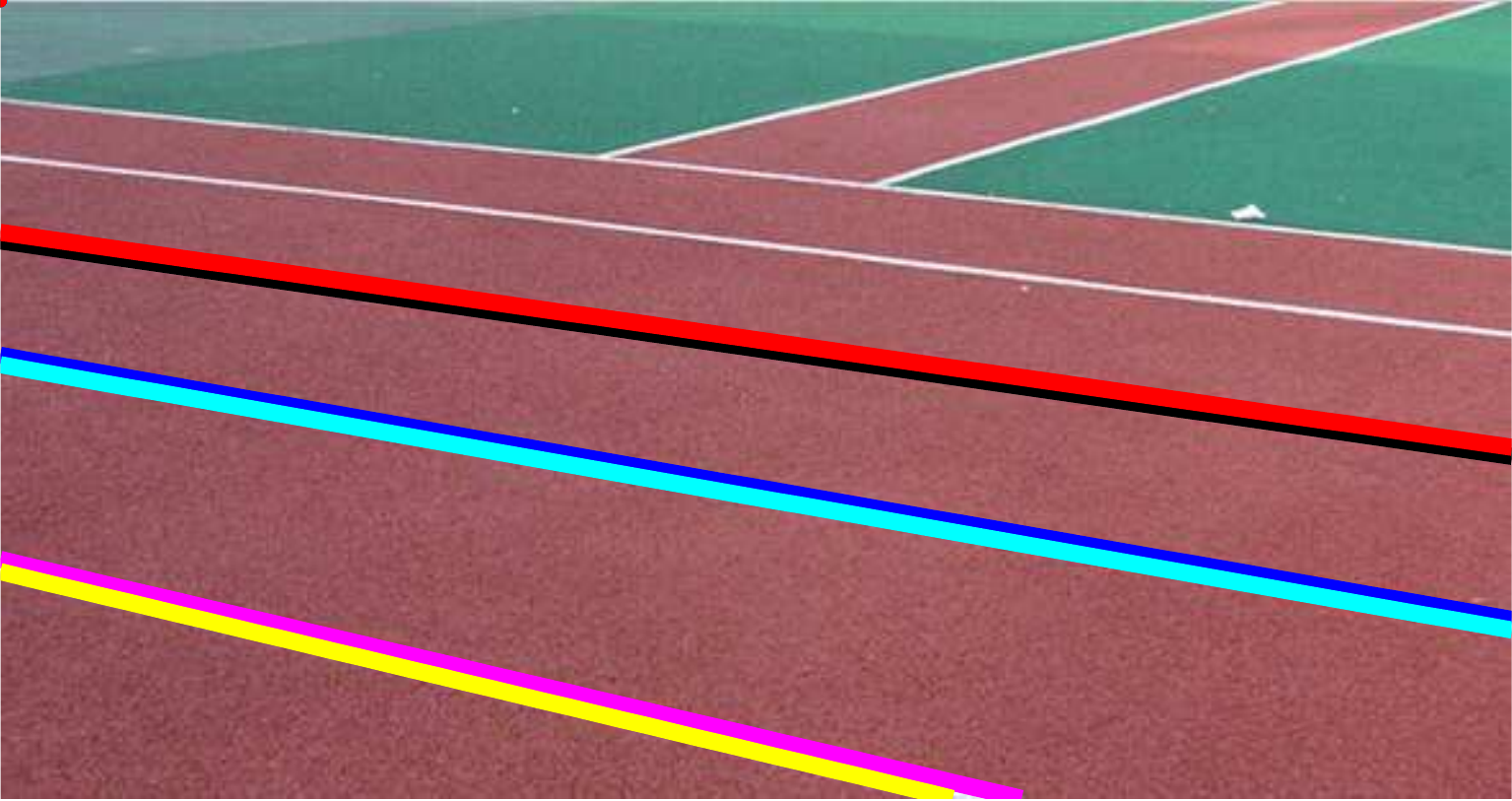}}
  \centerline{\includegraphics[width=0.90\textwidth]{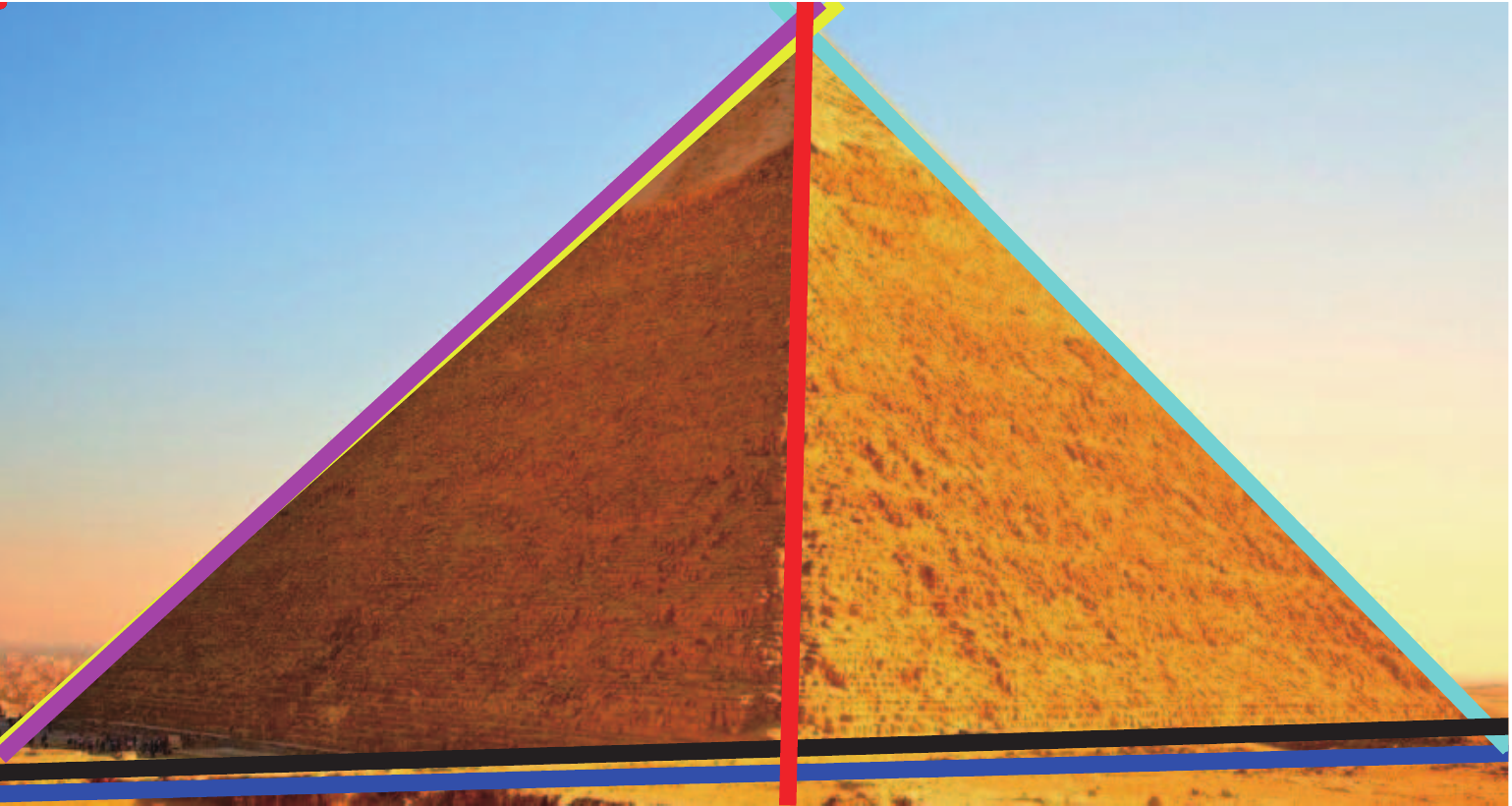}}
  \begin{center} (b) KF  \end{center}
\end{minipage}
\begin{minipage}[t]{.16\textwidth}
  \centering
  \centerline{\includegraphics[width=0.90\textwidth]{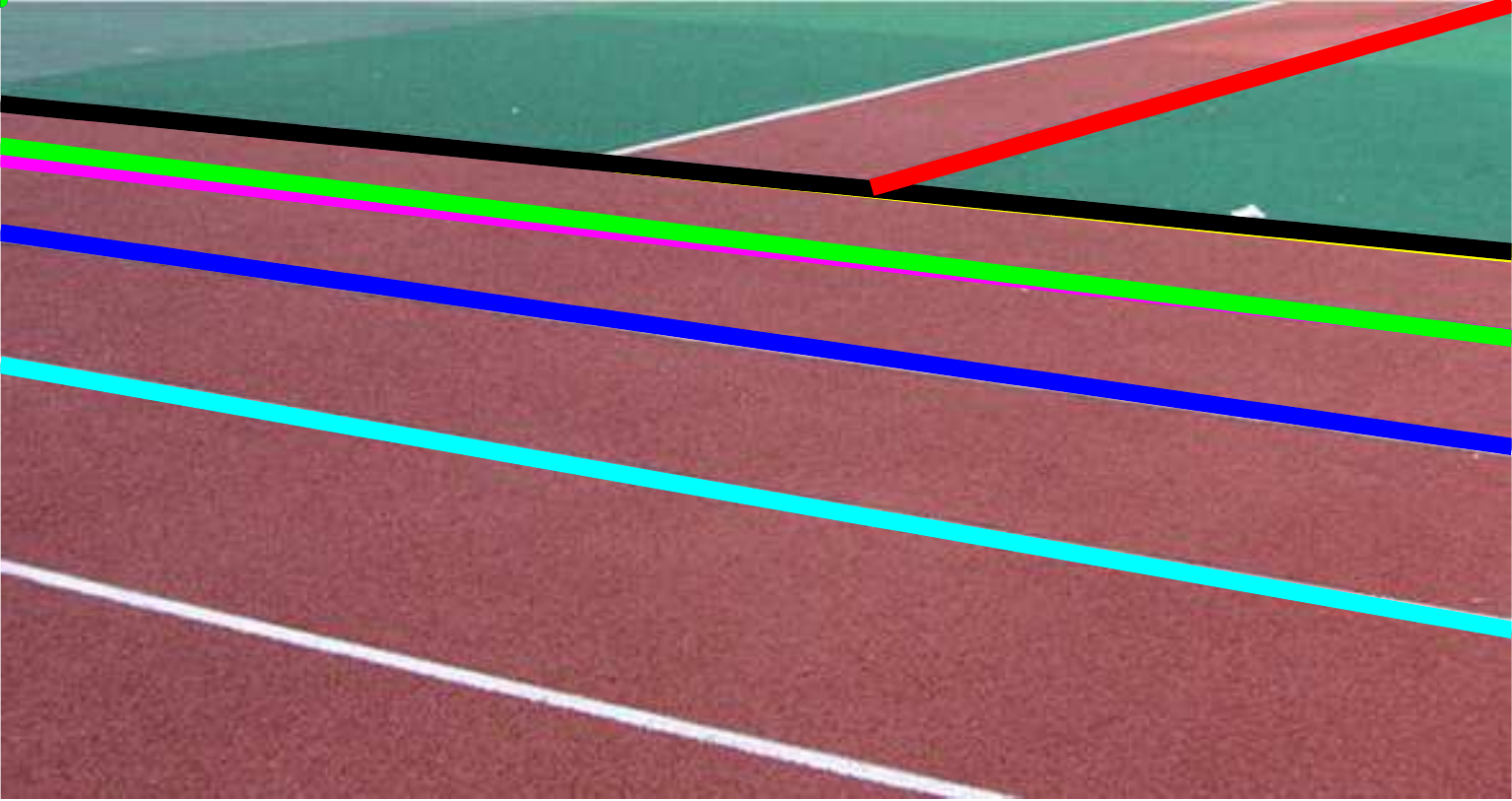}}
  \centerline{\includegraphics[width=0.90\textwidth]{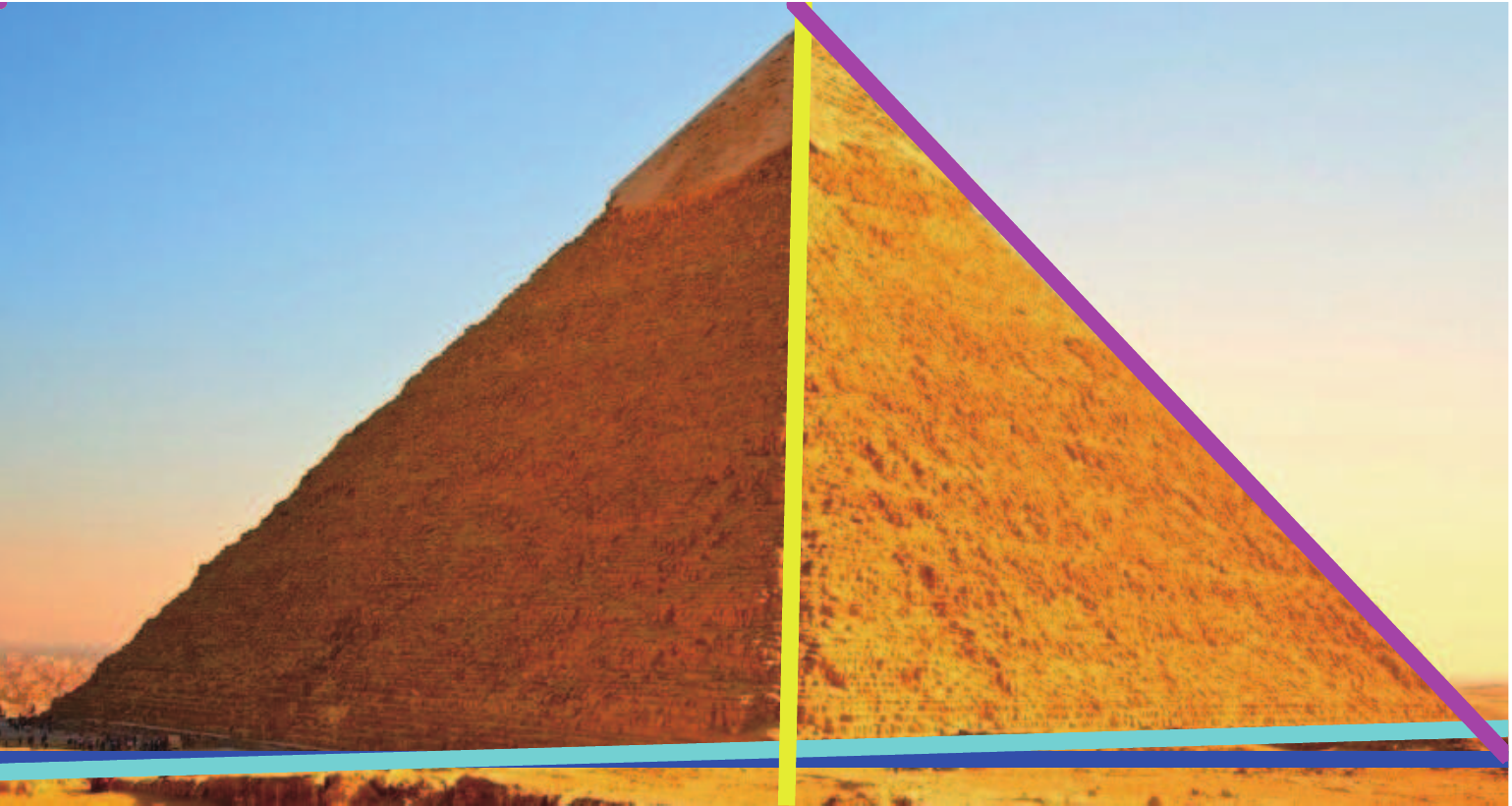}}
  \begin{center} (c) RCG  \end{center}
\end{minipage}
\begin{minipage}[t]{.16\textwidth}
  \centering
  \centerline{\includegraphics[width=0.90\textwidth]{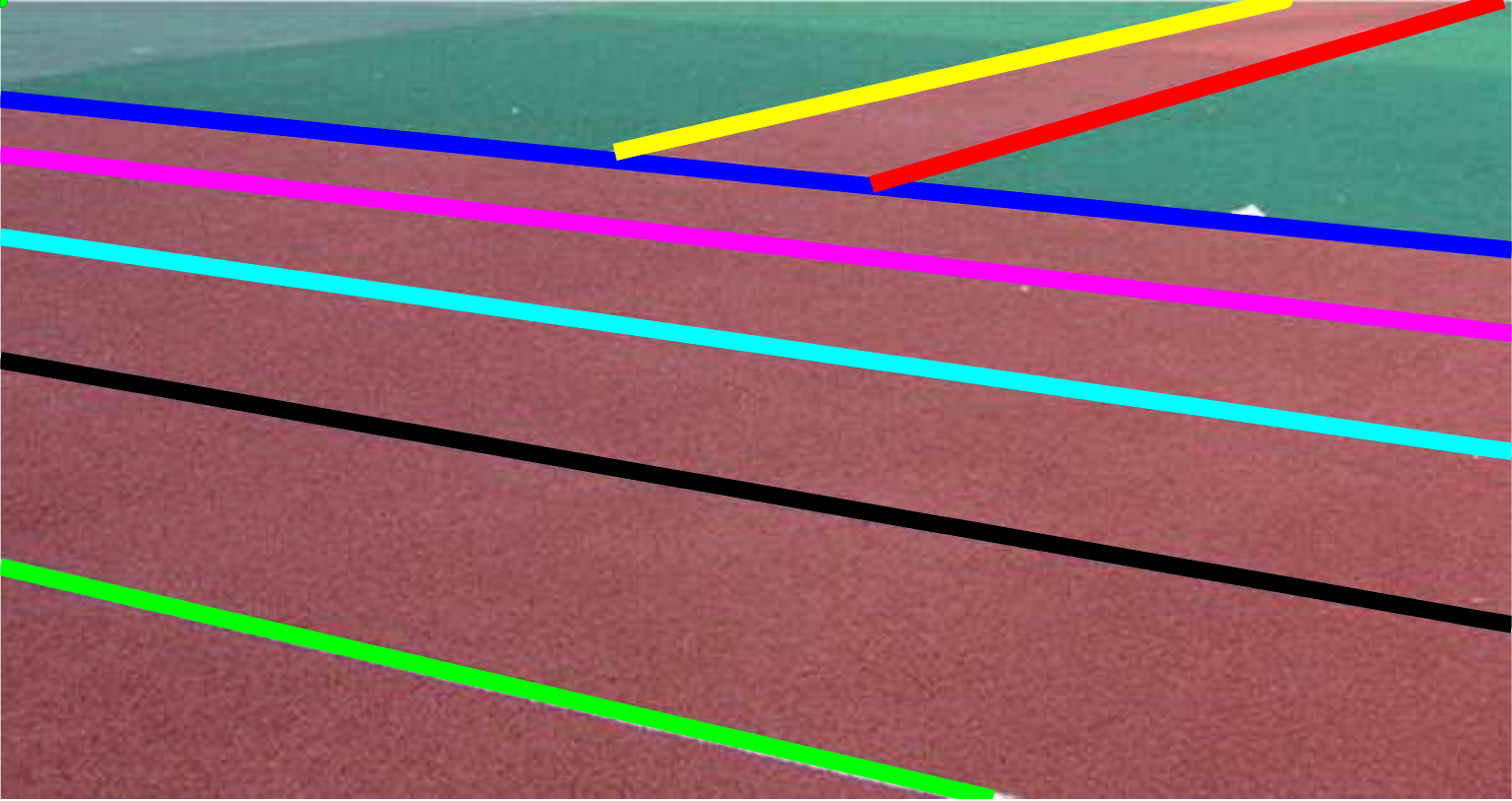}}
  \centerline{\includegraphics[width=0.90\textwidth]{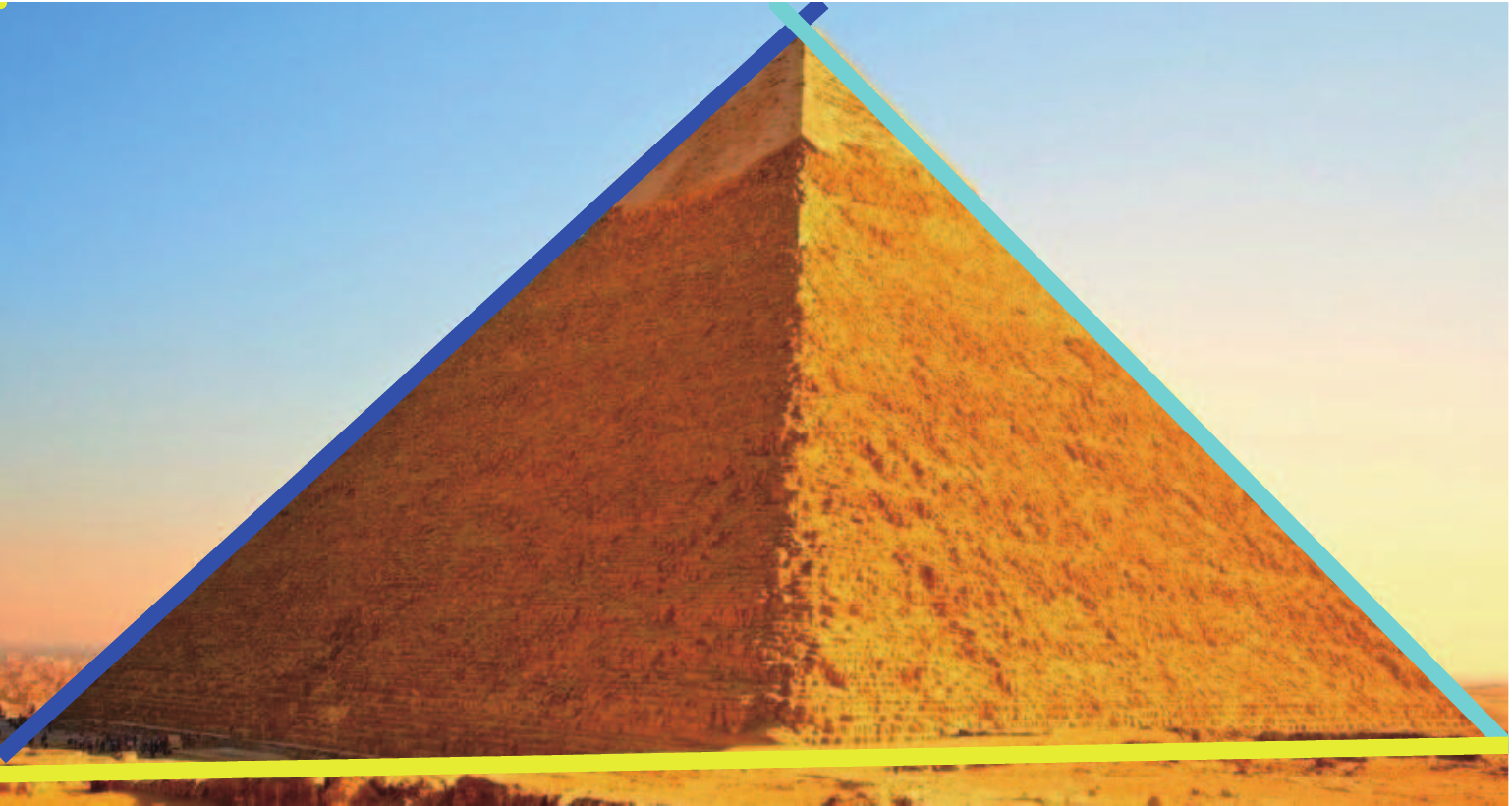}}
  \centerline{(d) AKSWH }\medskip
\end{minipage}
\begin{minipage}[t]{.16\textwidth}
  \centering
  \centerline{\includegraphics[width=0.90\textwidth]{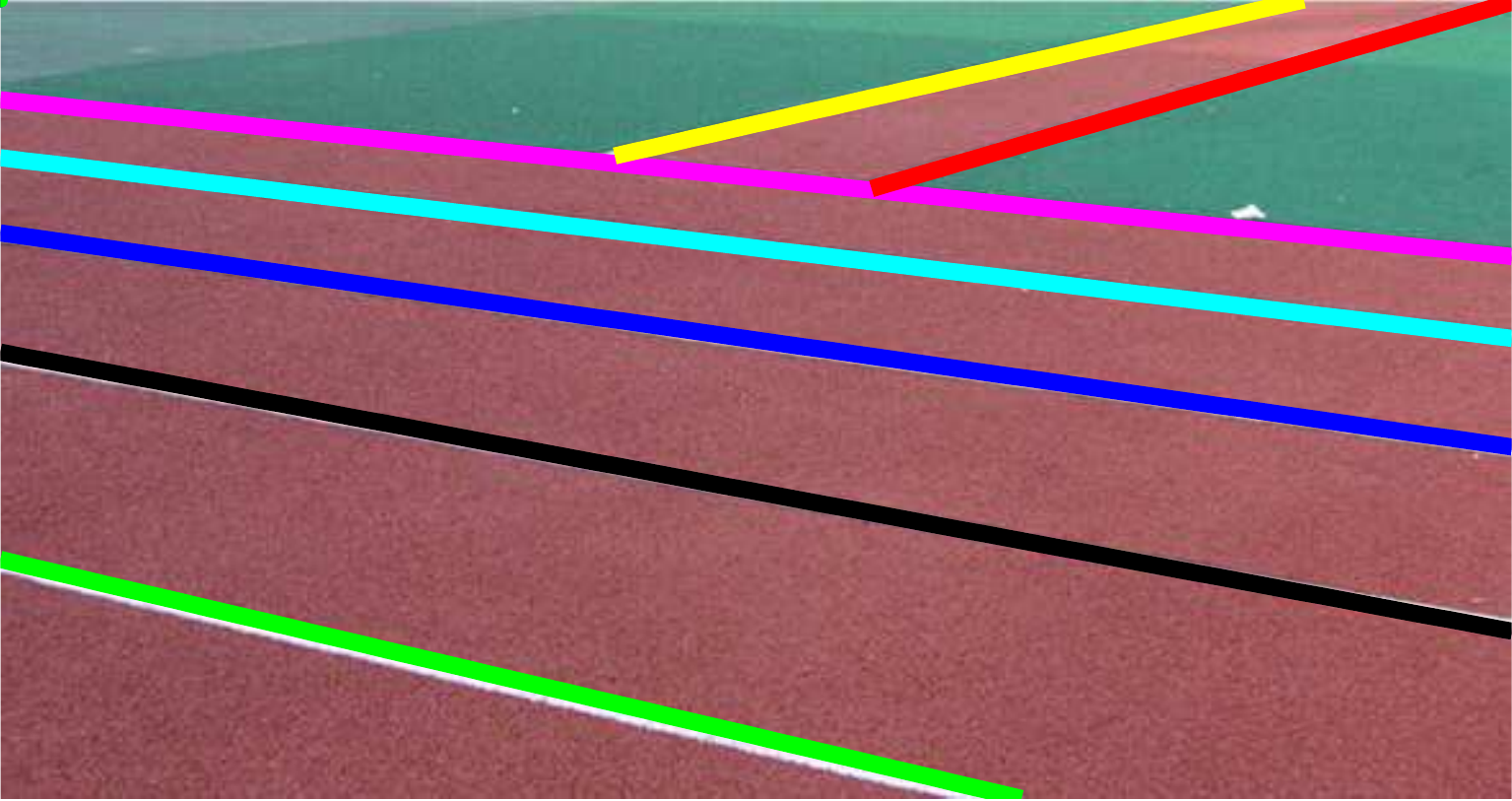}}
  \centerline{\includegraphics[width=0.90\textwidth]{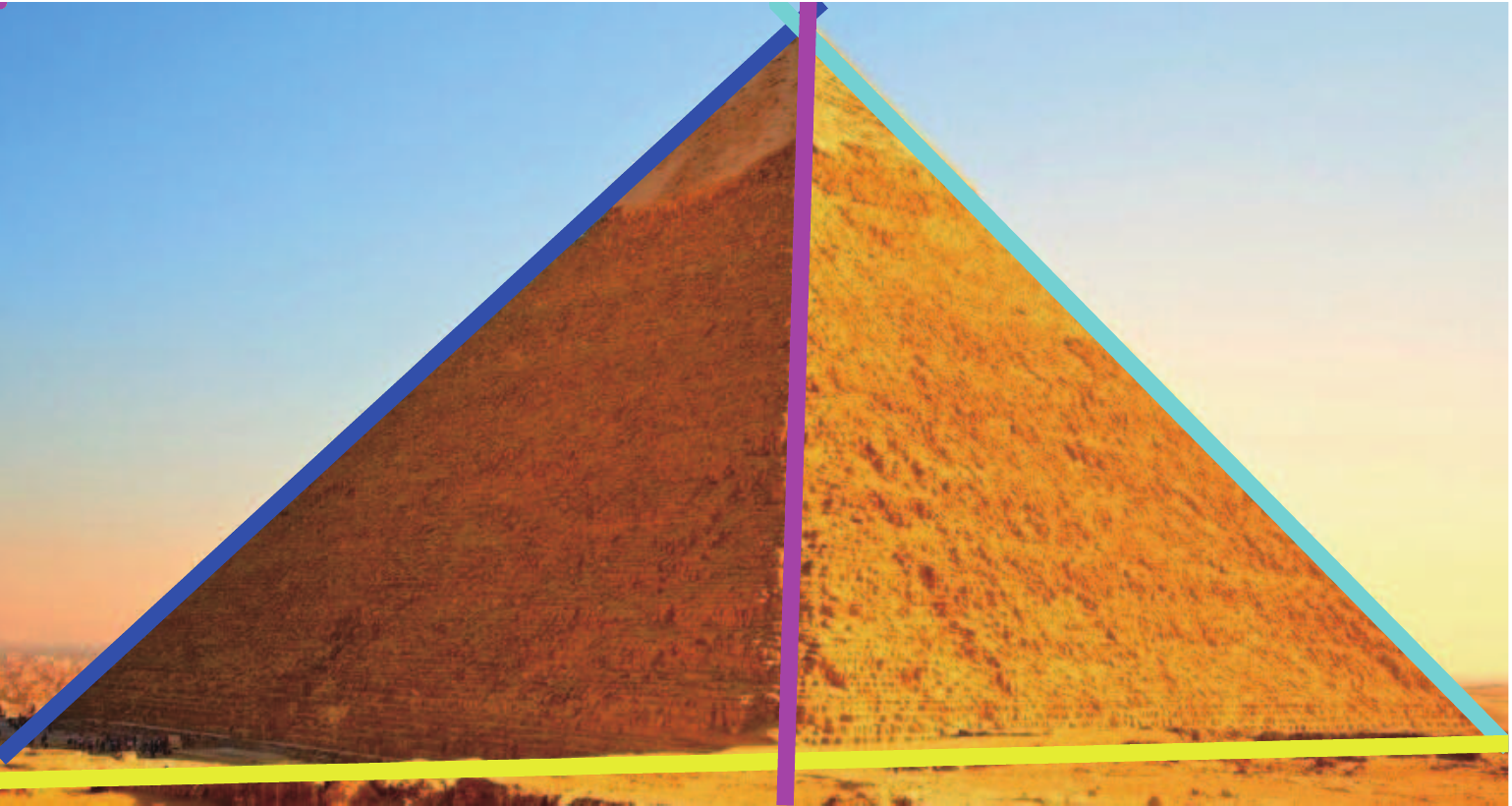}}
  \centerline{(e) T-linkage }\medskip
\end{minipage}
\begin{minipage}[t]{.16\textwidth}
  \centering
  \centerline{\includegraphics[width=0.90\textwidth]{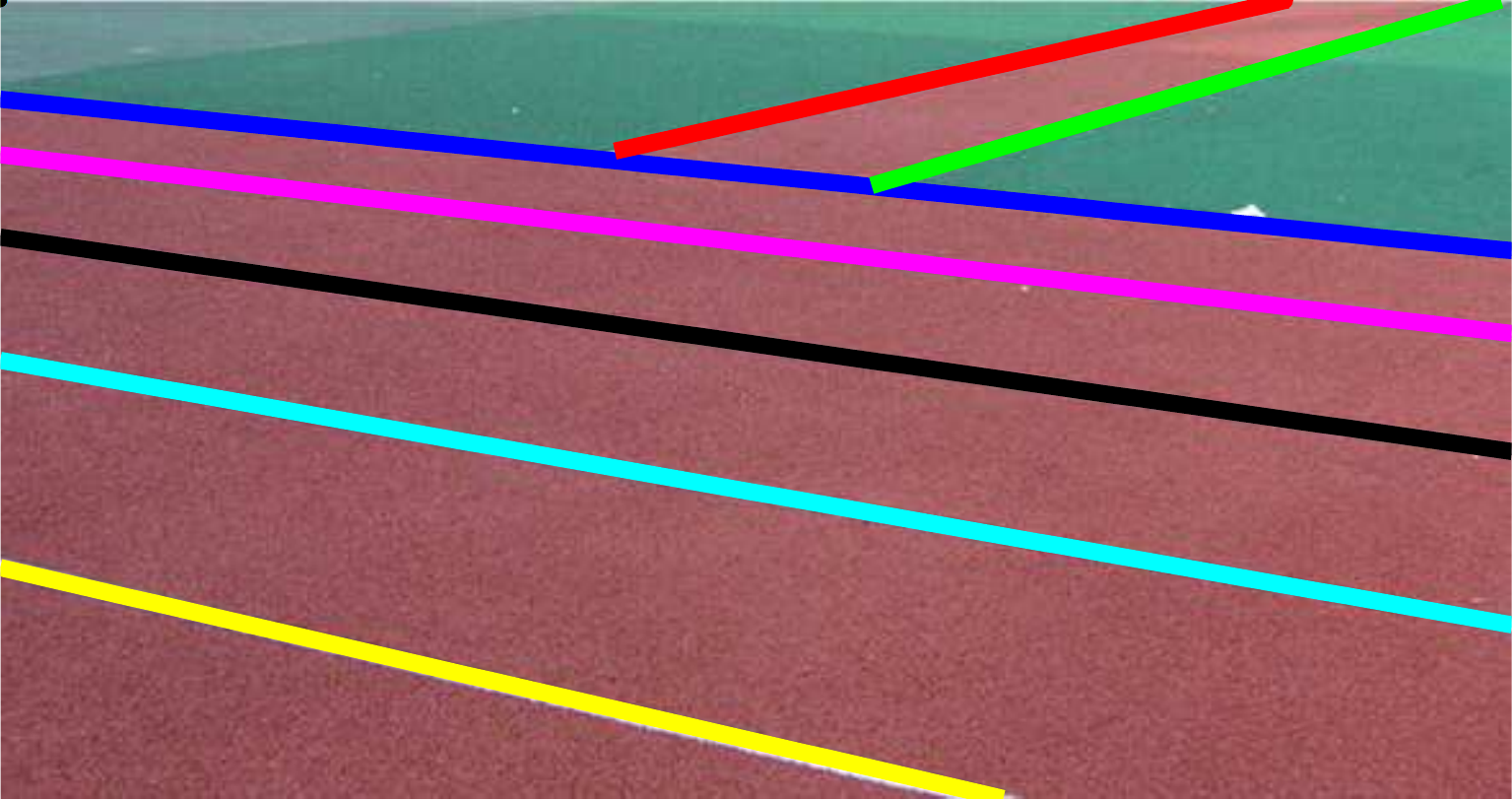}}
  \centerline{\includegraphics[width=0.90\textwidth]{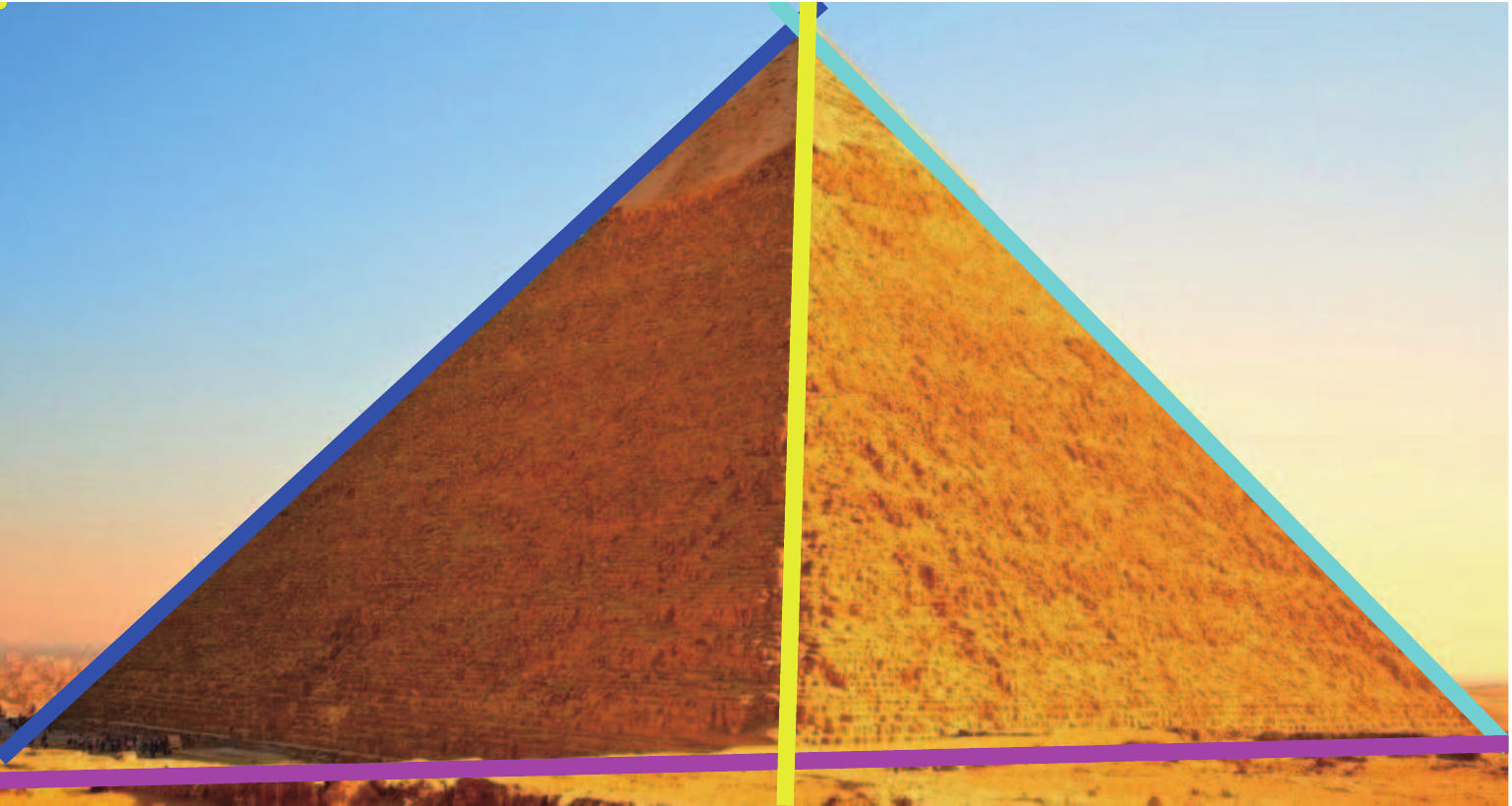}}
  \centerline{(f) MSH }\medskip
\end{minipage}
\hfill
\caption{Examples for line fitting. First (``tracks") and second (``pyramid") rows respectively fit seven and four lines. (a) The original data. (b) to (f) The results obtained by KF, RCG, AKSWH, T-linkage and MSH, respectively.}
\label{fig:linefiting}
\end{figure*}
\begin{figure*}
\centering
\begin{minipage}[t]{.16\textwidth}
  \centering
  \centerline{\includegraphics[width=0.92\textwidth]{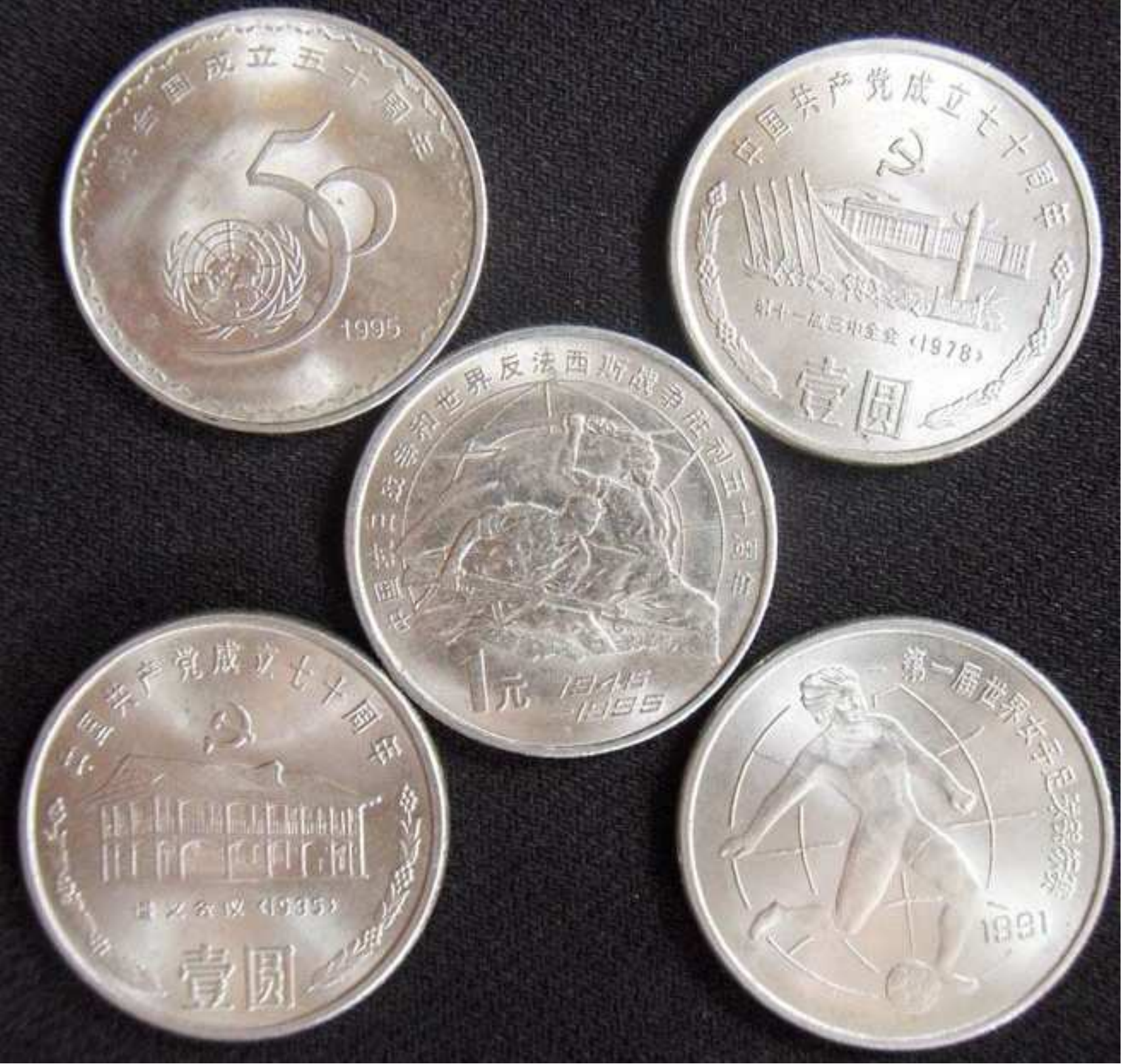}}
  \centerline{}
  \centerline{\includegraphics[width=0.92\textwidth]{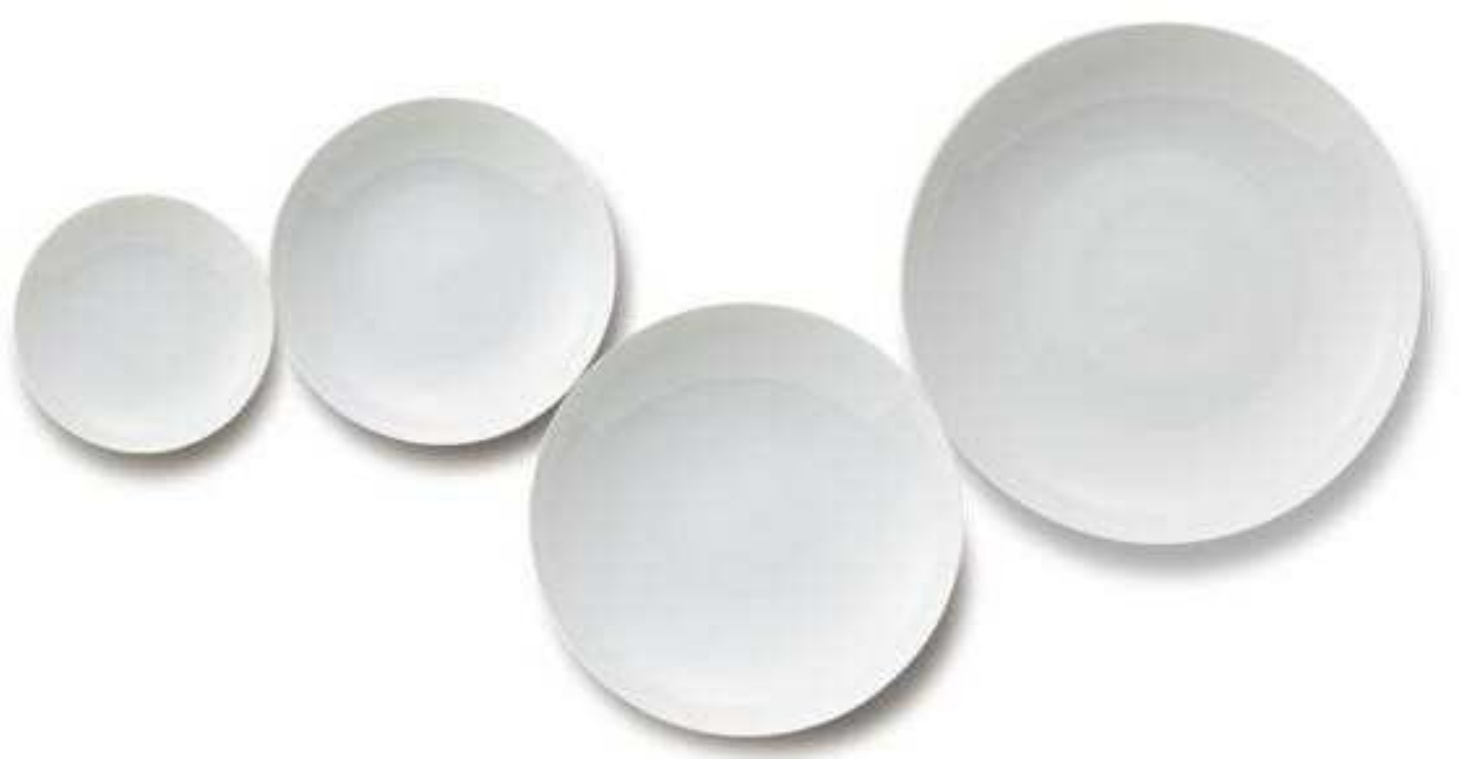}}
  \begin{center} (a) Datasets  \end{center}
\end{minipage}
\begin{minipage}[t]{.16\textwidth}
  \centering
  \centerline{\includegraphics[width=0.92\textwidth]{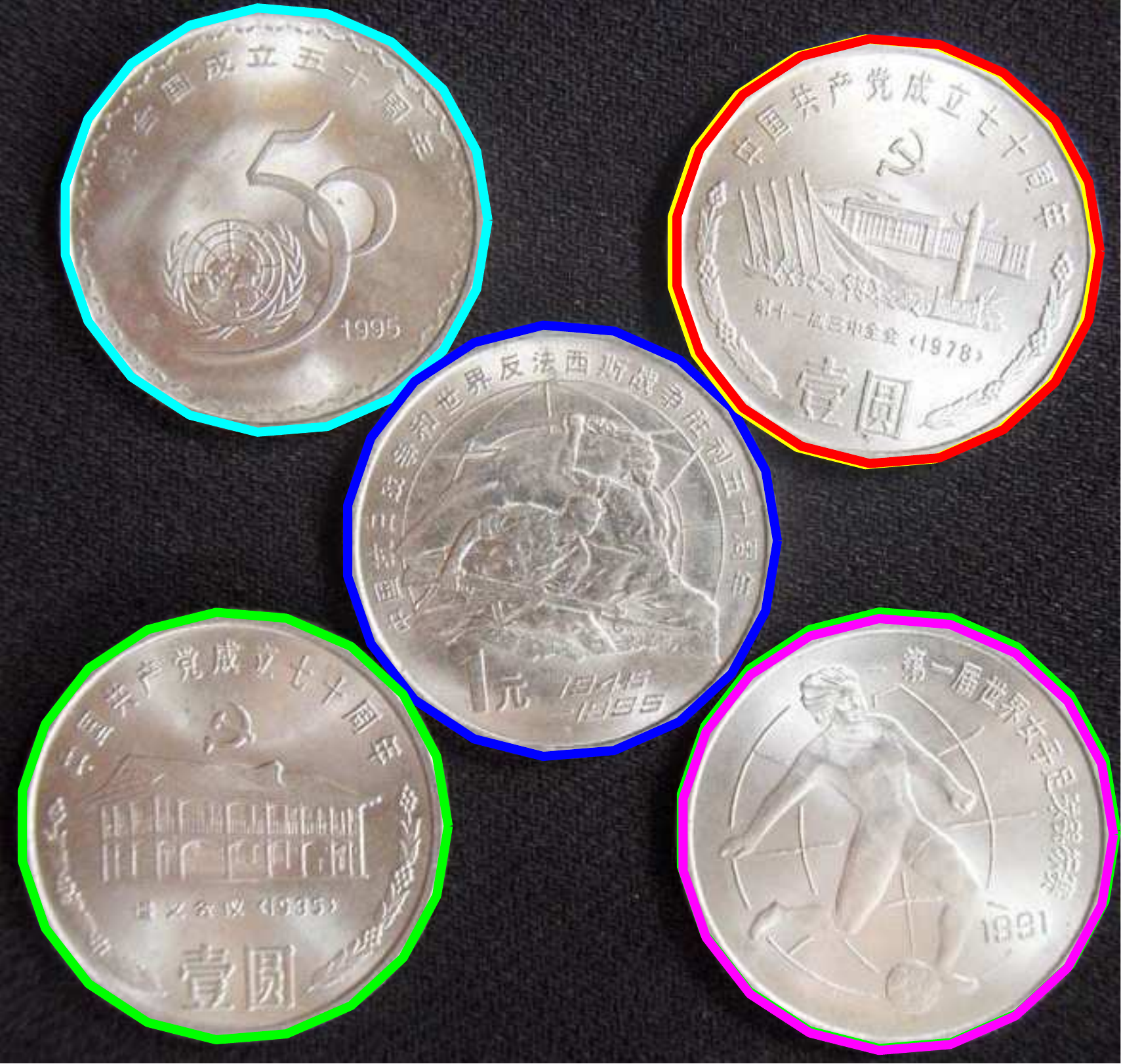}}
  \centerline{}
  \centerline{\includegraphics[width=0.92\textwidth]{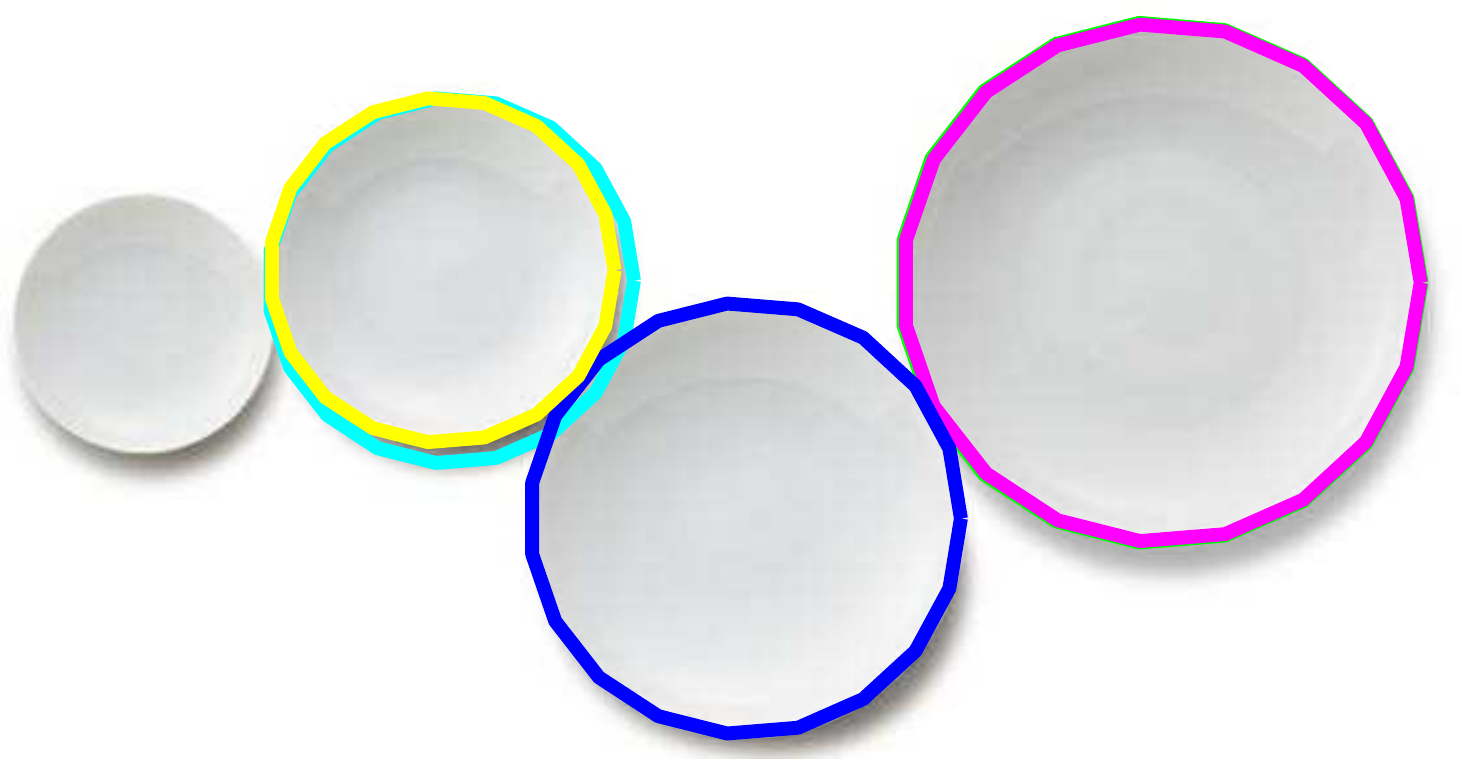}}
  \begin{center} (b) KF  \end{center}
\end{minipage}
\begin{minipage}[t]{.16\textwidth}
  \centering
  \centerline{\includegraphics[width=0.92\textwidth]{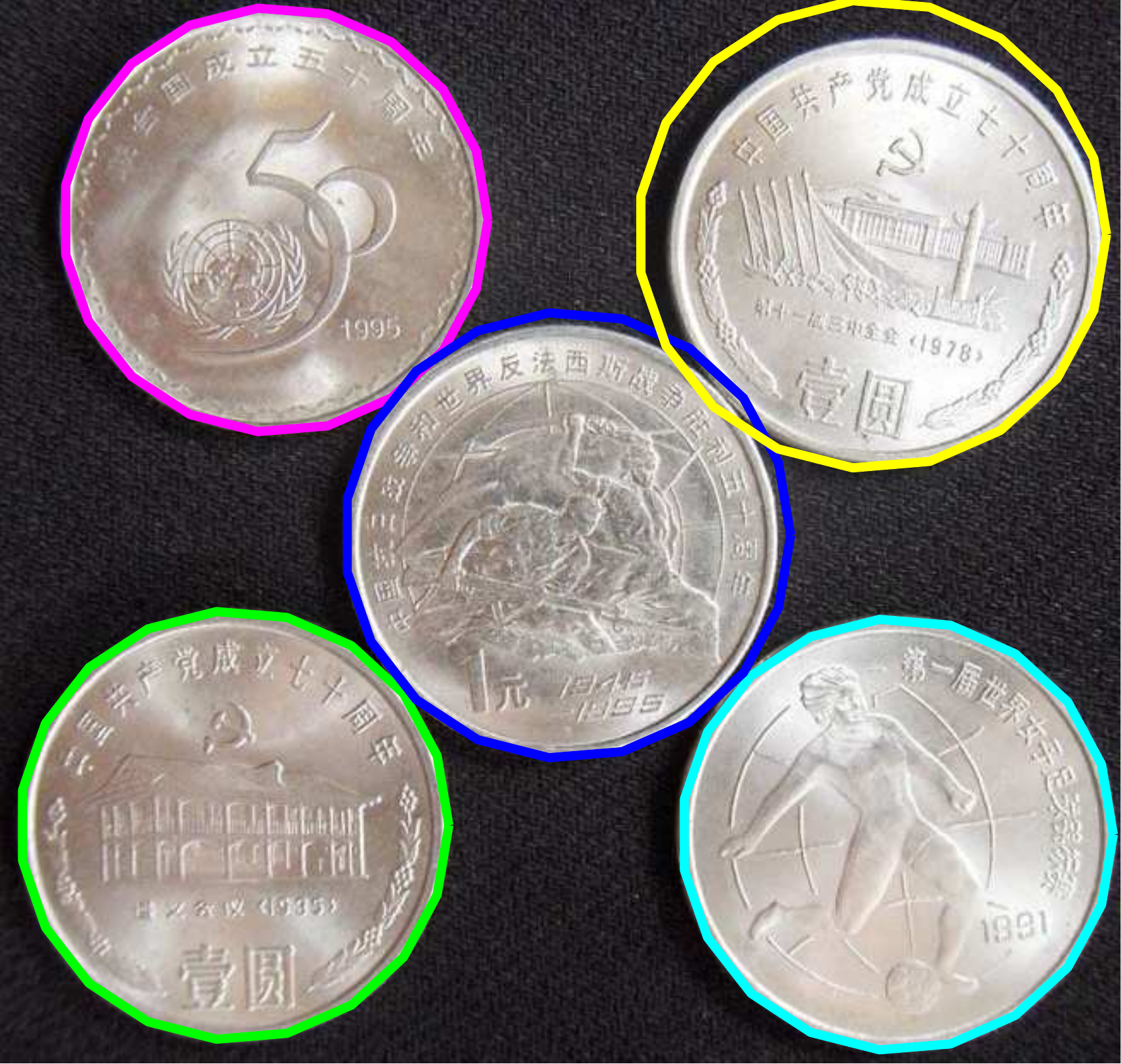}}
  \centerline{}
  \centerline{\includegraphics[width=0.92\textwidth]{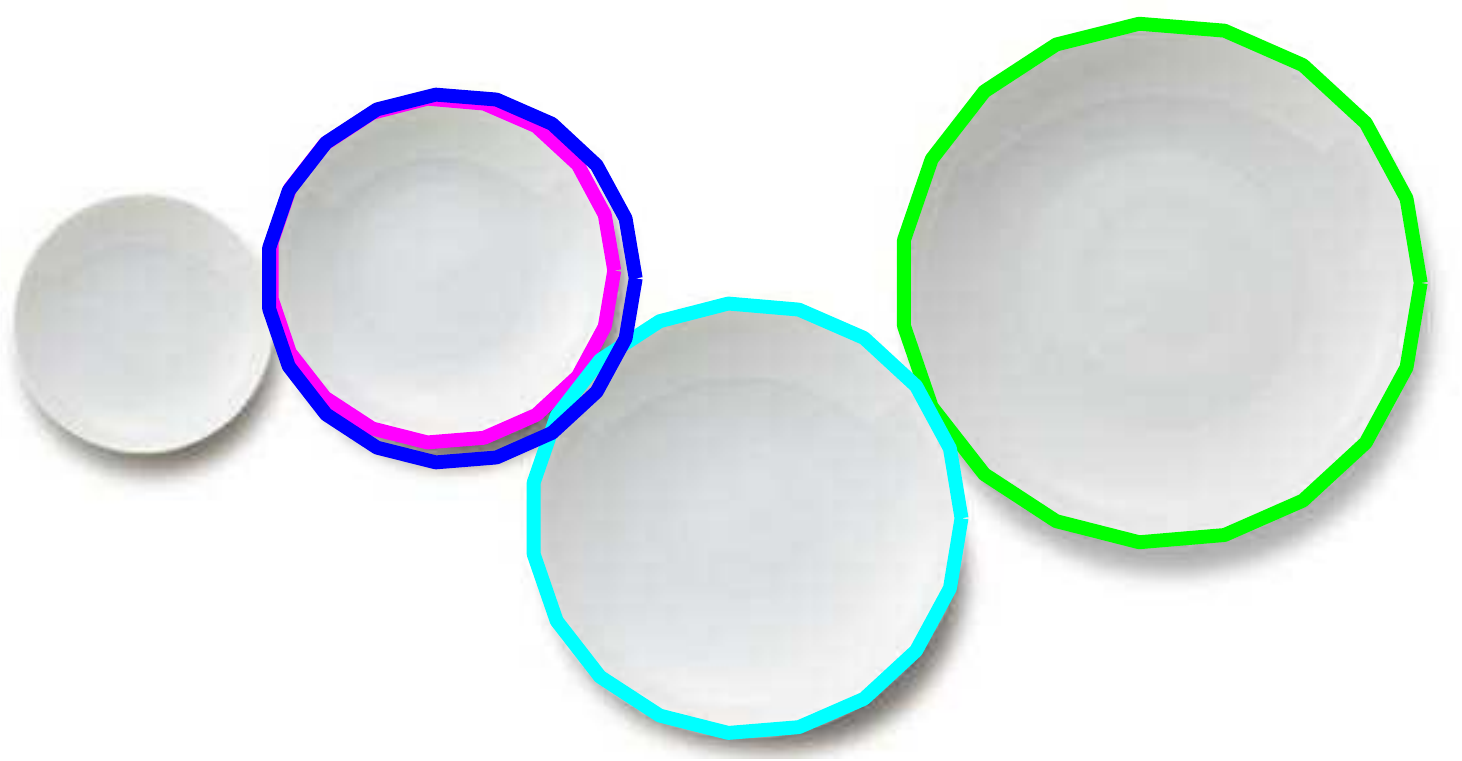}}
  \begin{center} (c) RCG  \end{center}
\end{minipage}
\begin{minipage}[t]{.16\textwidth}
  \centering
  \centerline{\includegraphics[width=0.92\textwidth]{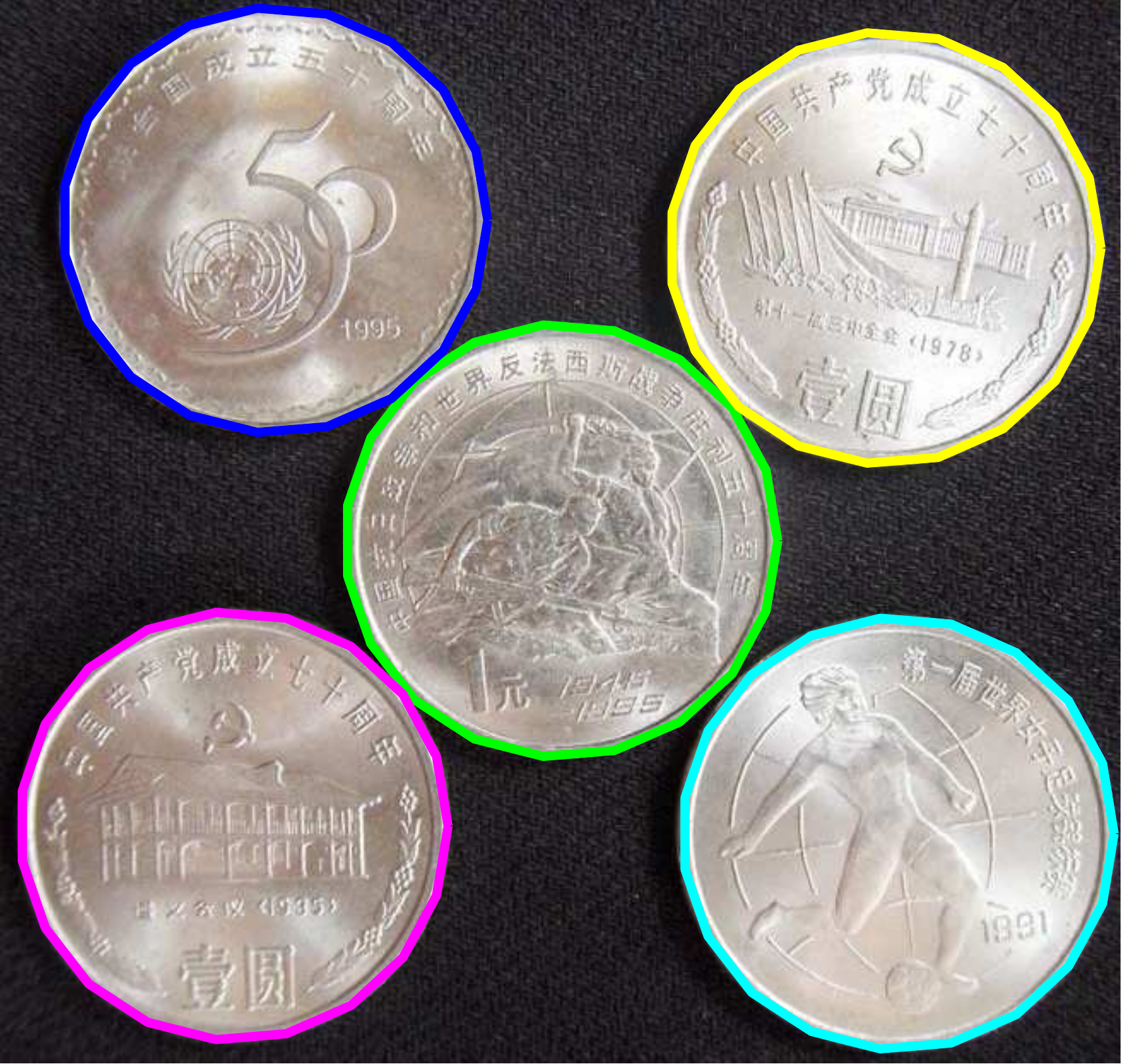}}
  \centerline{}
  \centerline{\includegraphics[width=0.92\textwidth]{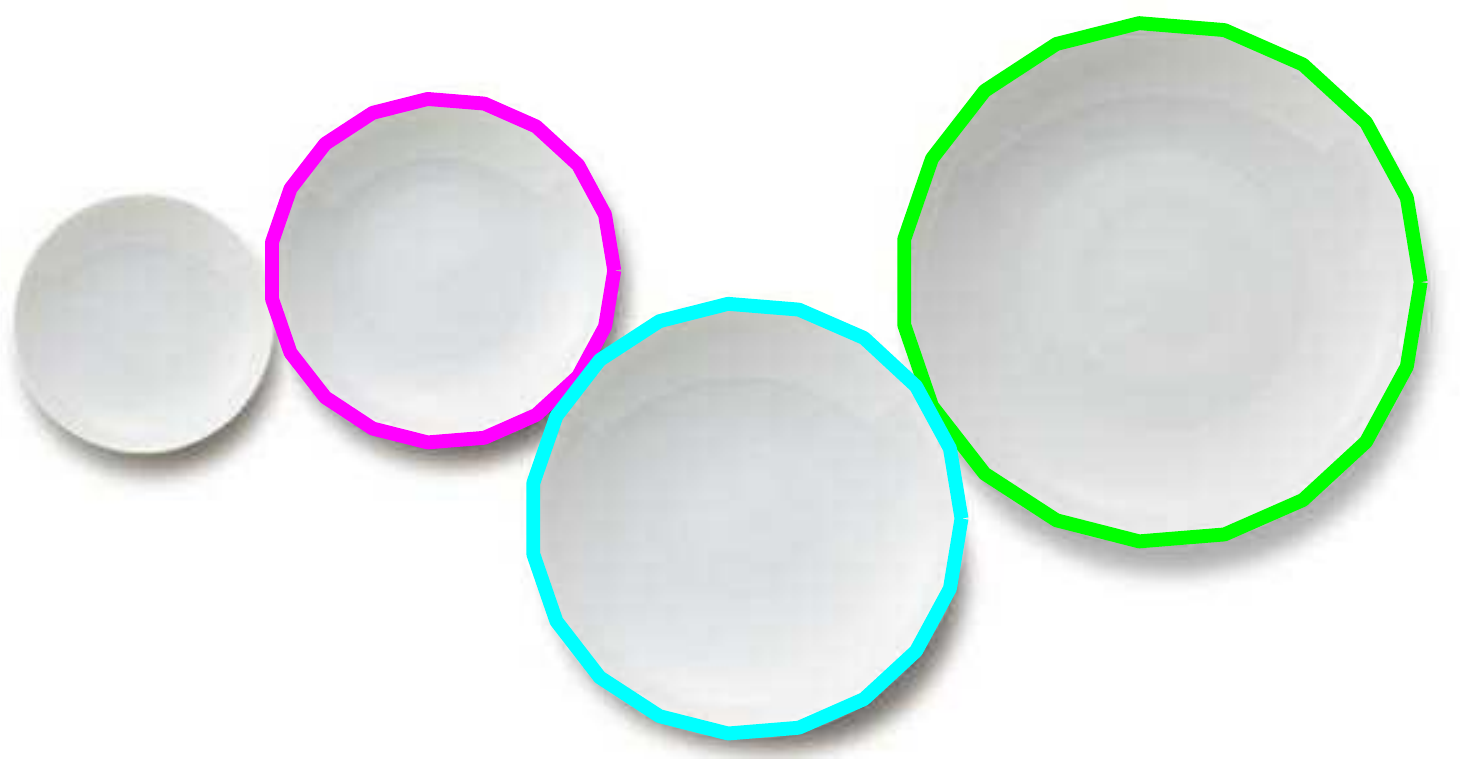}}
  \centerline{(d) AKSWH }\medskip
\end{minipage}
\begin{minipage}[t]{.16\textwidth}
  \centering
  \centerline{\includegraphics[width=0.92\textwidth]{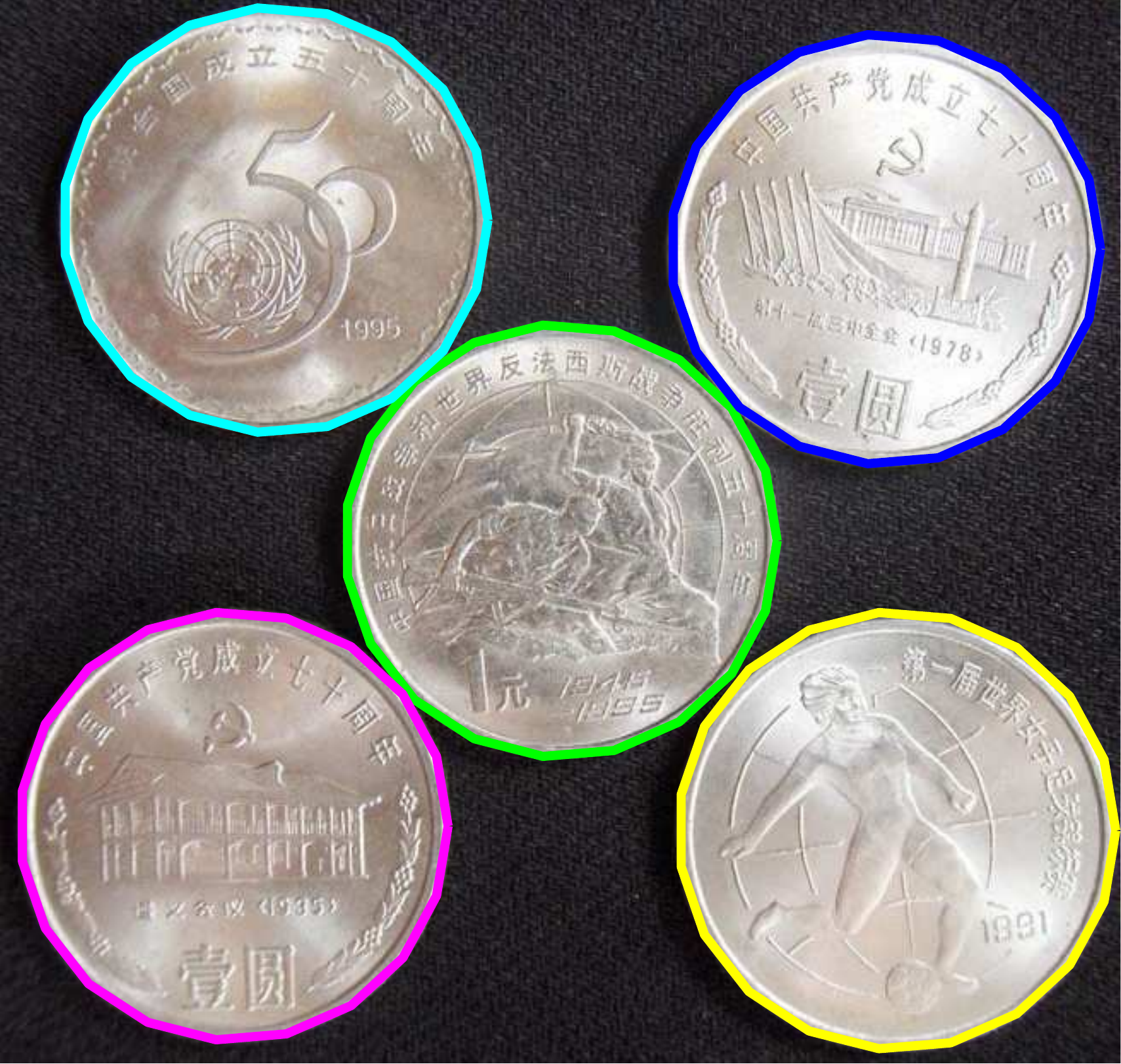}}
  \centerline{}
  \centerline{\includegraphics[width=0.92\textwidth]{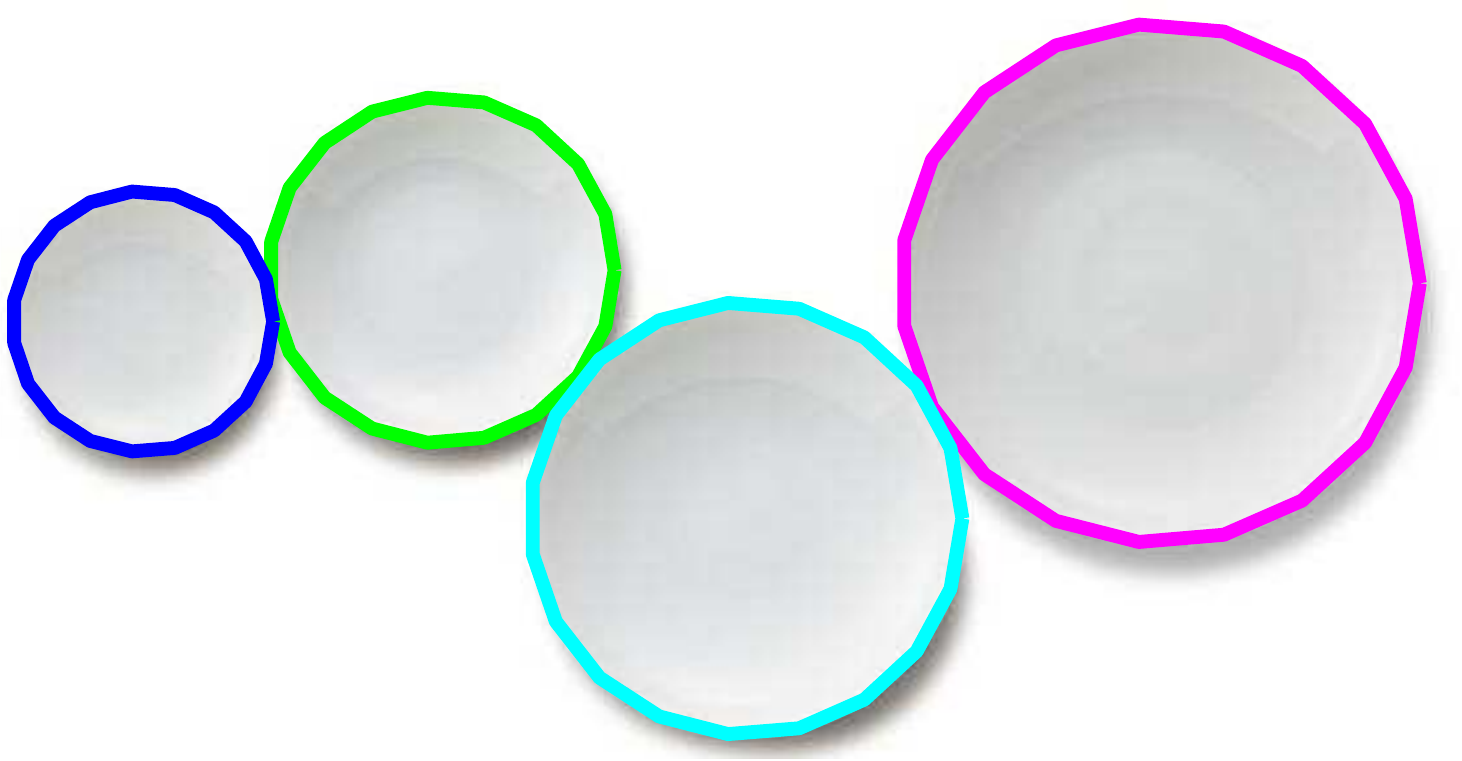}}
  \centerline{(e) T-linkage }\medskip
\end{minipage}
\begin{minipage}[t]{.16\textwidth}
  \centering
  \centerline{\includegraphics[width=0.92\textwidth]{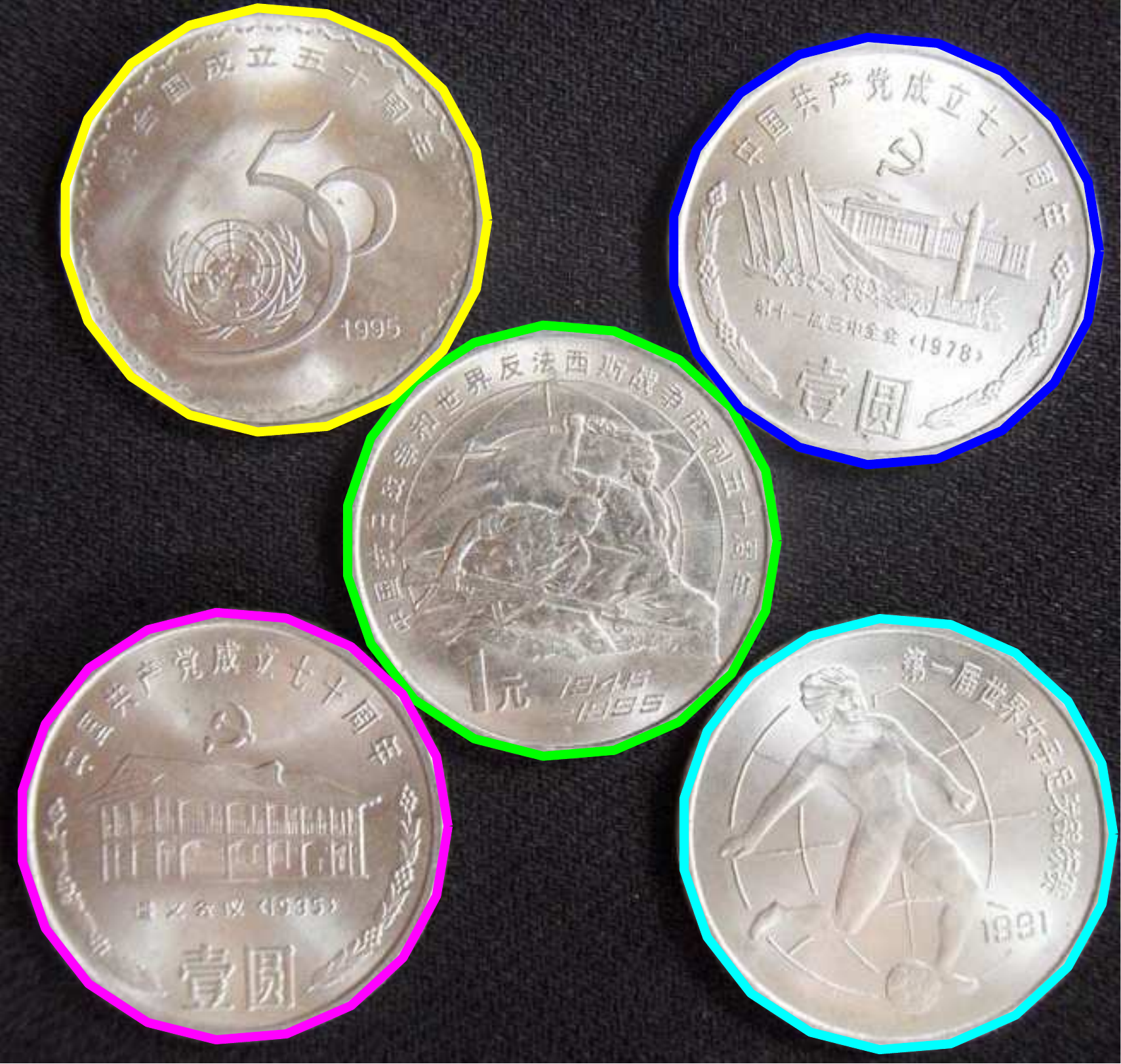}}
  \centerline{}
  \centerline{\includegraphics[width=0.92\textwidth]{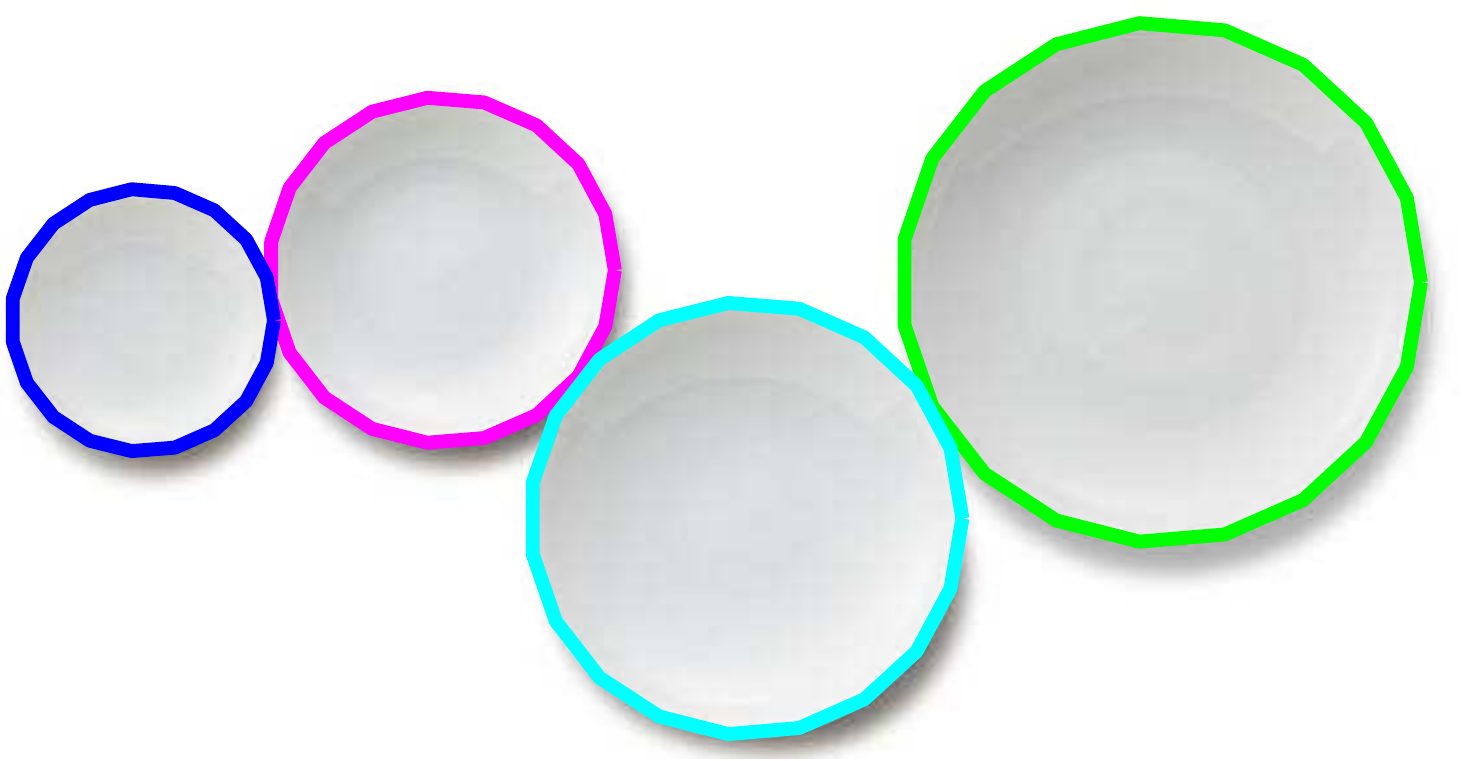}}
  \centerline{(f) MSH }\medskip
\end{minipage}
\hfill
\caption{Examples for circle fitting. First (``coins") and second (``bowls") rows respectively fit five and four circles. (a) The original data. (b) to (f) The results obtained by KF, RCG, AKSWH, T-linkage and MSH, respectively.}
\label{fig:circlefiting}
\end{figure*}

\begin{figure*}
\centering
\begin{minipage}[t]{.12\textwidth}
  \centering
  \centerline{\includegraphics[width=1.15\textwidth]{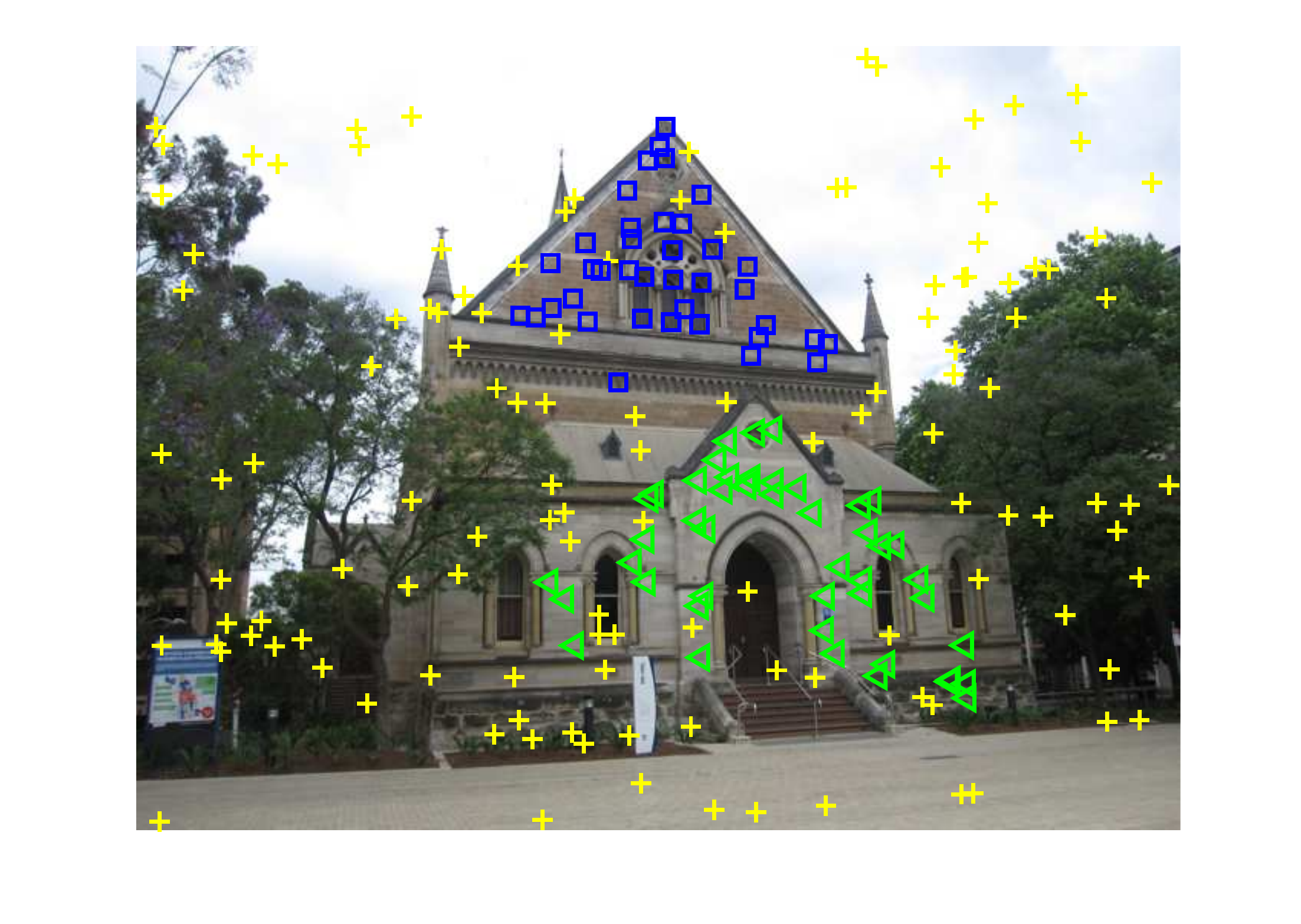}}
  \centerline{\includegraphics[width=1.15\textwidth]{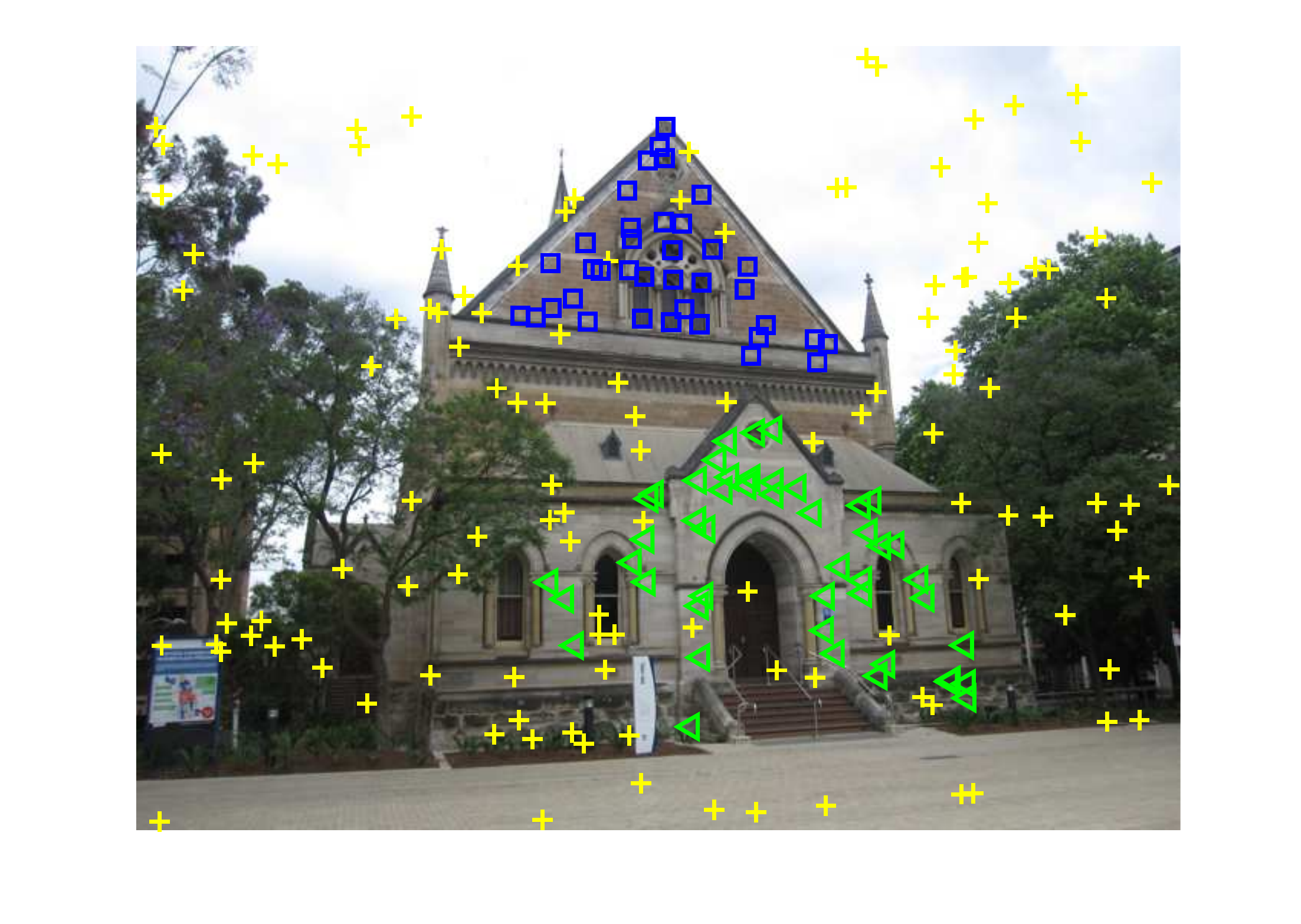}}
  \begin{center} (a) Elderhalla  \end{center}
\end{minipage}
\begin{minipage}[t]{.12\textwidth}
  \centering
  \centerline{\includegraphics[width=1.15\textwidth]{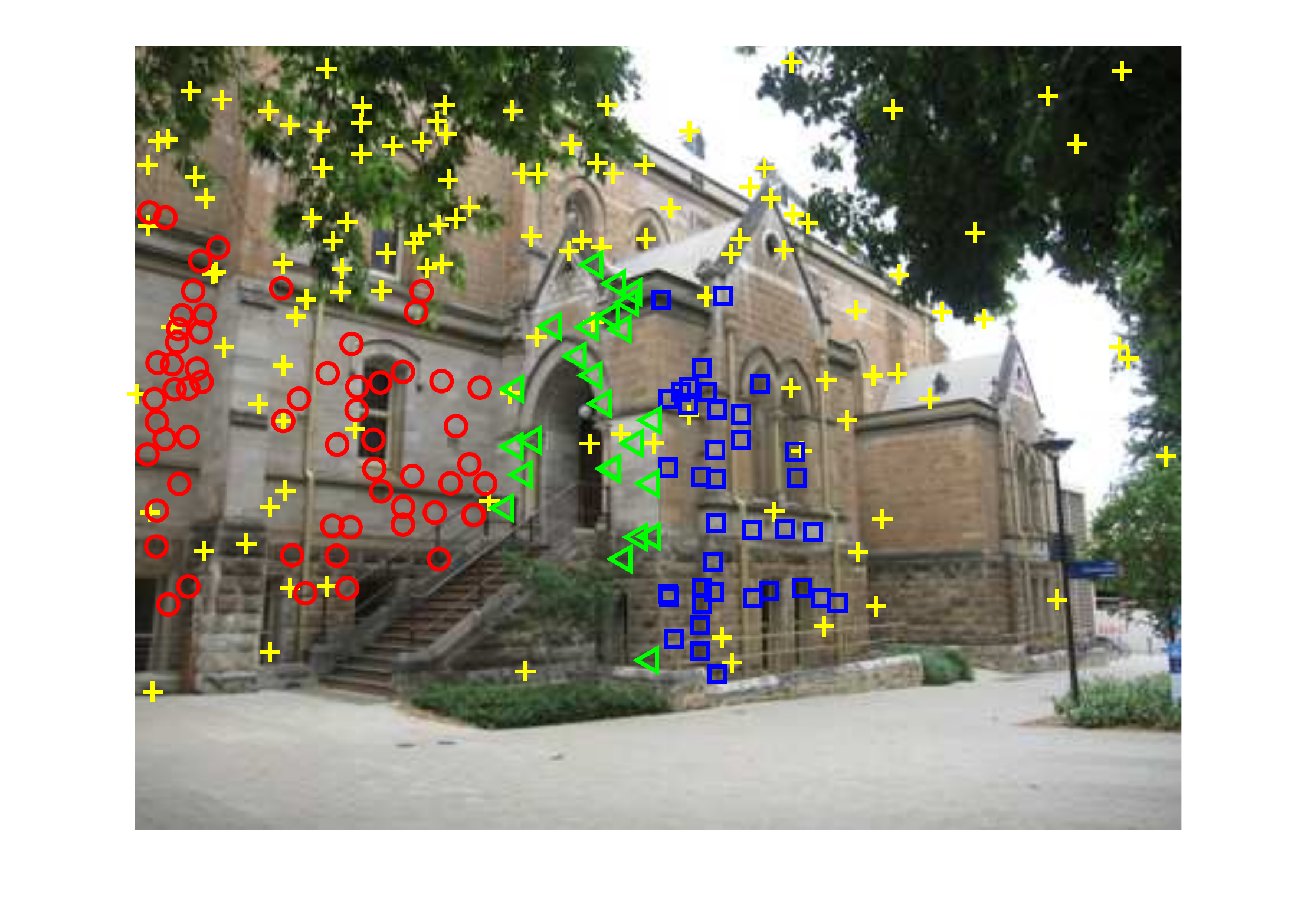}}
  \centerline{\includegraphics[width=1.15\textwidth]{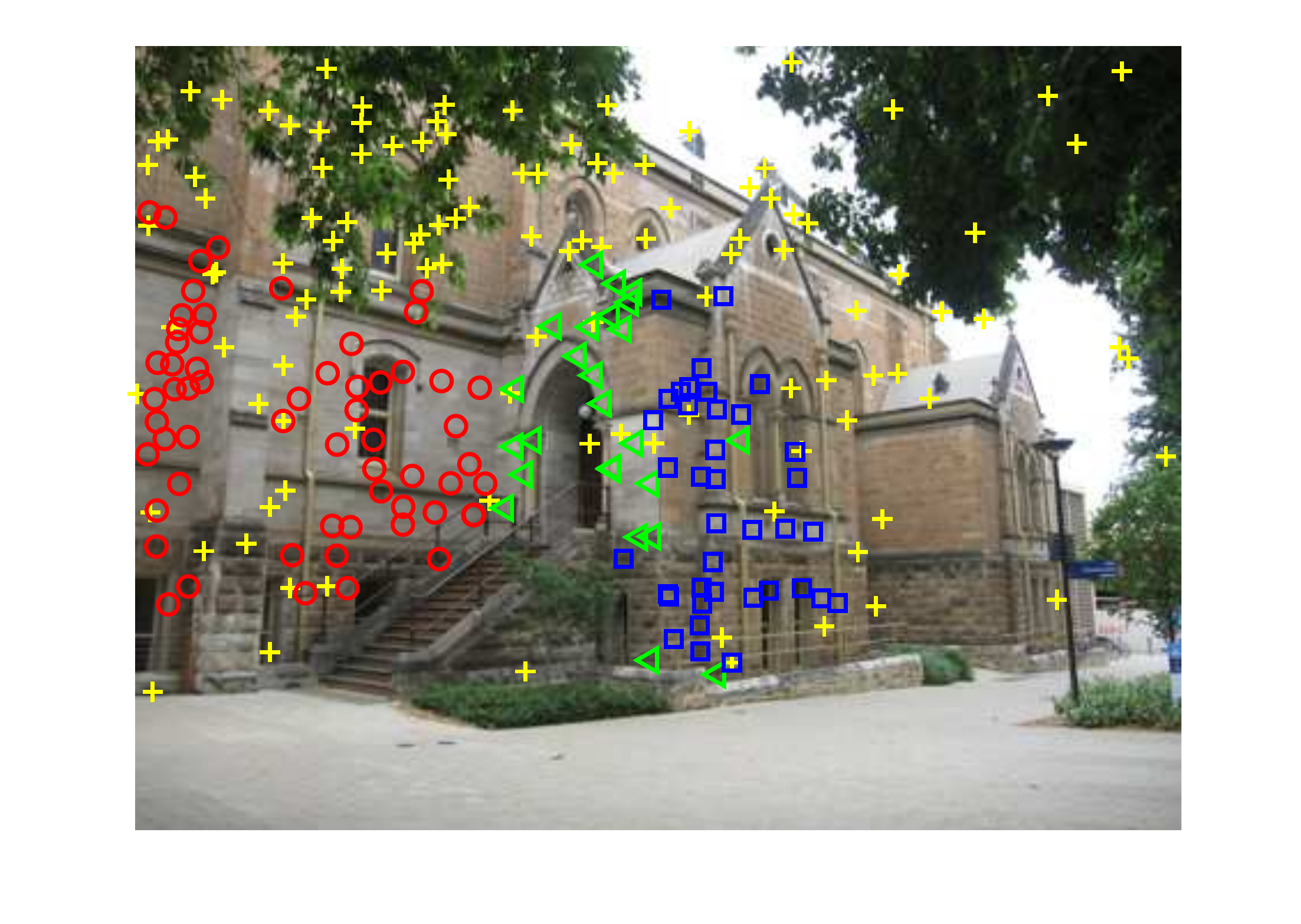}}
  \begin{center} (b) Elderhallb  \end{center}
\end{minipage}
\begin{minipage}[t]{.12\textwidth}
  \centering
  \centerline{\includegraphics[width=1.15\textwidth]{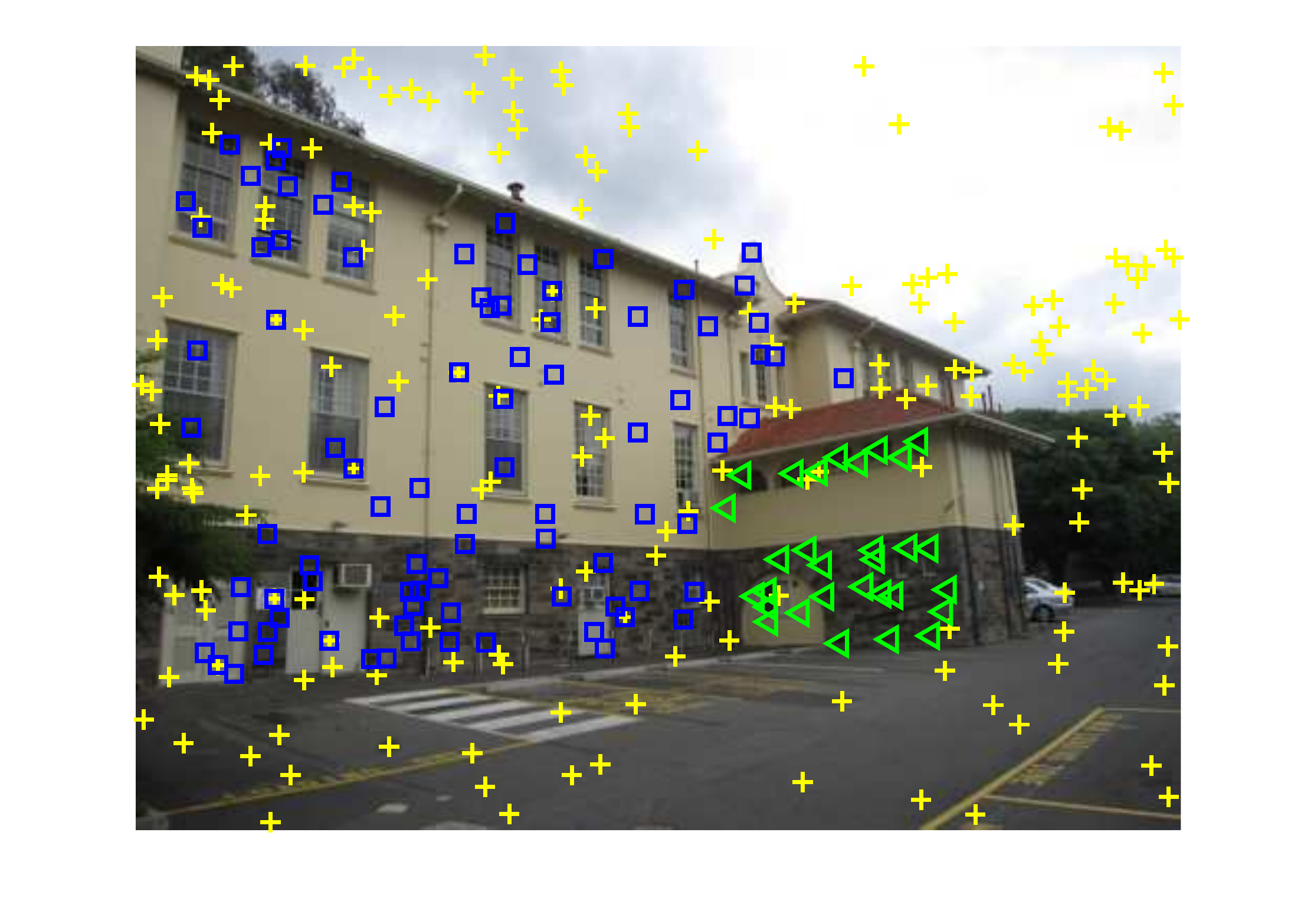}}
  \centerline{\includegraphics[width=1.15\textwidth]{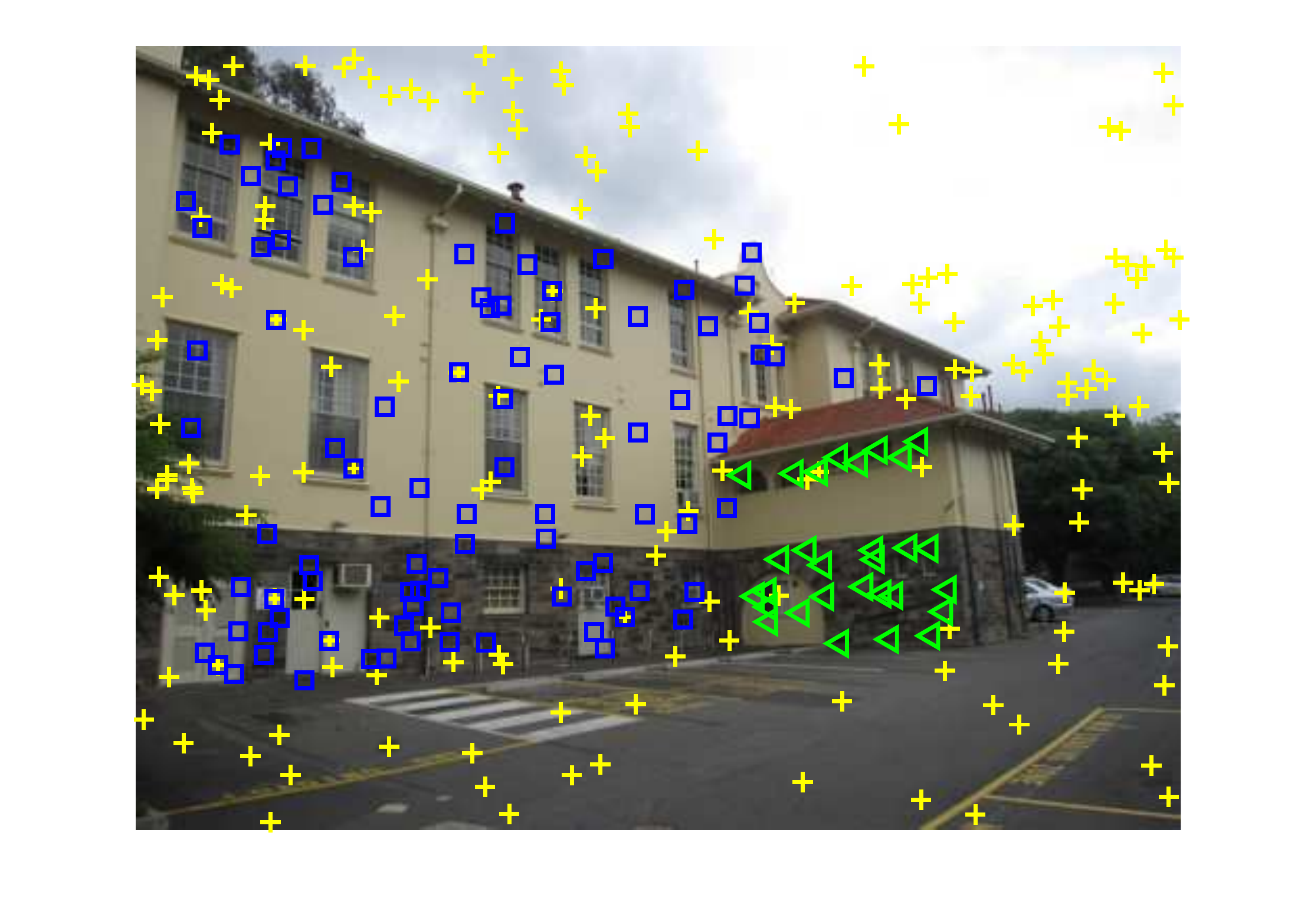}}
  \begin{center} (c) Hartley  \end{center}
\end{minipage}
\begin{minipage}[t]{.12\textwidth}
  \centering
  \centerline{\includegraphics[width=1.15\textwidth]{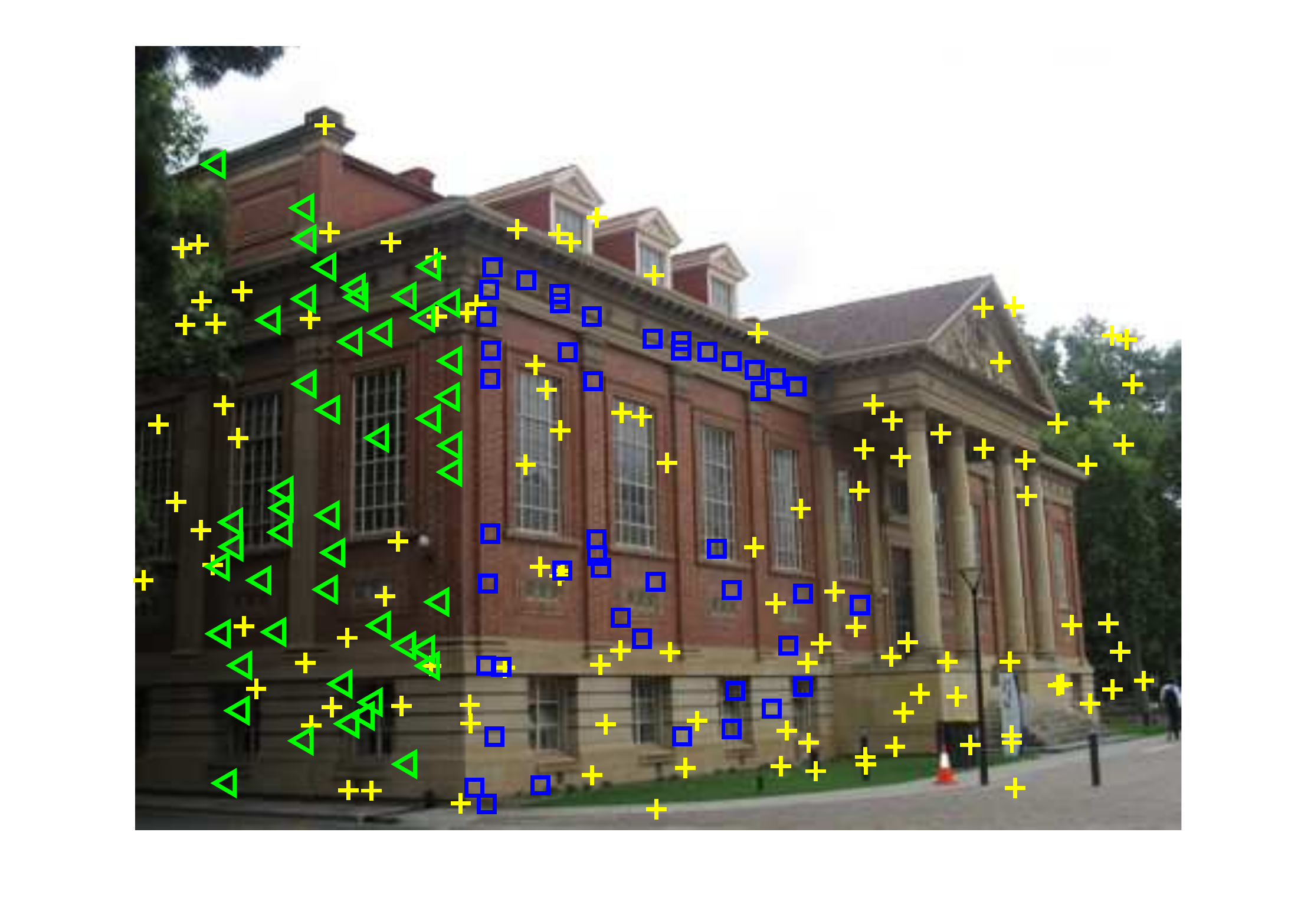}}
  \centerline{\includegraphics[width=1.15\textwidth]{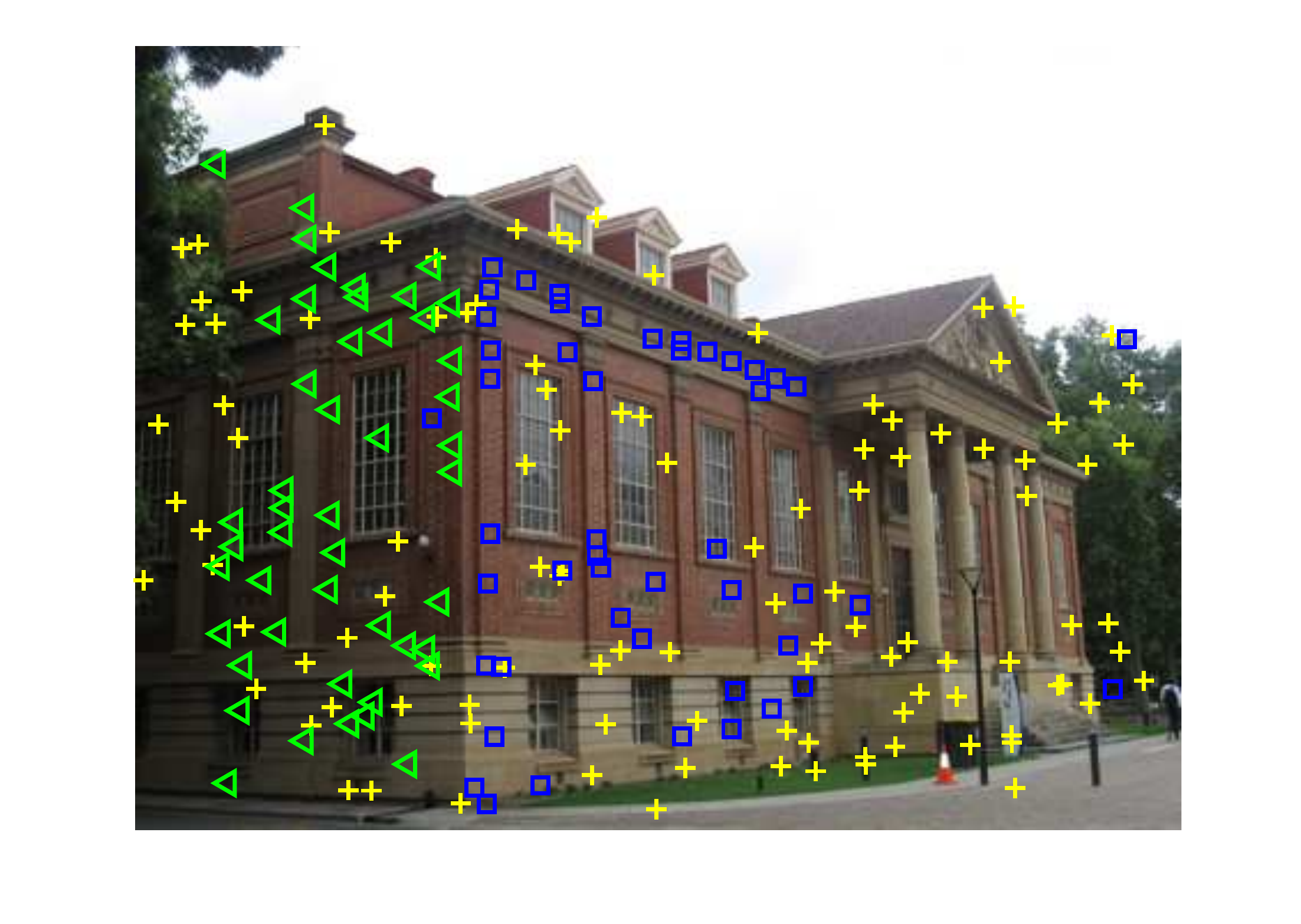}}
  \centerline{(d) Library }\medskip
\end{minipage}
\begin{minipage}[t]{.12\textwidth}
  \centering
  \centerline{\includegraphics[width=1.15\textwidth]{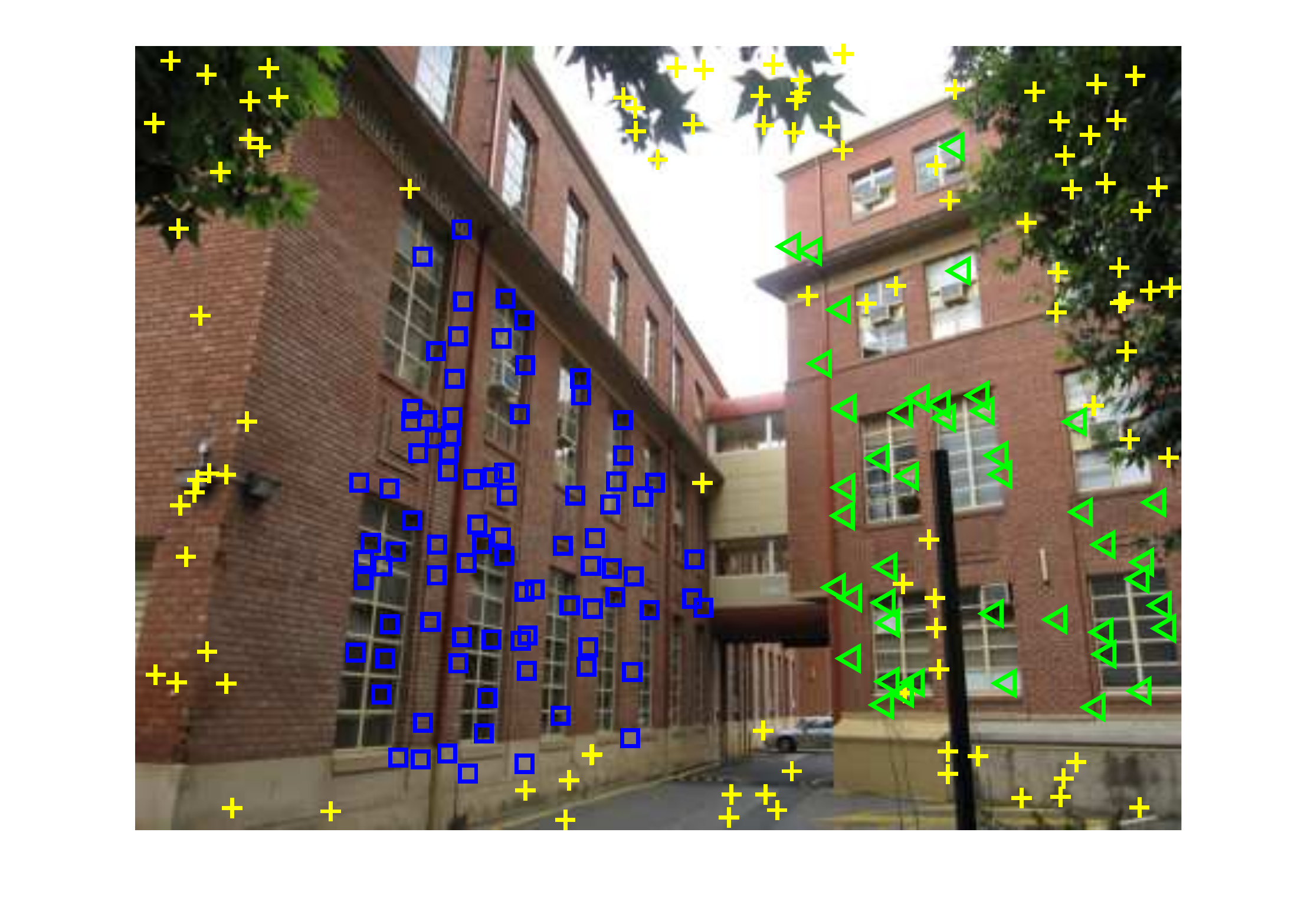}}
  \centerline{\includegraphics[width=1.15\textwidth]{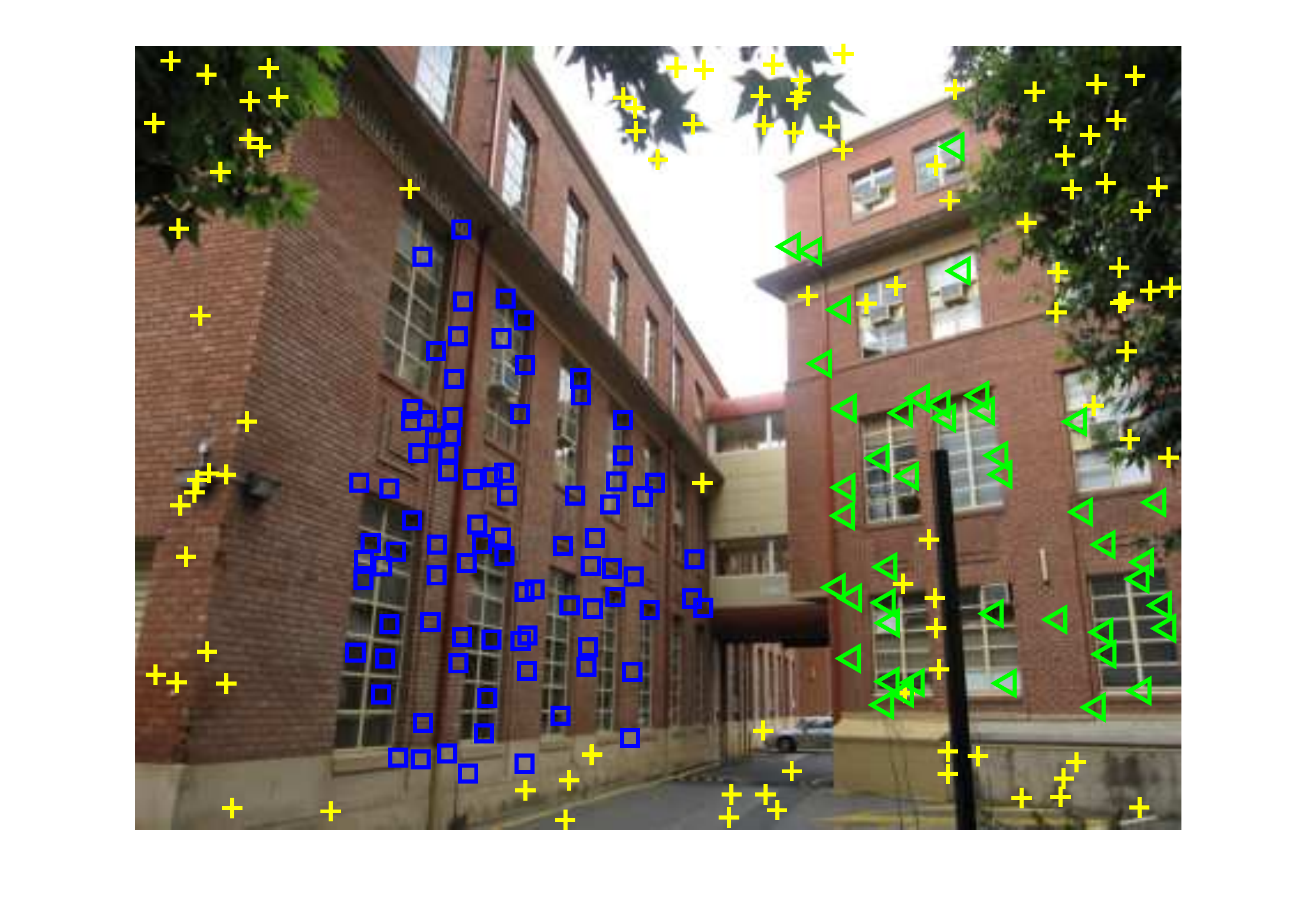}}
  \centerline{(e) Sene }\medskip
\end{minipage}
\begin{minipage}[t]{.12\textwidth}
  \centering
  \centerline{\includegraphics[width=1.15\textwidth]{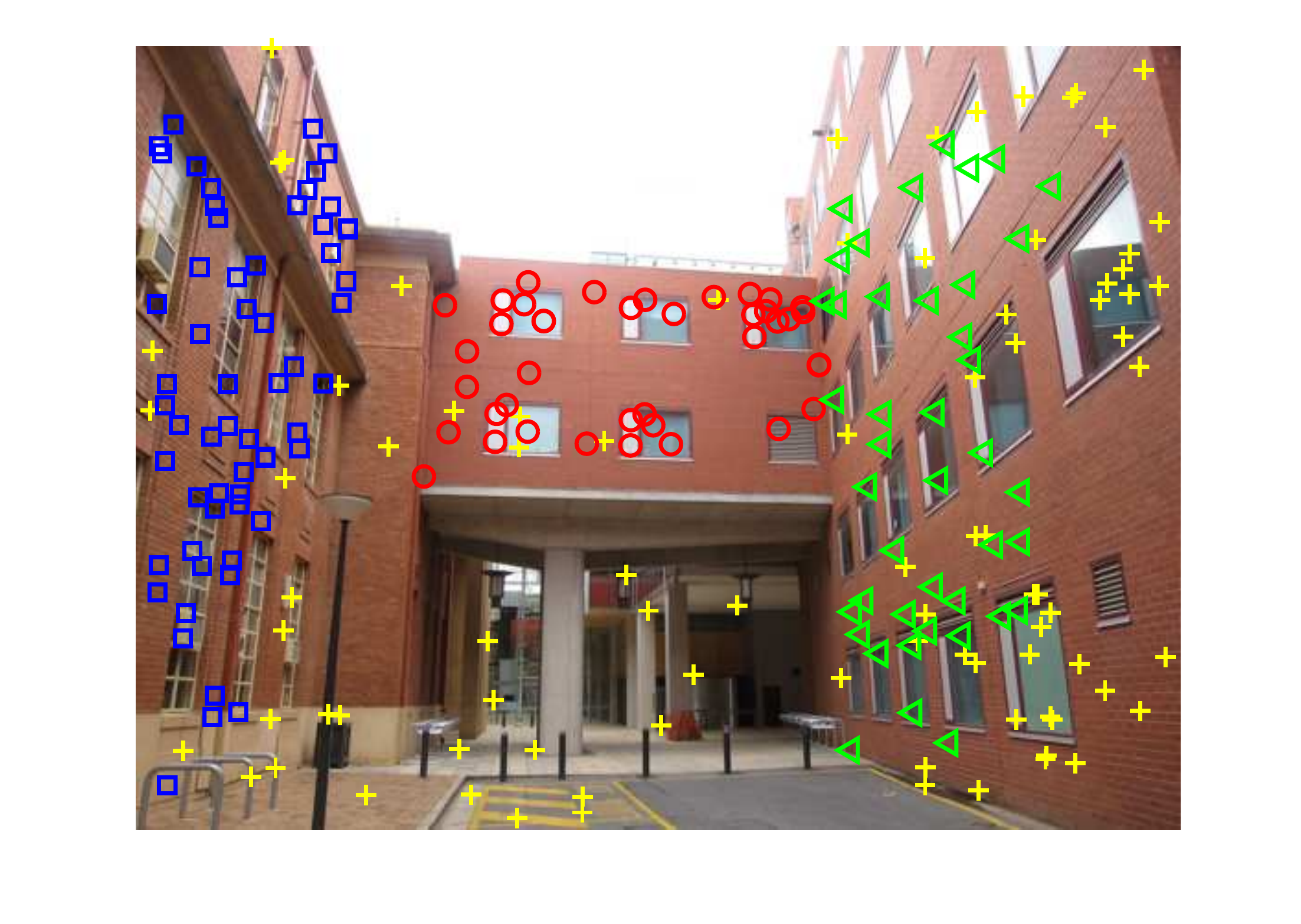}}
  \centerline{\includegraphics[width=1.15\textwidth]{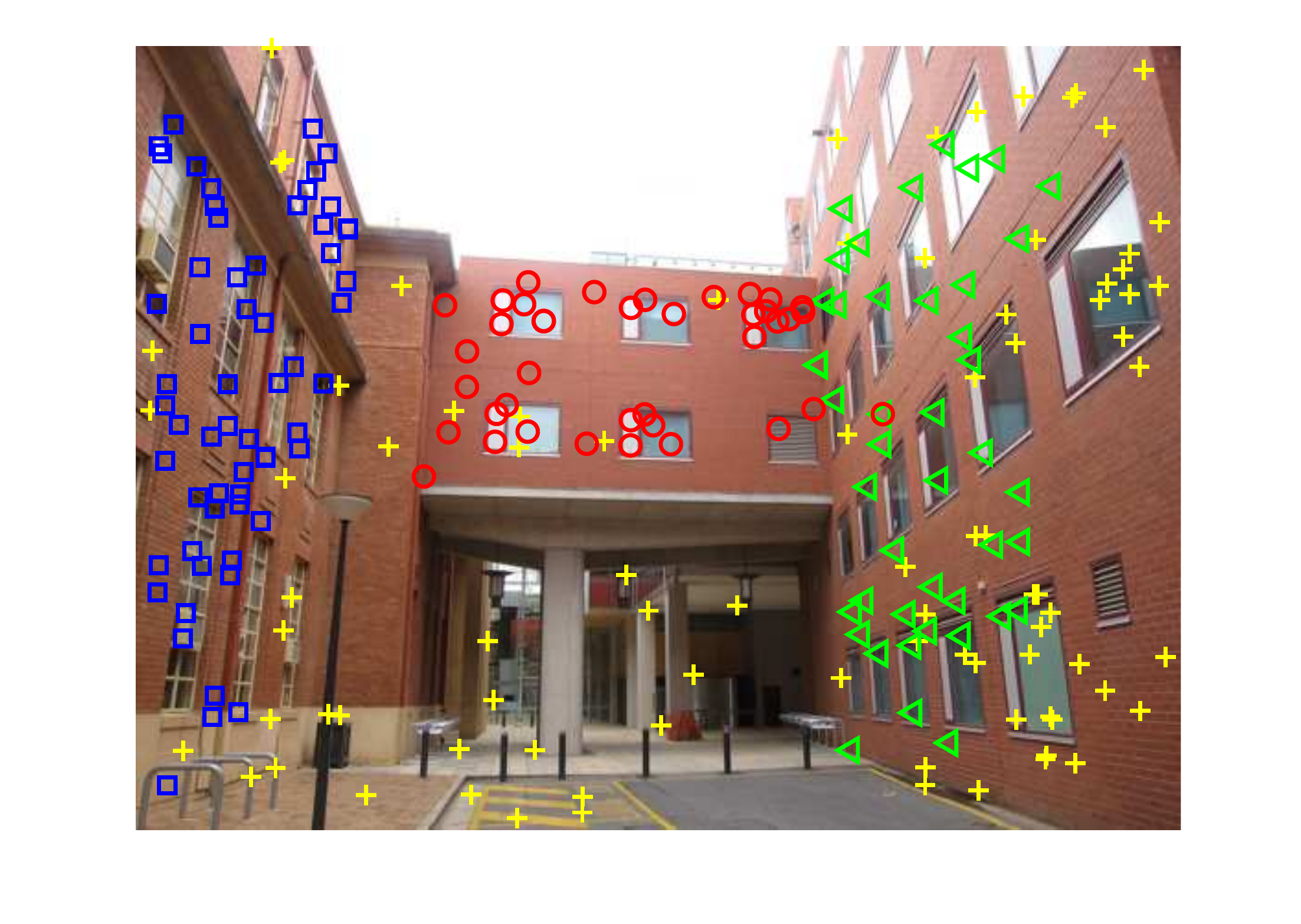}}
  \centerline{(f) Neem }\medskip
\end{minipage}
\begin{minipage}[t]{.12\textwidth}
  \centering
  \centerline{\includegraphics[width=1.15\textwidth]{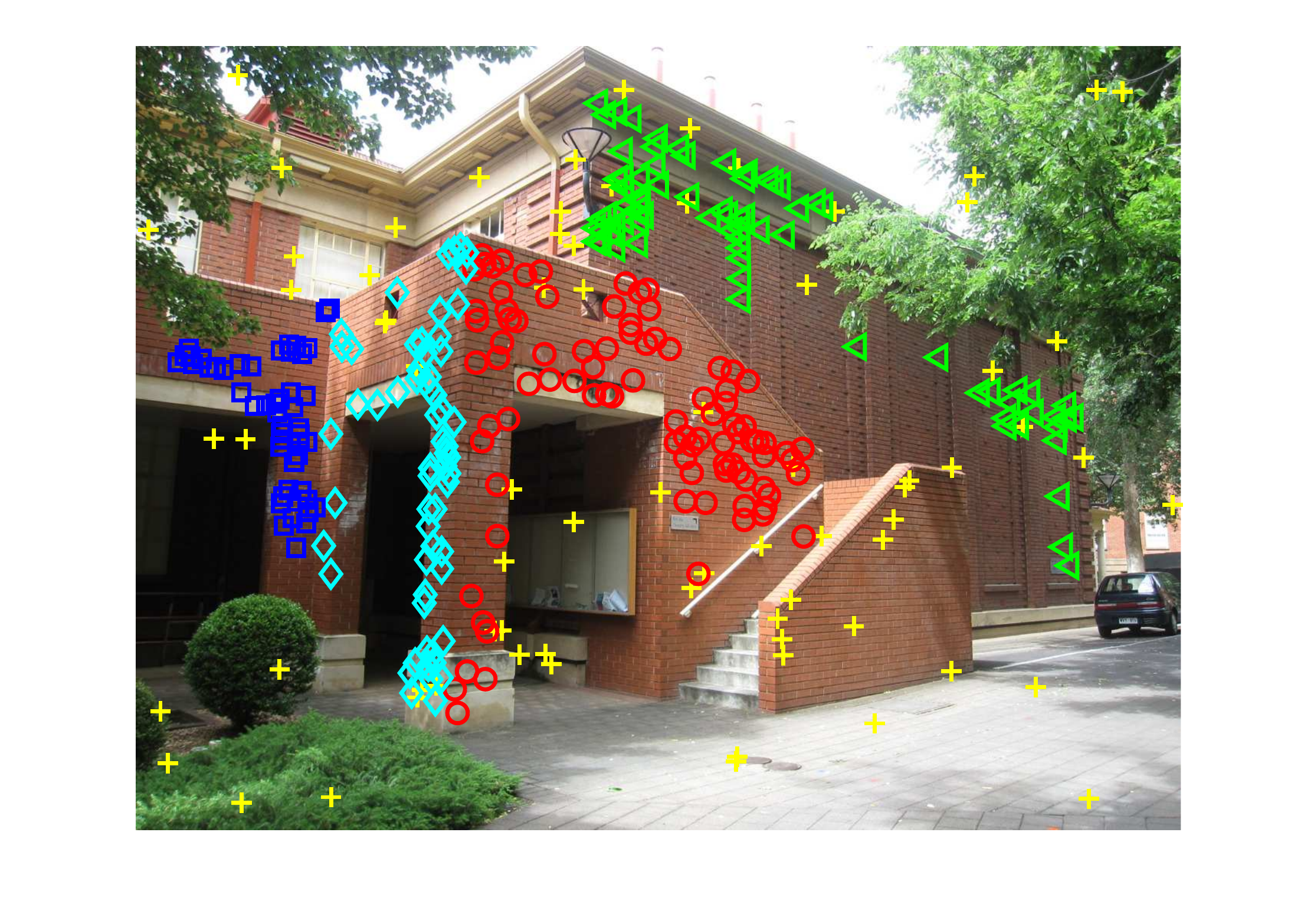}}
  \centerline{\includegraphics[width=1.15\textwidth]{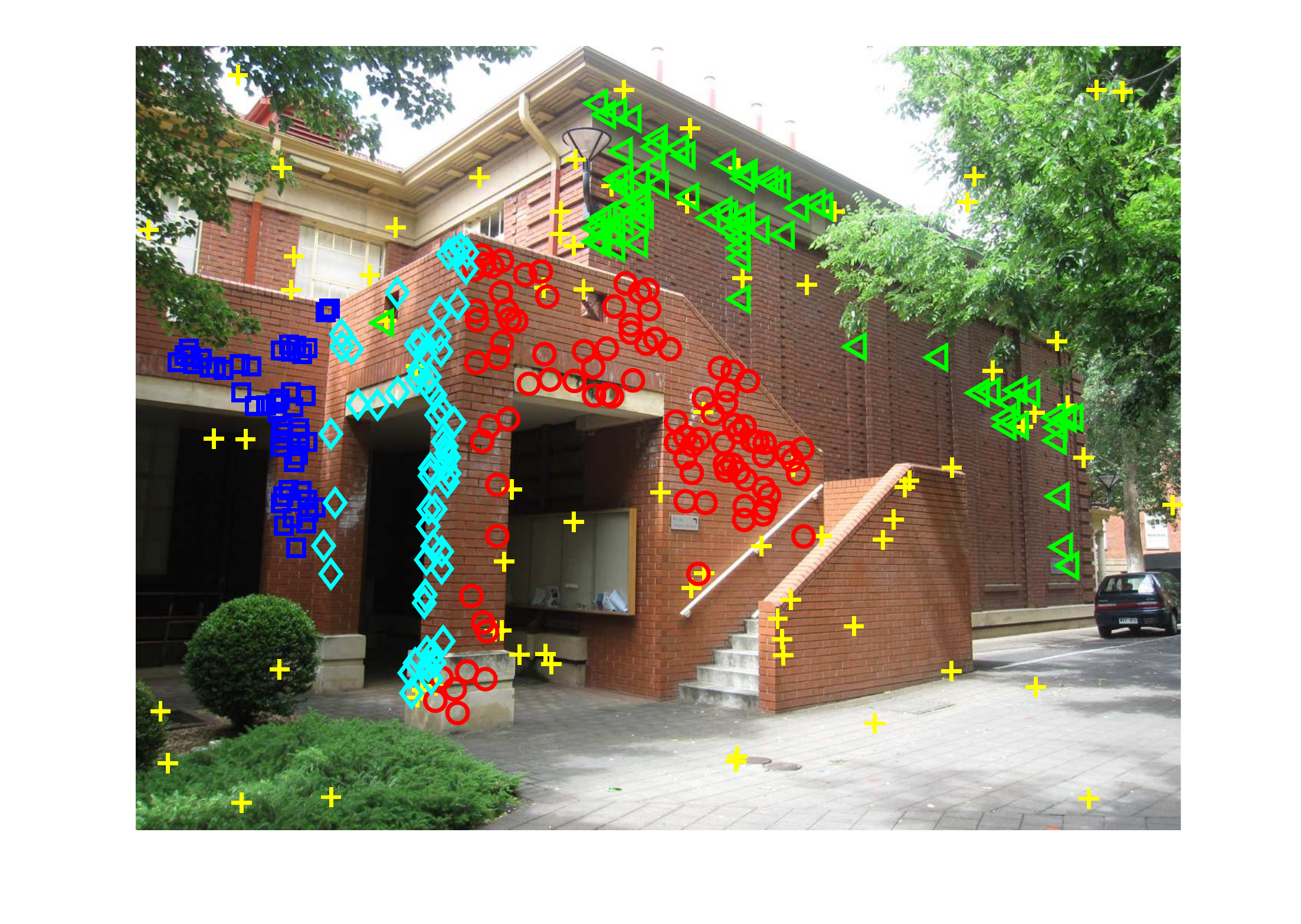}}
  \centerline{(g) Johnsona }\medskip
\end{minipage}
\begin{minipage}[t]{.12\textwidth}
  \centering
  \centerline{\includegraphics[width=1.15\textwidth]{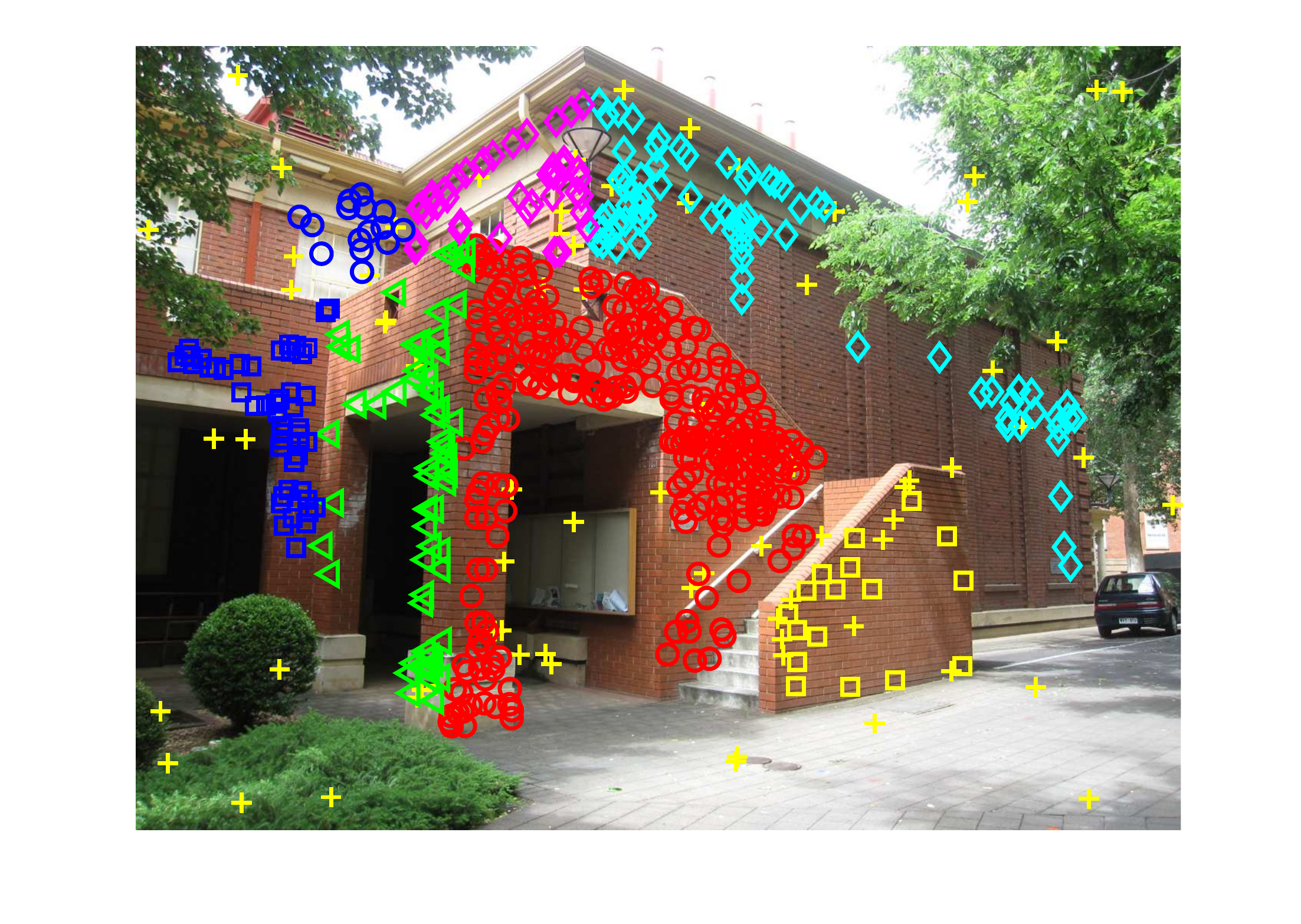}}
  \centerline{\includegraphics[width=1.15\textwidth]{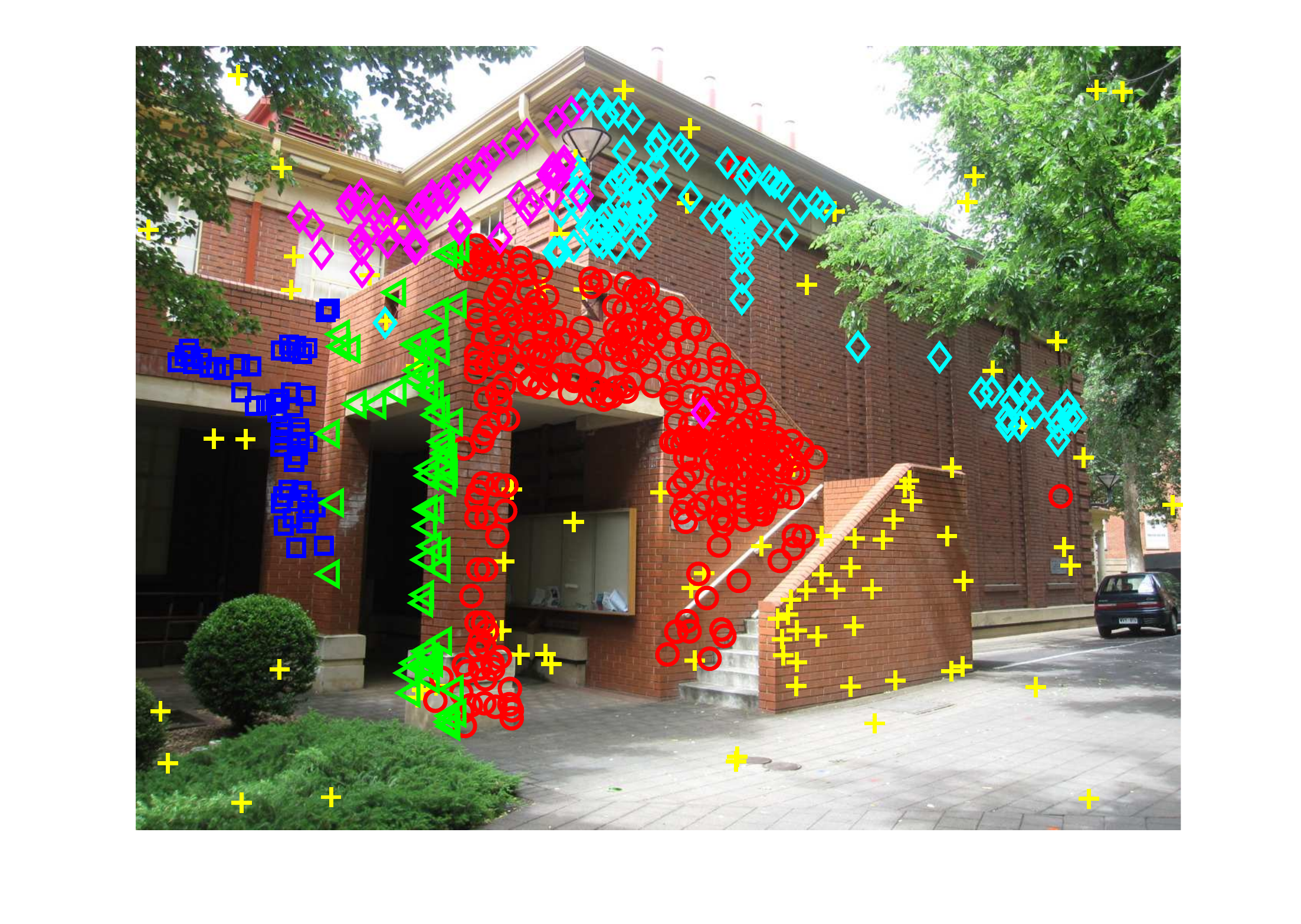}}
  \centerline{(h) Johnsonb }\medskip
\end{minipage}
\hfill
\caption{Homography based segmentation on eight image pairs. The first and second rows are the original images with the ground truth results and the segmentation results obtained by MSH, respectively. We do not show the results obtained by the other competing methods due to the space limit.}
\label{fig:homography}
\end{figure*}

\begin{table*}
\small\
  \caption{The fitting errors (in percentage) for homography based segmentation on eight dataset (the best results are boldfaced)}
  \medskip
\centering
\begin{tabular}{|c|c|c|c|c|c|c|c|c|c|c|c|}
\hline
\multirow{2}{*}{} & \multicolumn{2}{c|}{KF} & \multicolumn{2}{c|}{RCG}& \multicolumn{2}{c|}{AKSWH} & \multicolumn{2}{c|}{T-linkage}&\multicolumn{2}{c|}{MSH}\\
\cline{2-11}
& Avg. &  Min. & Avg. & Min. & Avg. &Min.& Avg. & Min.& Avg. & Min. \\
\hline
Elderhalla & {12.15} & {7.22} & {10.37} & {9.81}  & {0.98} & {\bf0.93} & {1.17} & {\bf0.93} & {\bf0.93} & {\bf0.93} \\
 \hline
 Elderhallb & {34.51} & {34.51} & {10.12} & {7.45}  & {13.06} & {11.34} & {12.63} & {11.76} & {\bf3.37} & {\bf1.96} \\
 \hline
Hartley   & {15.31} & {11.24} & {4.88} & {2.81} & {4.06} & {1.87} & {\bf2.50}  & {1.87}   & {2.81} & {\bf1.56} \\
 \hline
Library   & {13.19} & {10.84} & {9.77} & {9.77}  & {5.79} & {\bf1.40} & {4.65} & {1.29} & {\bf2.79} & {\bf1.40} \\
\hline
Sene      & {12.08} & {8.01} & {10.00} & {6.10}  & {2.00}  & {\bf0.00} & {0.44} & {\bf0.00} & {\bf0.24} & {\bf0.00}  \\
\hline
Neem      & {10.25} & {8.37} & {11.17} & {9.50}  & {5.56} & {1.66} & {3.82} & {\bf1.24} & {\bf2.90} & {\bf1.24} \\
\hline
Johnsona  & {25.74} & {12.43} & {23.06} & {10.22}  & {8.55}  & {2.41} & {4.03} & {2.68} & {\bf3.73} & {\bf1.88} \\
\hline
Johnsonb & {48.32} & {42.84} & {41.45} & {22.93} & {26.49}  & {22.65} & {18.39} & {12.40} & {\bf16.75} & {\bf9.86}  \\
\hline
\end{tabular}
 \label{table:homographytable}
\end{table*}
\begin{figure*}
\centering
\begin{minipage}[t]{.12\textwidth}
  \centering
  \centerline{\includegraphics[width=1.15\textwidth]{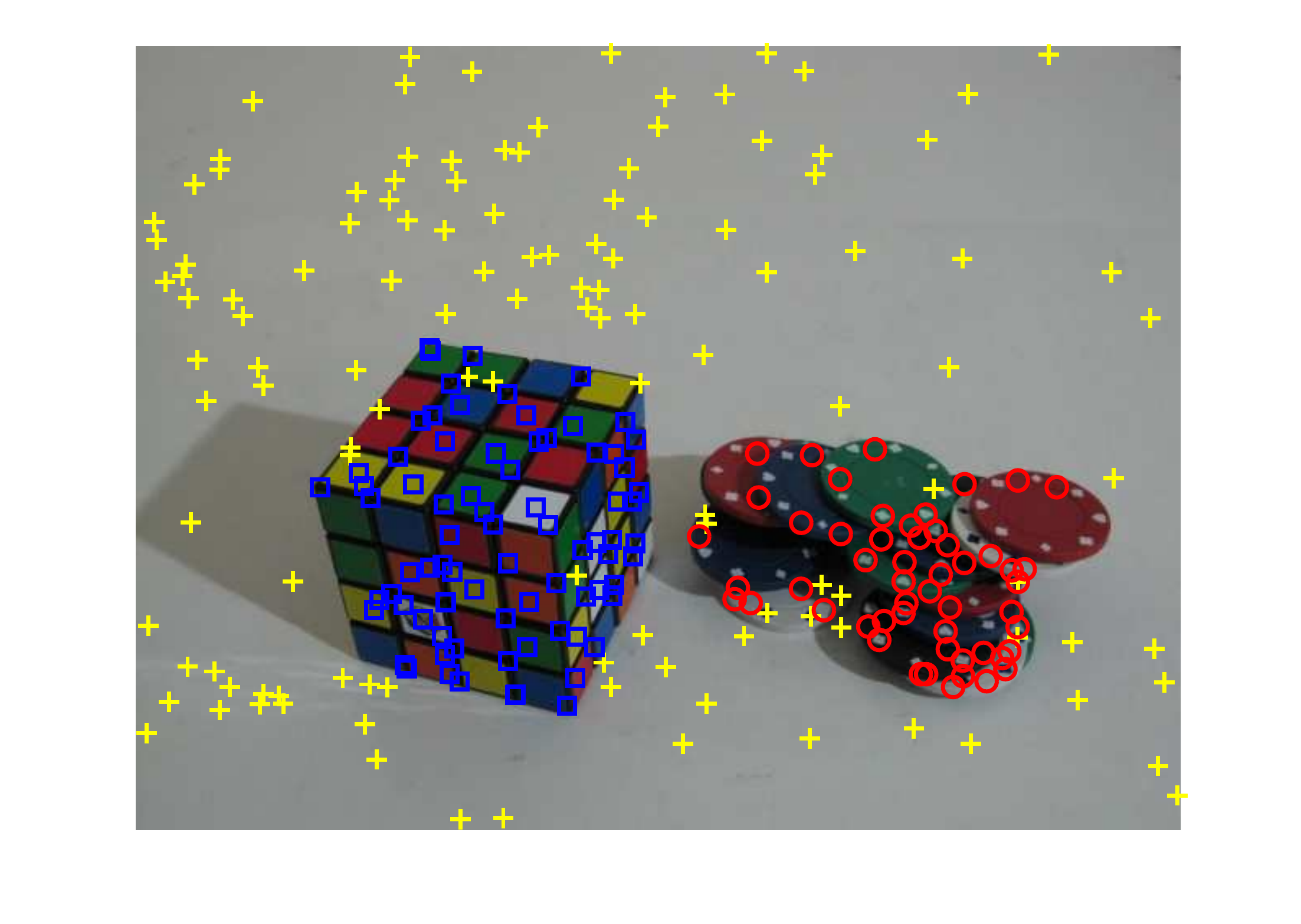}}
  \centerline{\includegraphics[width=1.15\textwidth]{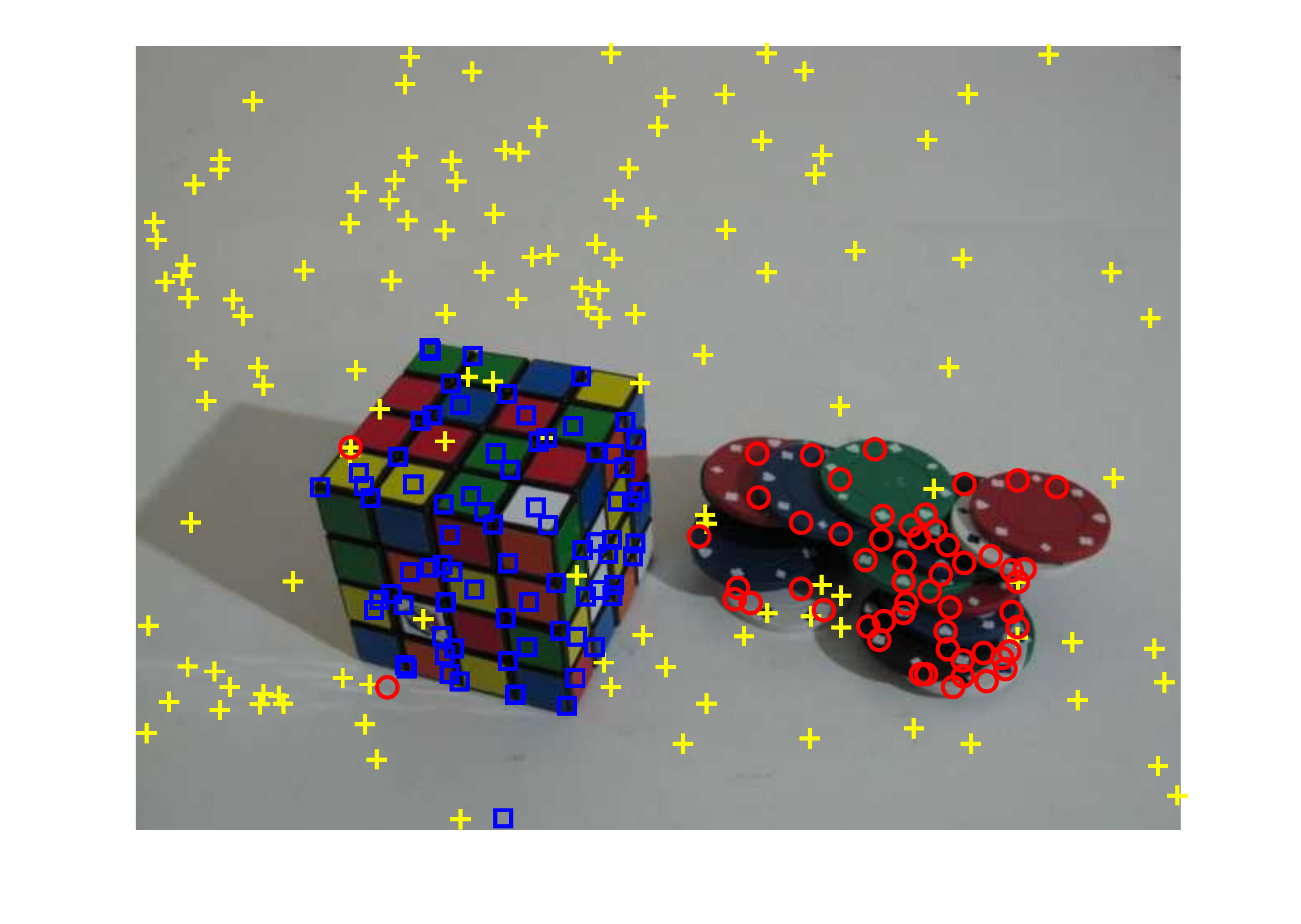}}
  \begin{center} (a)  \end{center}
\end{minipage}
\begin{minipage}[t]{.12\textwidth}
  \centering
  \centerline{\includegraphics[width=1.15\textwidth]{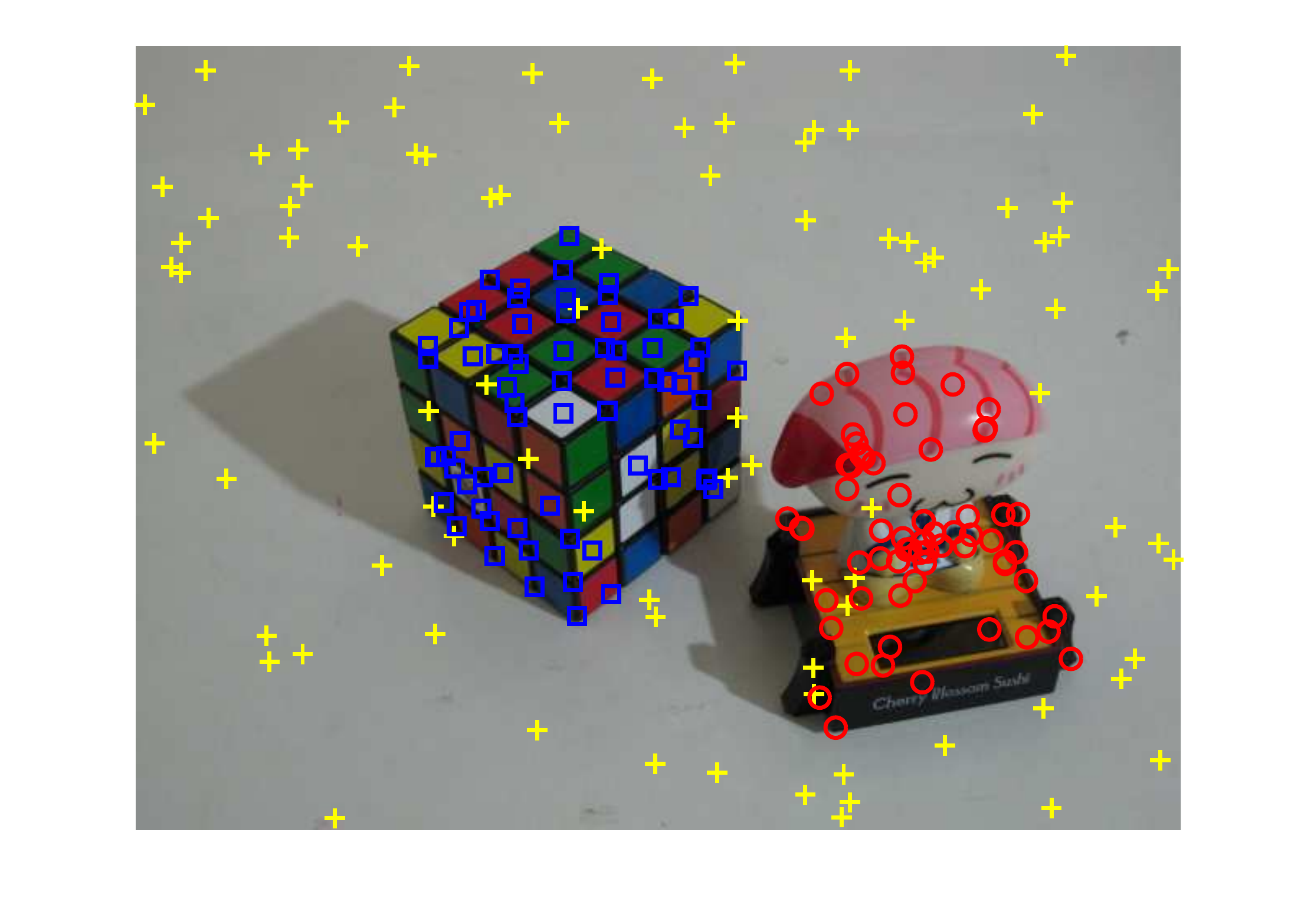}}
  \centerline{\includegraphics[width=1.15\textwidth]{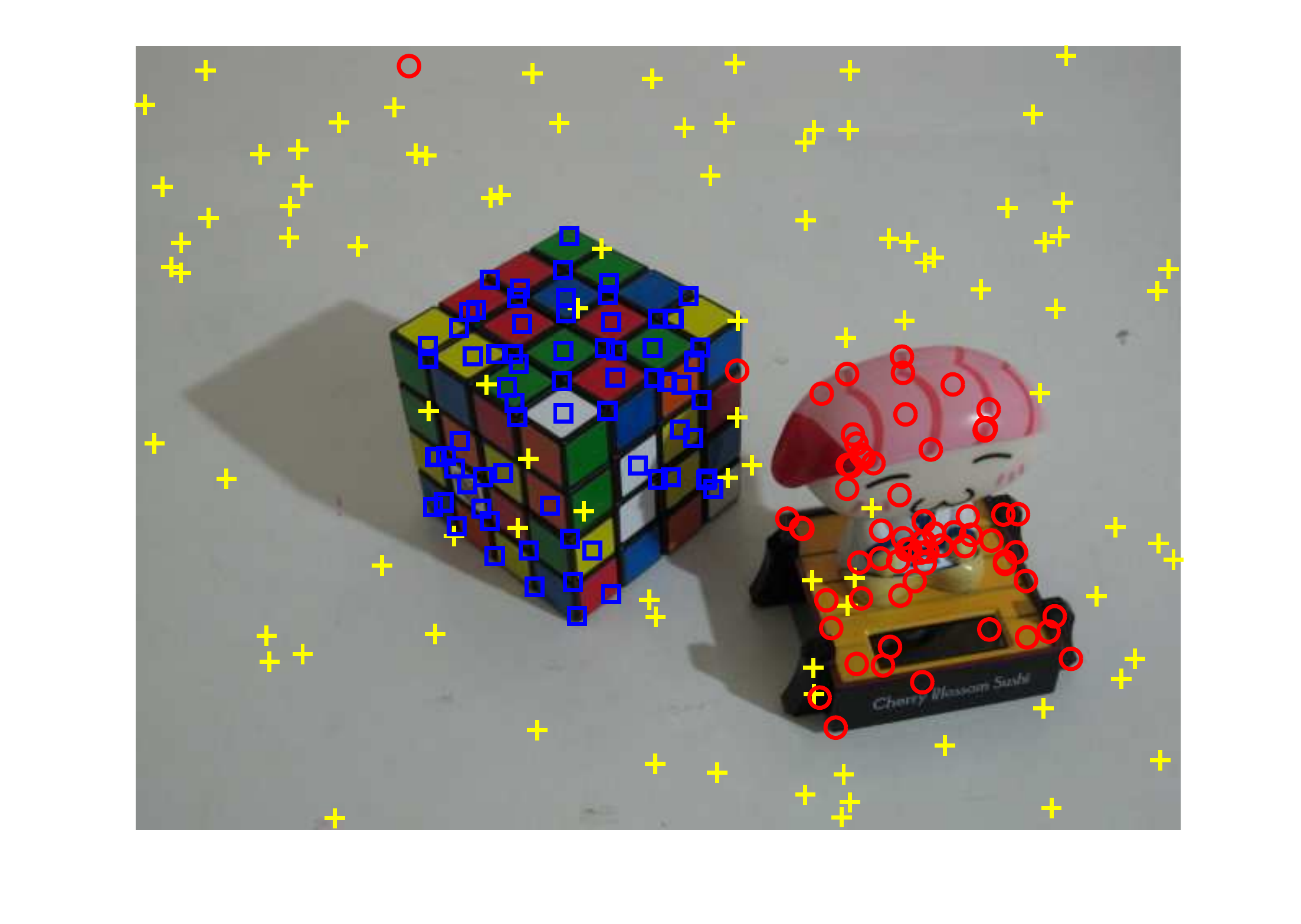}}
  \begin{center} (b)  \end{center}
\end{minipage}
\begin{minipage}[t]{.12\textwidth}
  \centering
  \centerline{\includegraphics[width=1.15\textwidth]{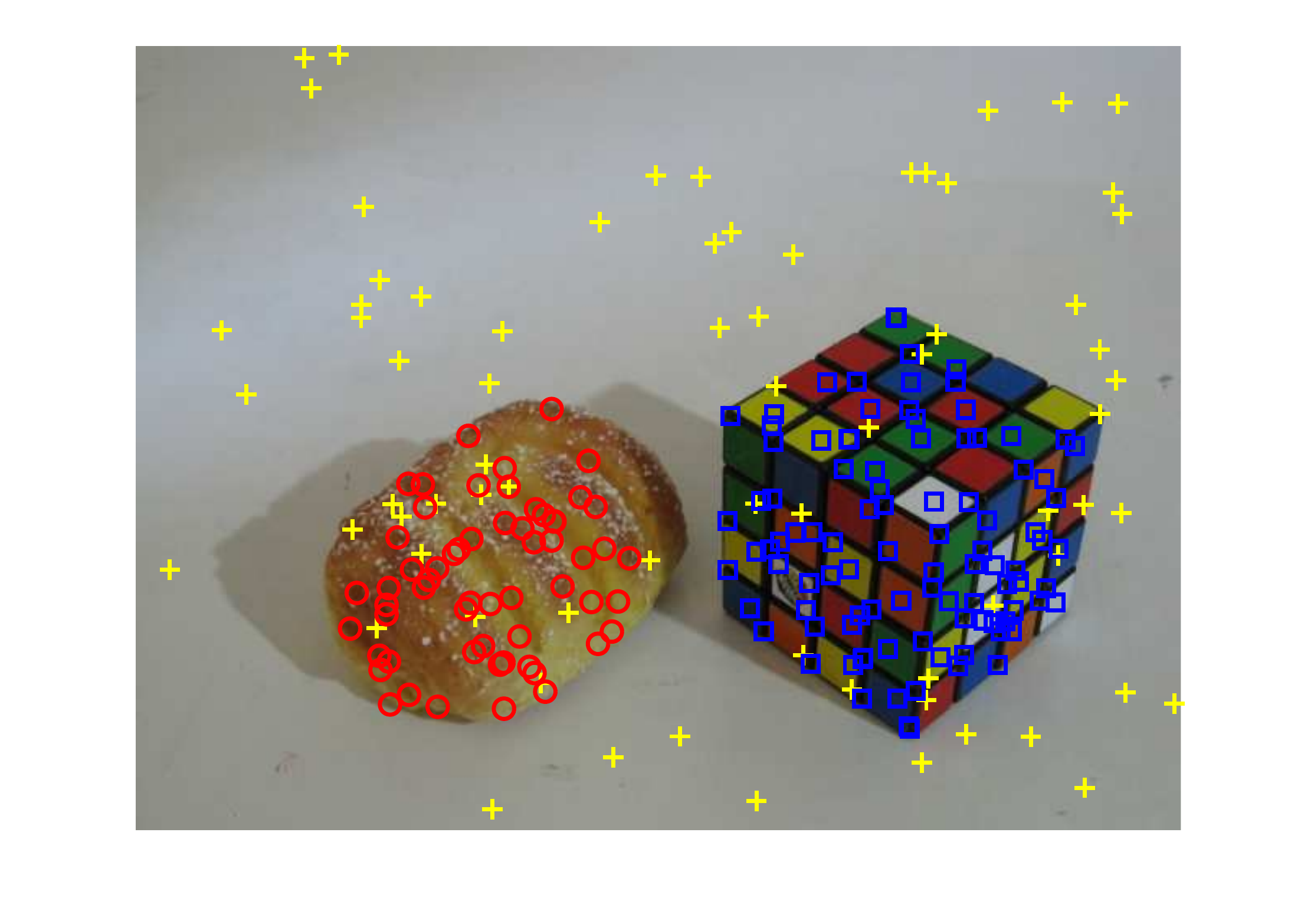}}
  \centerline{\includegraphics[width=1.15\textwidth]{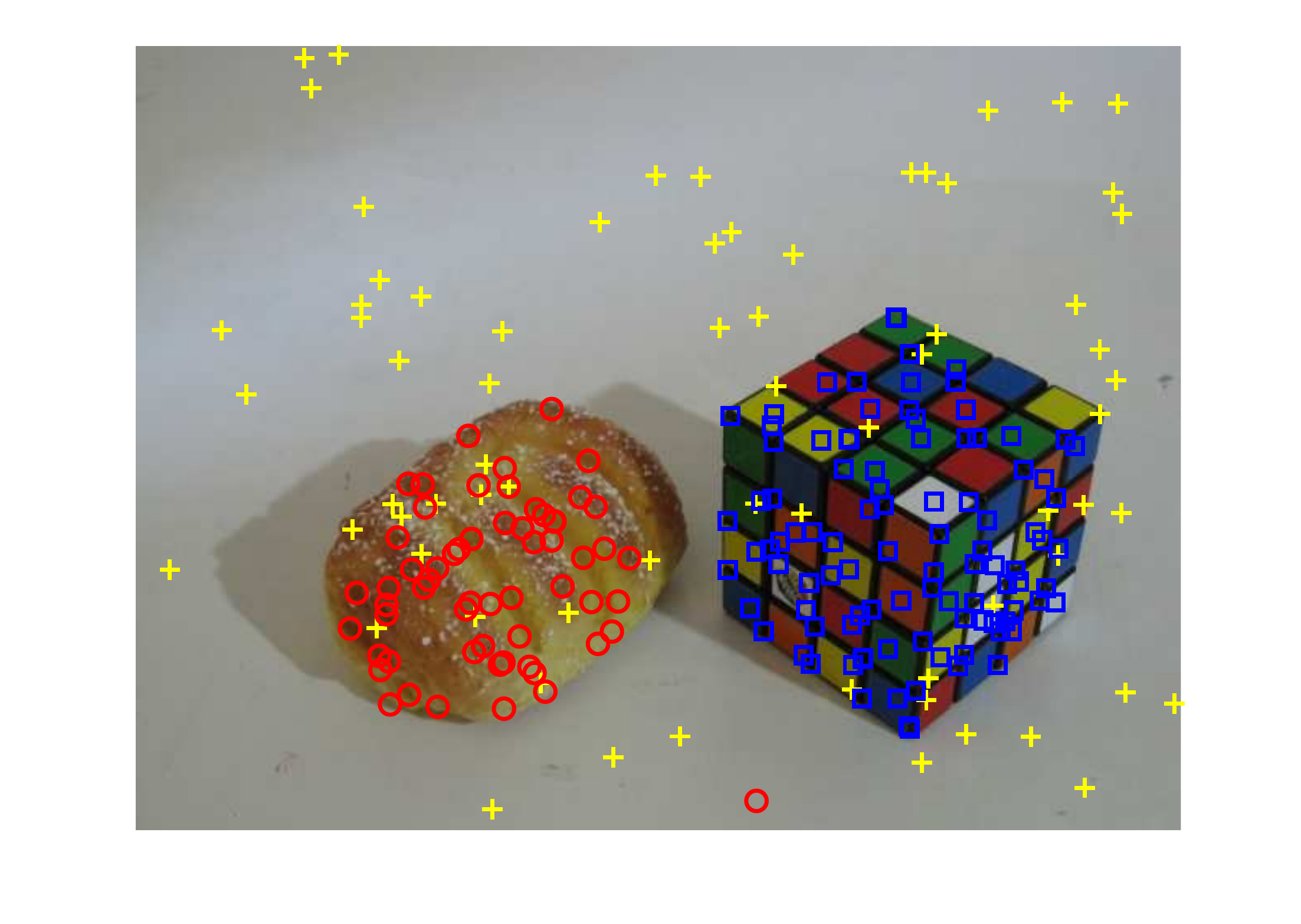}}
  \begin{center} (c) \end{center}
\end{minipage}
\begin{minipage}[t]{.12\textwidth}
  \centering
  \centerline{\includegraphics[width=1.15\textwidth]{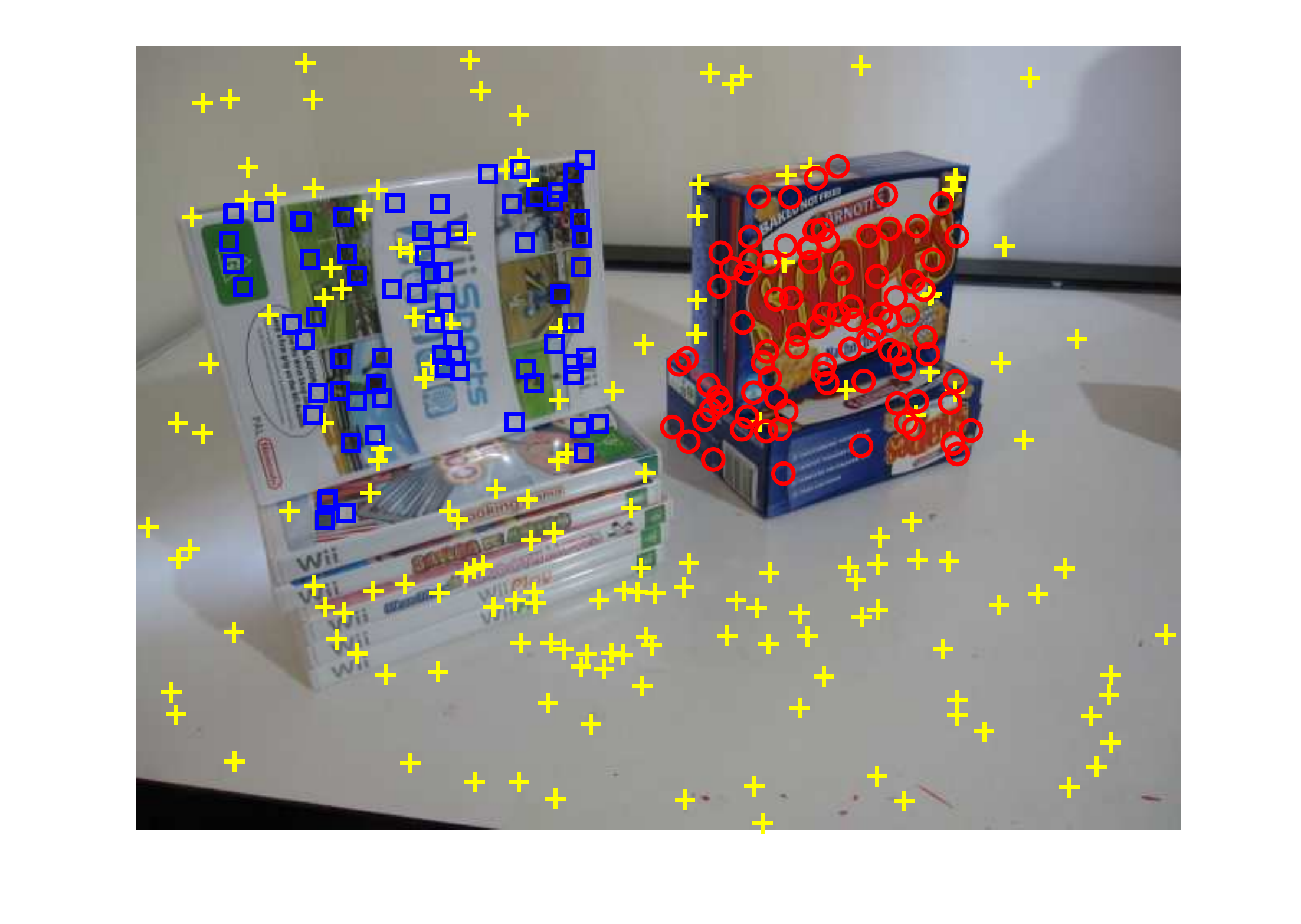}}
  \centerline{\includegraphics[width=1.15\textwidth]{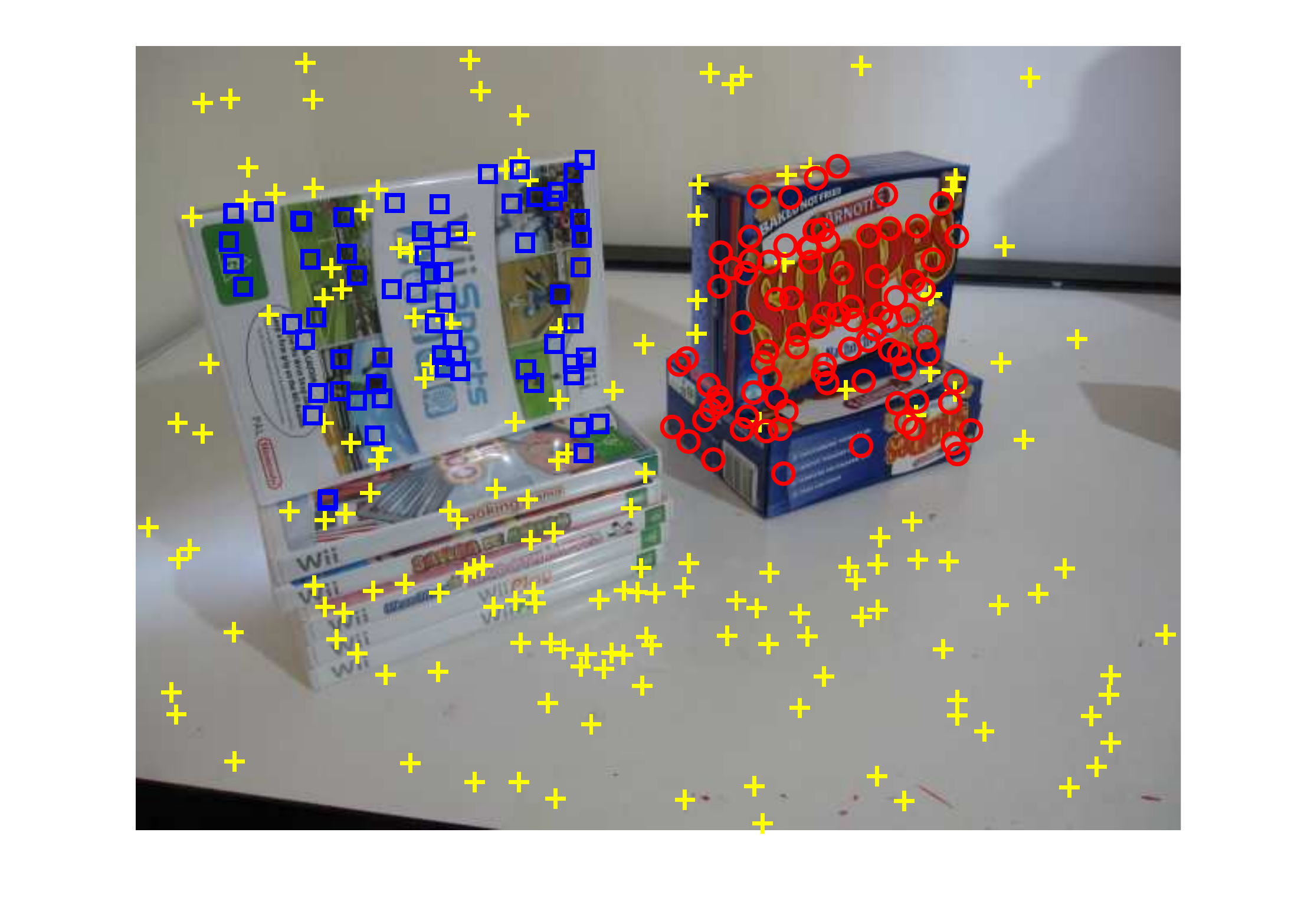}}
  \centerline{(d)  }\medskip
\end{minipage}
\begin{minipage}[t]{.12\textwidth}
  \centering
  \centerline{\includegraphics[width=1.15\textwidth]{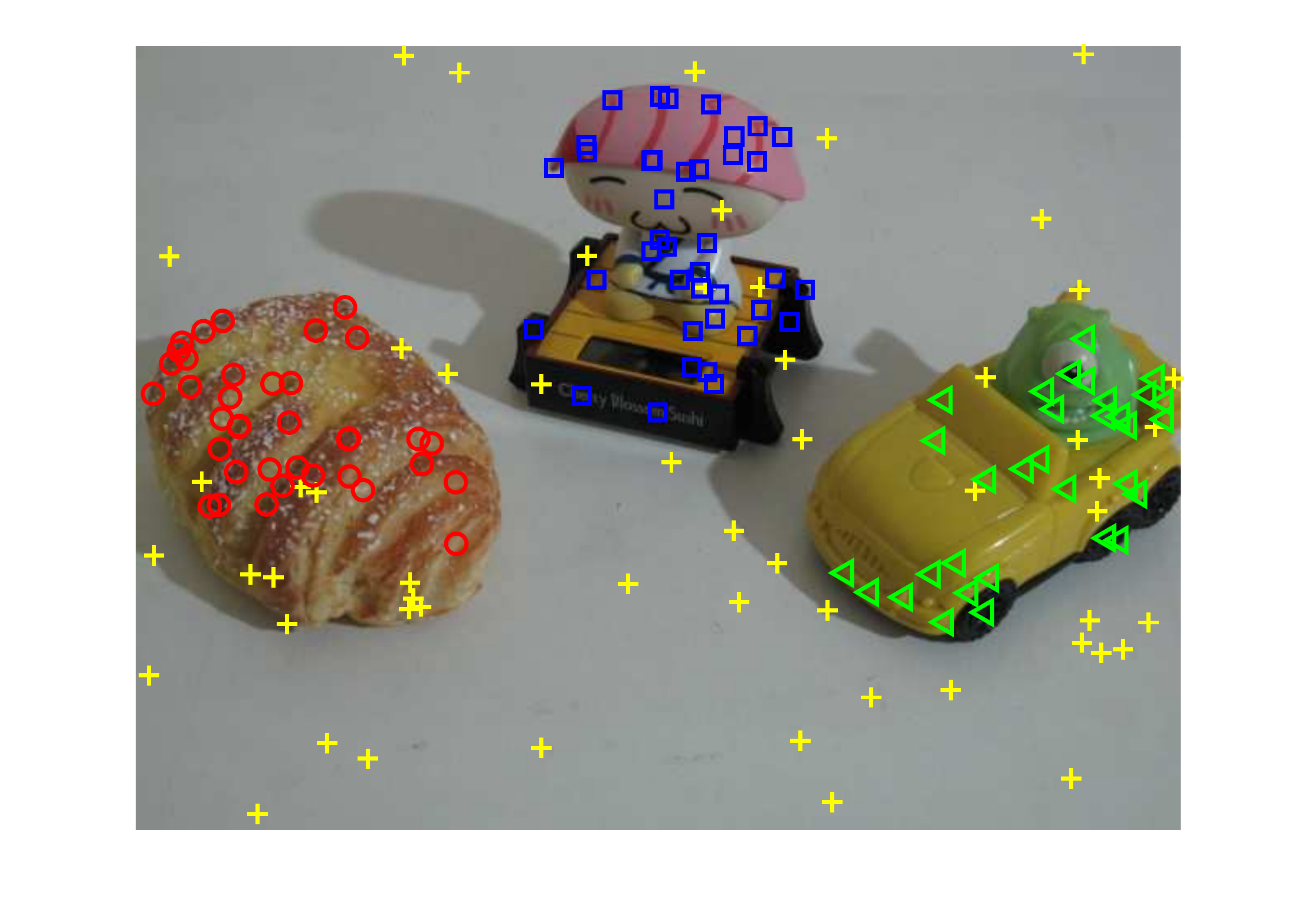}}
  \centerline{\includegraphics[width=1.15\textwidth]{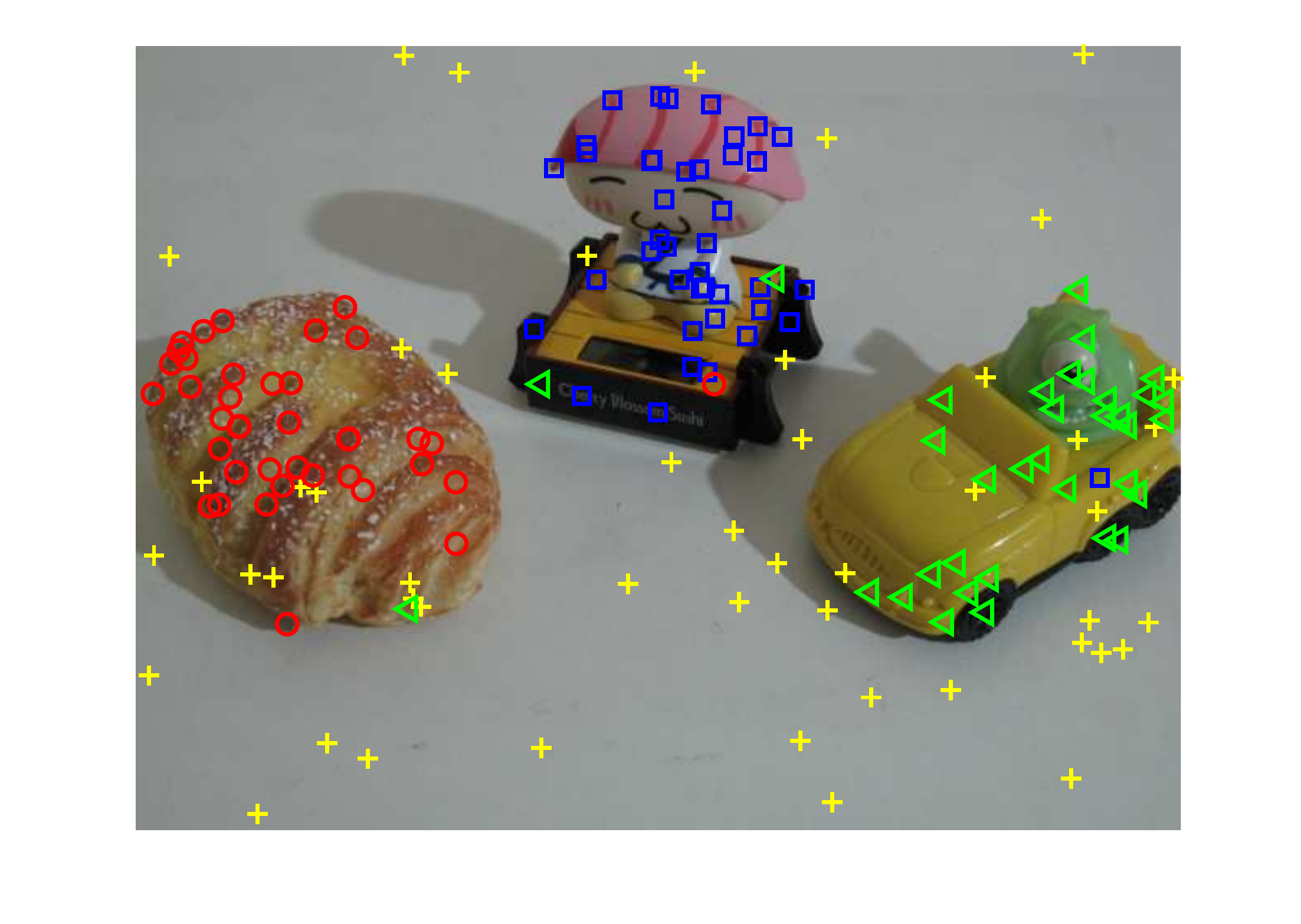}}
  \centerline{(e)  }\medskip
\end{minipage}
\begin{minipage}[t]{.12\textwidth}
  \centering
  \centerline{\includegraphics[width=1.15\textwidth]{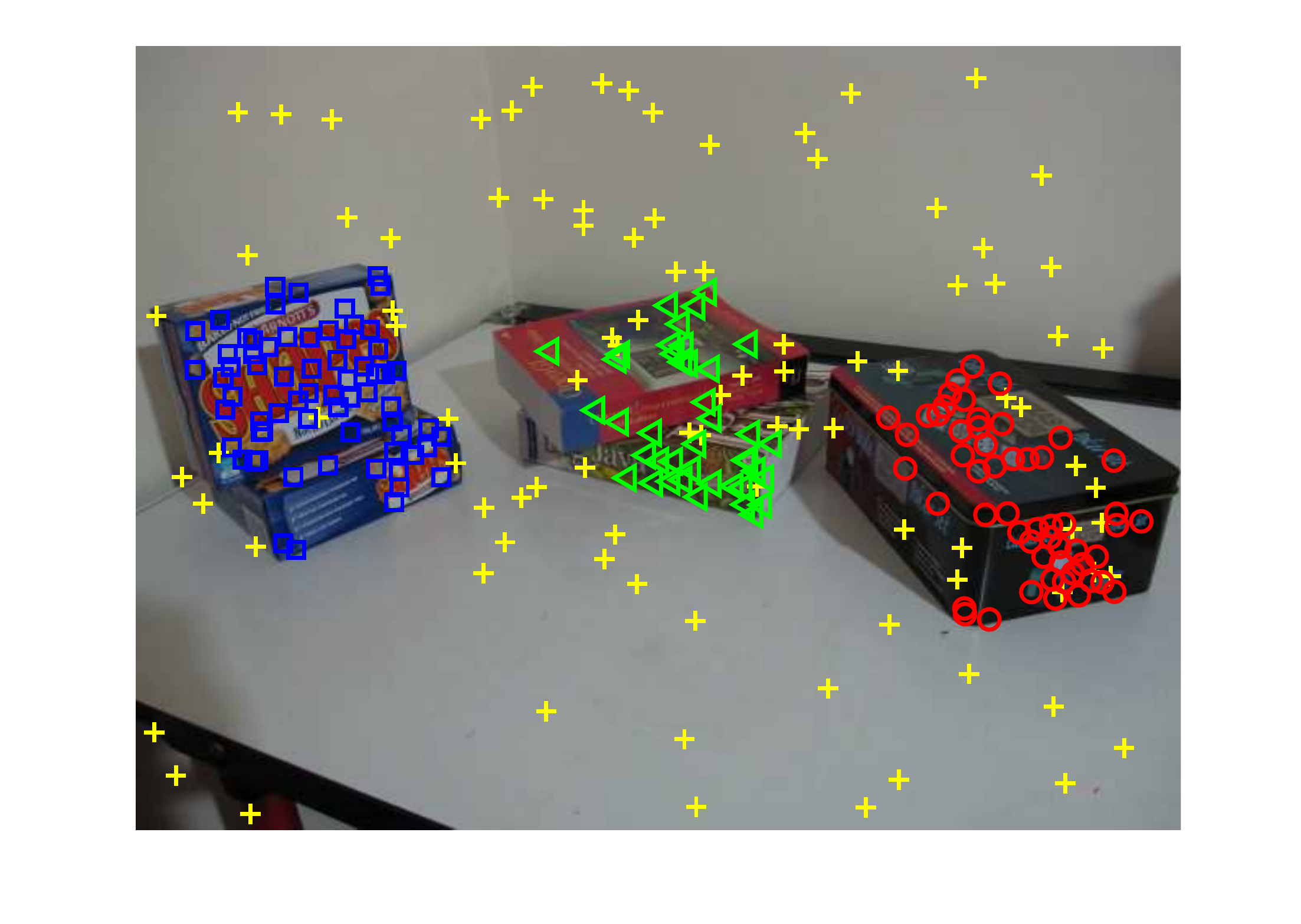}}
  \centerline{\includegraphics[width=1.15\textwidth]{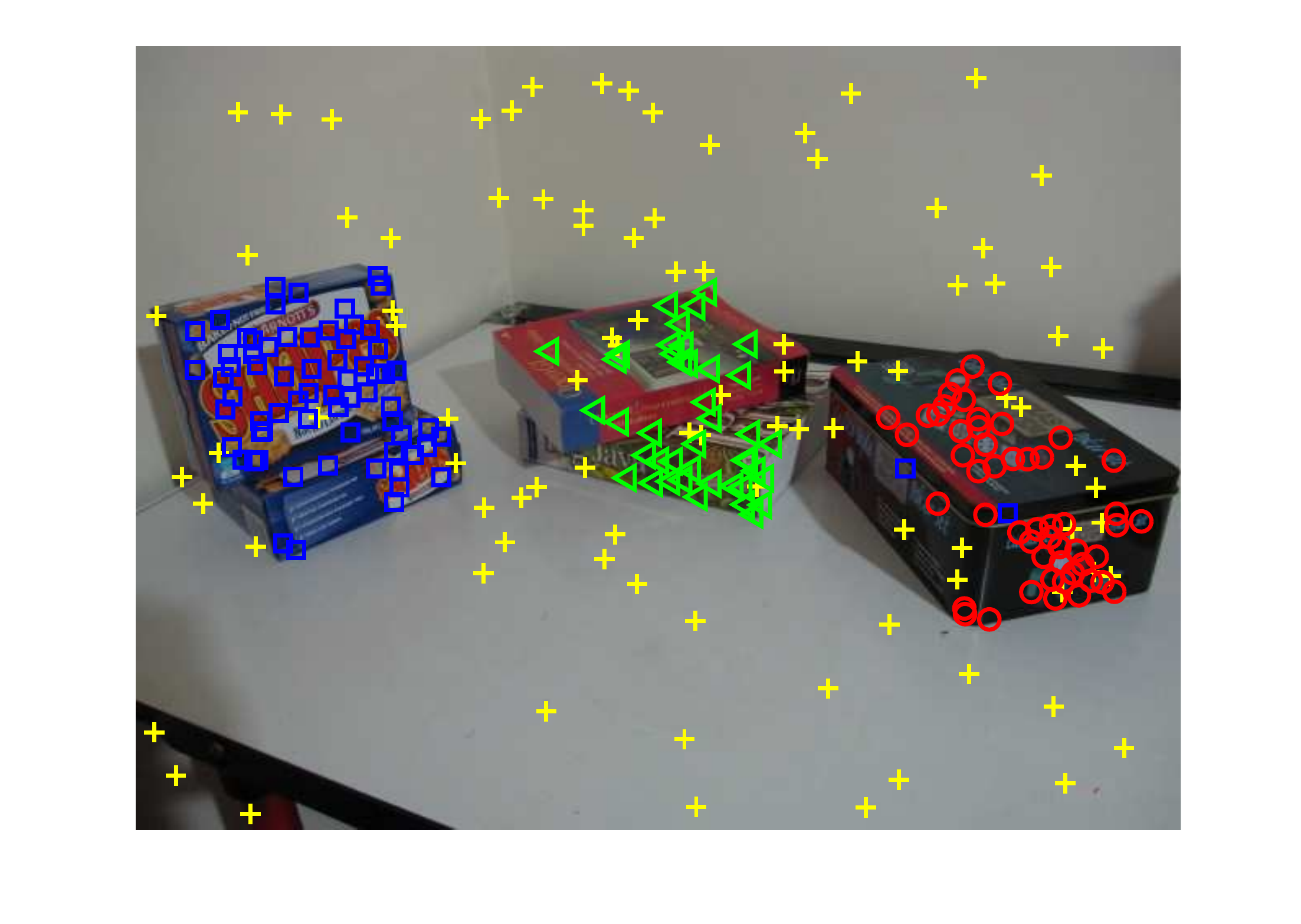}}
  \centerline{(f)  }\medskip
\end{minipage}
\begin{minipage}[t]{.12\textwidth}
  \centering
  \centerline{\includegraphics[width=1.15\textwidth]{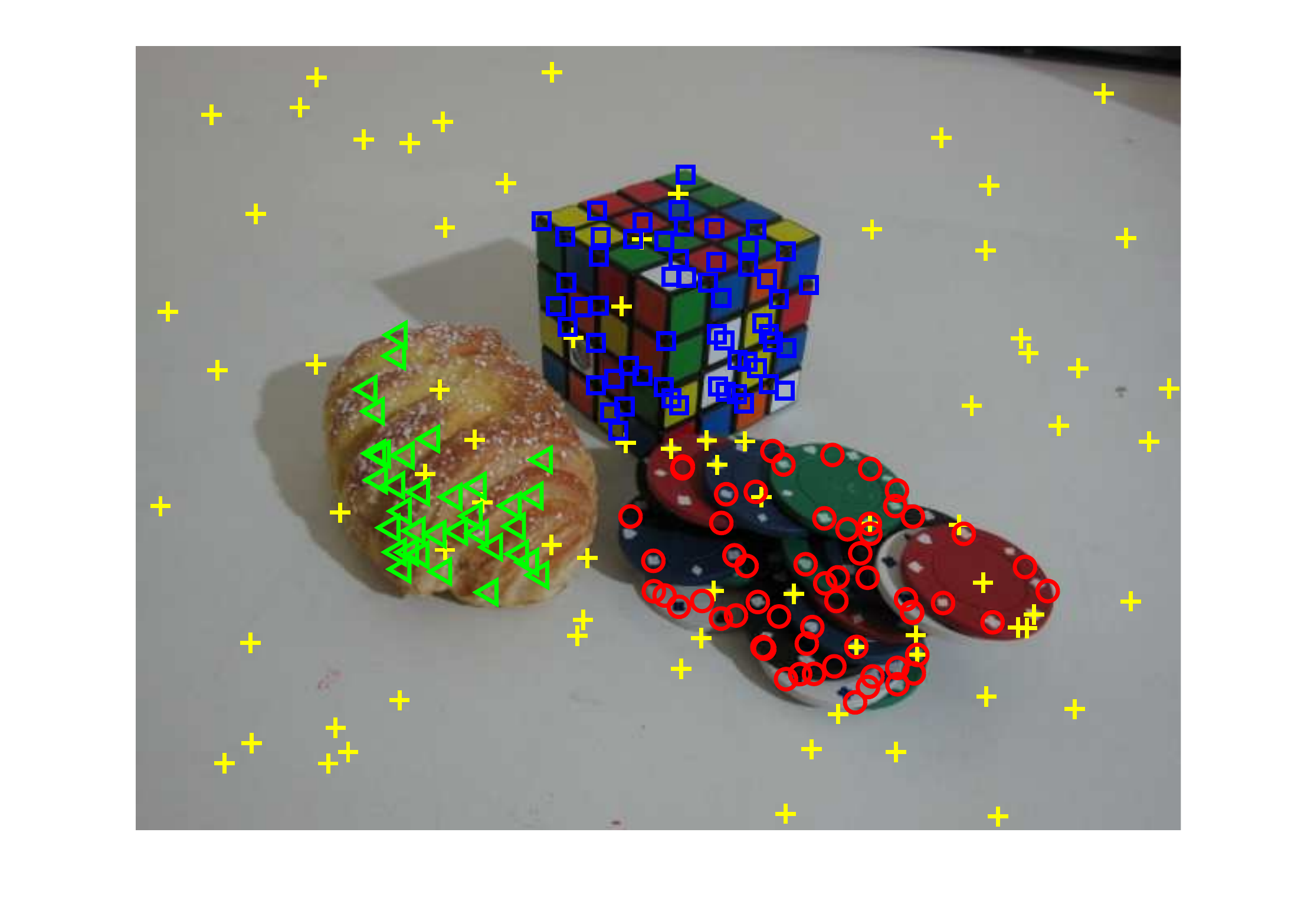}}
  \centerline{\includegraphics[width=1.15\textwidth]{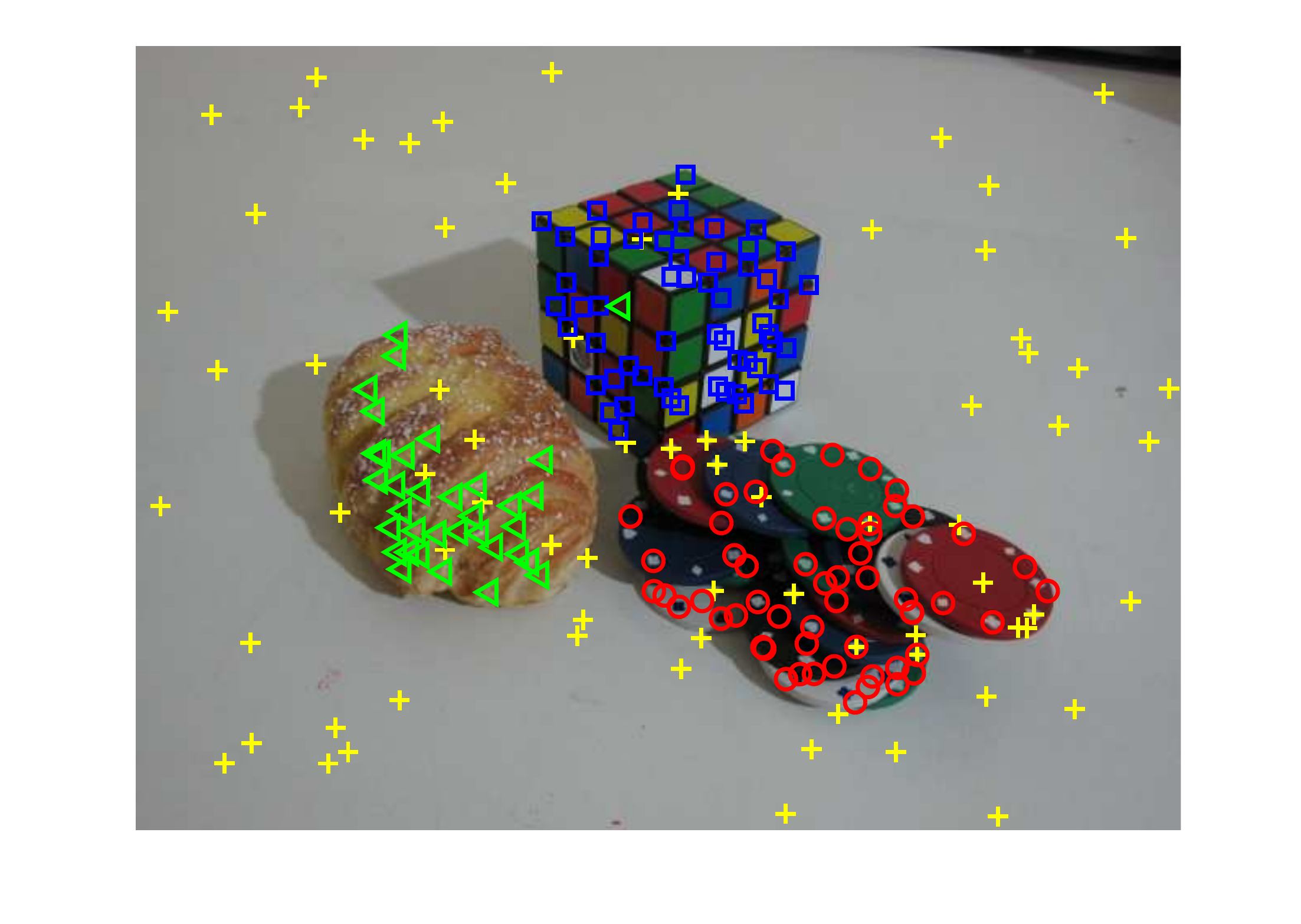}}
  \centerline{(g)  }\medskip
\end{minipage}
\begin{minipage}[t]{.12\textwidth}
  \centering
  \centerline{\includegraphics[width=1.15\textwidth]{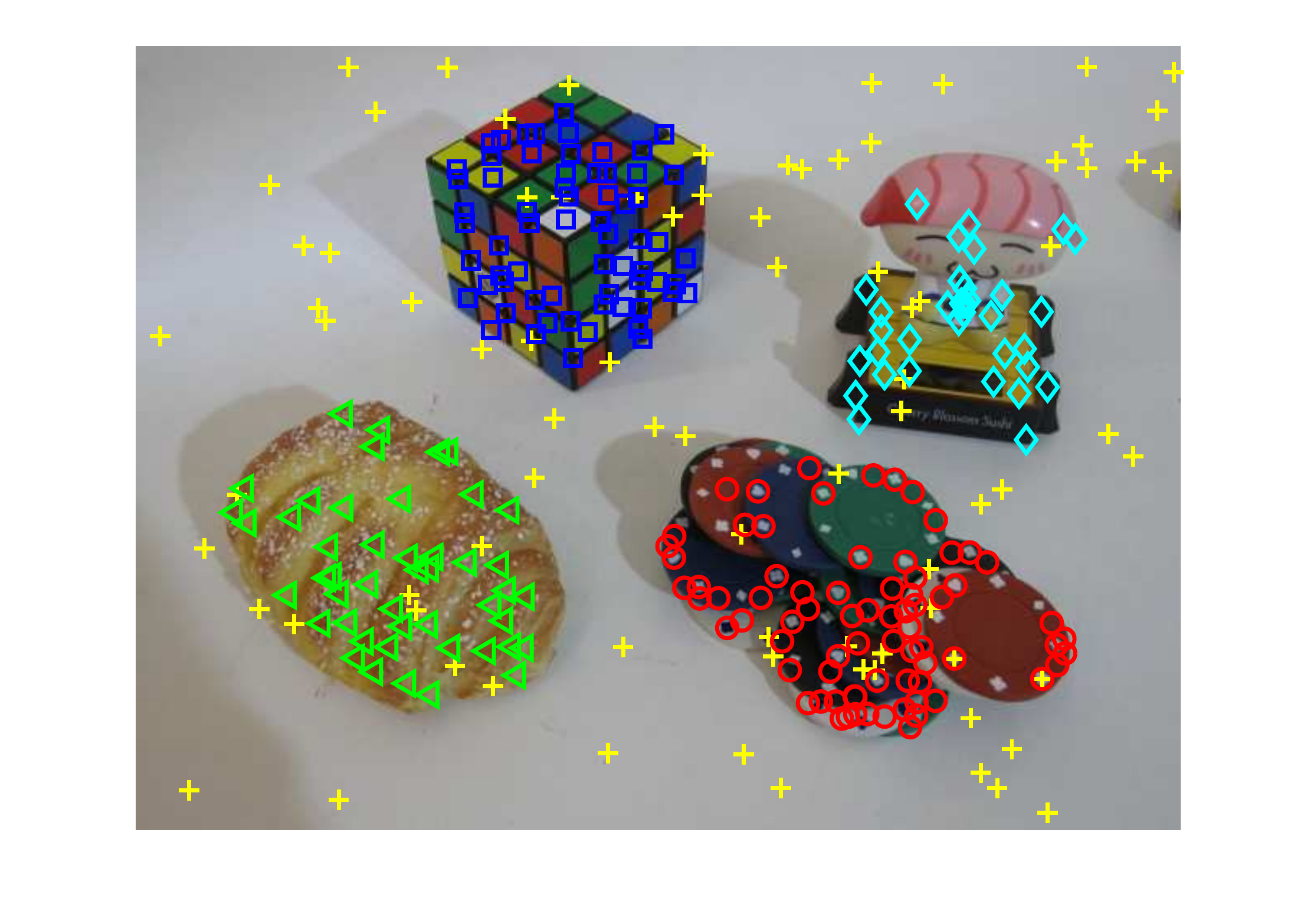}}
  \centerline{\includegraphics[width=1.15\textwidth]{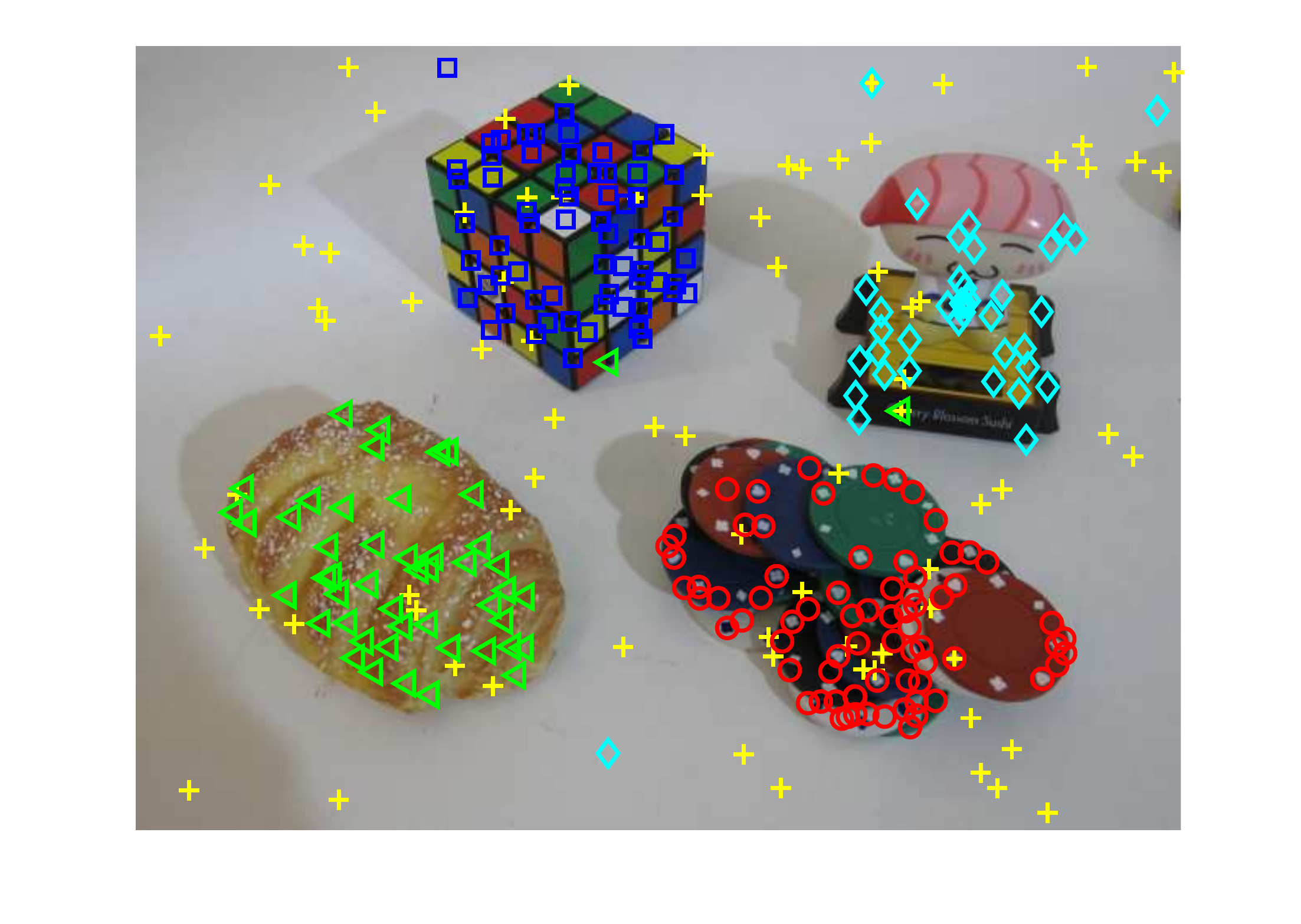}}
  \centerline{(h)  }\medskip
\end{minipage}
\hfill
\caption{Two-view based motion segmentation on eight image pairs, namely (a) Cubechips, (b) Cubetoy, (c) Breadcube, (d) Gamebiscuit, (e) Breadtoycar, (f) Biscuitbookbox, (g) Breadcubechips and (h) Cubebreadtoychips. The first and second rows are the original images with the ground truth results and the segmentation results obtained by MSH, respectively.}
\label{fig:motionsegmentation}
\end{figure*}
\begin{table*}
\small\
  \caption{The fitting errors (in percentage) for two-view based motion segmentation on eight dataset (the best results are boldfaced)}
\medskip
\centering
\begin{tabular}{|c|c|c|c|c|c|c|c|c|c|c|}
\hline
\multirow{2}{*}{} & \multicolumn{2}{c|}{KF} & \multicolumn{2}{c|}{RCG}& \multicolumn{2}{c|}{AKSWH} & \multicolumn{2}{c|}{T-linkage}&\multicolumn{2}{c|}{MSH}\\
\cline{2-11}
& Avg. &  Min. & Avg. & Min. & Avg. &Min.& Avg. & Min.& Avg. & Min. \\
\hline
Cubechips &{8.42}&{4.23}& {13.43}&{9.52}&{4.72}& {\bf2.11}&{5.63}&{2.46}& {\bf3.80}&{\bf2.11} \\
 \hline
Cubetoy   & {12.53}& {2.81} & {13.35}& {10.92} & {7.23} & {4.02} & {5.62} & {4.82} & {\bf3.21}& {\bf1.61}\\
 \hline
Breadcube & {14.83} & {4.13}& {12.60} & {8.07} & {5.45} & {1.42}& {4.96} & {1.32} & {\bf2.69}& {\bf0.83} \\
\hline
Gamebiscuit & {13.78}& {5.10}& {9.94} & {3.96} & {7.01}& {5.18}& {7.32} & {3.54}  & {\bf3.72}& {\bf1.22} \\
\hline
Breadtoycar & {16.87} & {14.55} & {26.51}& {19.54}& {9.04} &{8.43}& {\bf4.42}& {\bf4.00} & {6.63}& {4.55} \\
\hline
Biscuitbookbox & {16.06}& {14.29} & {16.87}& {14.36} &{8.54} & {4.99 }& {1.93} &{\bf1.16}&  {\bf1.54}& {\bf1.16}\\
\hline
Breadcubechips  & {33.43} & {21.30}& {26.39} & {20.43} & {7.39}& {3.41}& {\bf1.06} & {0.86} & {1.74} & {\bf 0.43} \\
\hline
Cubebreadtoychips& {31.07}& {22.94}& {37.95}& {20.80} & {14.95}& {13.15} & {\bf3.11}& {\bf3.00} &{4.28}& {3.57} \\
\hline
\end{tabular}
 \label{table:motionsegmentation}
\end{table*}
\subsection{Synthetic Data}
\label{sec:SyntheticData}
We evaluate the performance of the five fitting methods on line fitting using four challenging synthetic data in the $3$D space (see Fig.~\ref{fig:fivelines}). We repeat the experiment $50$ times and report the standard, the average and the best results of the fitting errors obtained by the competing methods, respectively, in Table~\ref{table:3Dlinetable}. We also show the corresponding average fitting results obtained by all the competing methods in Fig.~\ref{fig:fivelines}(b) to Fig.~\ref{fig:fivelines}(f).

From Fig.~\ref{fig:fivelines} and Table~\ref{table:3Dlinetable}, we can see that: (1) For the ``three lines'' data, the three lines are completely separable in the $3$D space, and the five fitting methods succeed in fitting all the three lines. However, MSH achieves the best performance among the five fitting methods. (2) For the ``four lines" data, the four lines intersect at one point. The five fitting methods succeed in estimating the number of the lines in data, but the data clustering based methods (i.e., KF and T-linkage), can not effectively deal with the data points near the intersection. In contrast, RCG, AKSWH and MSH correctly fit the four lines with lower fitting errors, while MSH achieves the lowest fitting error. (3) For the ``five lines" data, there exist two intersections. As mentioned before, the data points near the intersections are not correctly segmented by both KF and T-linkage, which causes these two methods to obtain high fitting errors. RCG correctly fits four lines but wrongly fits one. This is because the dense subgraph representing a potential structure in data is not effectively detected by RCG. In contrast, the parameter space based methods (i.e., AKSWH and MSH) are not very sensitive to data distribution. Both AKSWH and MSH correctly fit all the five lines with low fitting errors. (4) For the ``six lines" data, RCG correctly fits five of the six lines, and T-linkage wrongly estimates the number of lines in data. KF achieves the worst performance among the five fitting methods. In contrast, both AKSWH and MSH correctly fit the six lines. This challenging dataset further shows the superiority of the parameter space based methods over the other types of fitting methods.

\subsection{Real Images}
\subsubsection{Line Fitting}
\label{sec:linefitting}
We evaluate the performance of all the competing fitting methods using real images for line fitting (see Fig.~\ref{fig:linefiting}). For the ``tracks" image, which includes seven lines, there are $6,704$ edge points detected by the Canny operator \cite{canny1986computational}. As shown in Fig.~\ref{fig:linefiting}, AKSWH, T-linkage and MSH correctly fit all the seven lines. RCG correctly estimates the number of the lines but some lines are overlapped and two lines are missed because the potential structures in data are not correctly estimated during detecting the dense subgraphs. KF only correctly fits three out of the seven lines because many inliers belonging to the other four lines are wrongly removed.

For the ``pyramid" image, which includes four lines with a large number of outliers, and there are $5,576$ edge points detected by the Canny operator. KF, T-linkage and MSH succeed in fitting all the four lines, but KF wrongly estimates the number of lines. In contrast, both RCG and AKSWH only correctly fit three out of the four lines although RCG successfully estimates the number of lines in data. AKSWH can detect four lines after clustering hypotheses, but two lines are wrongly fused during the fusion step in AKSWH.

\subsubsection{Circle Fitting}
\label{sec:circlefitting}
We evaluate the performance of the five fitting methods using real images for circle fitting (see Fig.~\ref{fig:circlefiting}). For the ``coins" image, which includes five circles with similar number of inliers, there are $4,595$ edge points detected by the Canny operator. As shown in Fig.~\ref{fig:circlefiting}, AKSWH, T-linkage and MSH correctly fit all the five circles. In contrast, two model hypotheses estimated by KF overlap to one circle, and RCG correctly fits only four out of the five circles.

For the ``bowls" image, which includes four circles with obviously unbalanced numbers of inliers, $1,689$ edge points are detected by the Canny operator. We can see that two estimated circles by both KF and RCG overlap in the image. AKSWH correctly fits three circles but misses one circle because most of model hypotheses generated for the circle with a small number of inlier data points are removed when AKSWH selects significant model hypotheses. In contrast, both T-linkage and MSH succeed in fitting all the four circles in this challenging case.

\subsubsection{Homography Based Segmentation}
\label{sec:homographbasedsegmentation}
We also evaluate the performance of the five fitting methods using the eight real image pairs from the AdelaideRMF dataset~\cite{wong2011dynamic}\footnote{\url{http://cs.adelaide.edu.au/~hwong/doku.php?id=data}} for homography based segmentation. We repeat each experiment 50 times, and show the average and the minimum fitting errors in Table~\ref{table:homographytable}. The fitting results obtained by MSH are also shown in Fig.~\ref{fig:homography}.

From Fig.~\ref{fig:homography} and Table~\ref{table:homographytable}, we can see that MSH obtains accurate results, achieving the lowest average fitting errors in $7$ out of $8$ data and the lowest minimum fitting errors in all the eight data. Both AKSWH and T-linkage succeed in fitting $7$ out of $8$ data with low fitting errors. In contrast, KF and RCG achieve worse results. We note that many outliers are clustered with inliers when KF uses the proximity sampling and RCG is very sensitive to its parameters when there exists many bad model hypotheses.
\subsubsection{Two-view Based Motion Segmentation}
\label{sec:motionsegmentation}
For the two-view based motion segmentation problem, we use the eight real image pairs from the AdelaideRMF dataset~\cite{wong2011dynamic} to quantitatively compare the performance of MSH with the other four competing fitting methods. We also report the average and the minimum fitting errors in Table~\ref{table:motionsegmentation} by repeating each experiment 50 times. The fitting results obtained by MSH are also shown in Fig.~\ref{fig:motionsegmentation}.

From Fig.~\ref{fig:motionsegmentation} and Table~\ref{table:motionsegmentation}, we can see that both KF and RCG achieve bad results and fail in most cases. This is because when a large number of model hypotheses are generated for two-view based motion segmentation to cover all the model instances in data, a large proportion of bad model hypotheses may lead to inaccurate similarity measure between data points, which results in a wrong estimate of the parameters and of the number of model instances by KF and RCG. AKSWH achieves better results than both KF and RCG on average fitting errors. However, AKSWH may remove some good model hypotheses that correspond to model instances when it selects significant hypotheses especially for the unbalanced data, which results in a high fitting error. T-linkage and MSH succeed in fitting all the eight data with low fitting errors, while MSH obtains relatively better results (as shown in Fig.~\ref{fig:motionsegmentation}) and achieves the lowest average fitting errors in $5$ out of $8$ data, and the lowest minimum fitting errors in $6$ out of $8$ data.

\section{Conclusions}
\label{sec:conclusion}
This paper formulates geometric model fitting as a mode-seeking problem on a hypergraph in which each vertex represents a model hypothesis and each hyperedge denotes a data point. Based on the hypergraph, we propose a novel mode-seeking algorithm (MSH), which searches for authority peaks by analyzing the similarity between vertices. MSH simultaneously estimates the number and the parameters of model instances in the parameter space, which can alleviate sensitivity to unbalanced data effectively. MSH is scalable to large scale problems. Results on both synthetic data and real images have demonstrated that the proposed method significantly outperforms several other start-of-the-art fitting methods.

\section*{Acknowledgment}
\small{This work was supported by the National Natural Science Foundation of China under Grants 61472334, 61170179, and 61571379, and supported by the Fundamental Research Funds for the Central Universities under Grant 20720130720. David Suter acknowledged funding under ARC DPDP130102524.}

\end{document}